%% file: main.tex
\documentclass[10pt,twocolumn,letterpaper]{article}

\usepackage[pagenumbers]{cvpr} 


\definecolor{cvprblue}{rgb}{0.21,0.49,0.74}
\usepackage[pagebackref,breaklinks,colorlinks,allcolors=cvprblue]{hyperref}

\usepackage{graphicx}
\usepackage{booktabs}
\usepackage{colortbl}
\usepackage{algorithm}
\usepackage{amssymb}
\usepackage{pifont}
\usepackage{cuted}
\usepackage{array}
\usepackage{xcolor}
\usepackage{mdframed}
\usepackage{adjustbox}
\usepackage{varwidth}
\definecolor{darkred}{rgb}{0.839, 0.007, 0.125}

\newcommand*\samethanks[1][\value{footnote}]{\footnotemark[#1]}

\newcommand{\cmark}{\ding{51}}
\newcommand{\xmark}{\ding{55}}

\usepackage{algpseudocode}
\def\paperID{8122} 
\def\confName{CVPR}
\def\confYear{2025}

\title{Bridging the Vision-Brain Gap with an Uncertainty-Aware Blur Prior}
\author{
    Haitao Wu$^{1}$\quad Qing Li$^{1,2}$\quad
    Changqing Zhang$^{1}$\thanks{Corresponding authors.}\quad 
    Zhen He$^{2}$\samethanks \quad Xiaomin Ying$^{2}$\samethanks \vspace{4pt}\\
    $^{1}$College of Intelligence and Computing, Tianjin University\\
    $^{2}$Beijing Institute of Basic Medical Sciences\\ 
    \tt\{wuhaitao, liqing0315, zhangchangqing\}@tju.edu.cn\\
    \tt{hezhen.bio@gmail.com, yingxmbio@foxmail.com} \\
}

\begin{document}
\maketitle
\input{sec/0_abstract}    
\input{sec/1_introduction}
\input{sec/2-related_work}
\input{sec/3_theory}
\input{sec/4_method}

\input{sec/5_experiments_results}
\input{sec/6_conclusion}
\clearpage
{
    \small
    \bibliographystyle{ieeenat_fullname}
    \bibliography{main}
}

\input{sec/X_suppl}

\end{document}

%% file: sec/0_abstract.tex
\begin{abstract}
Can our brain signals faithfully reflect the original visual stimuli, even including high-frequency details? Although human perceptual and cognitive capacities enable us to process and remember visual information, these abilities are constrained by several factors, such as limited attentional resources and the finite capacity of visual memory. When visual stimuli are processed by human visual system into brain signals, some information is inevitably lost, leading to a discrepancy known as the \textbf{System GAP}.
Additionally, perceptual and cognitive dynamics, along with technical noise in signal acquisition, degrade the fidelity of brain signals relative to the visual stimuli, known as the \textbf{Random GAP}.
When encoded brain representations are directly aligned with the corresponding pretrained image features, the System GAP and Random GAP between paired data challenge the model, requiring it to bridge these gaps.
However, in the context of limited paired data, these gaps are difficult for the model to learn, leading to overfitting and poor generalization to new data. 
To address these GAPs, we propose a simple yet effective approach called the \textbf{Uncertainty-aware Blur Prior (UBP)}.
It estimates the uncertainty within the paired data, reflecting the mismatch between brain signals and visual stimuli. Based on this uncertainty, UBP dynamically blurs the high-frequency details of the original images, reducing the impact of the mismatch and improving alignment.
Our method achieves a top-1 accuracy of \textbf{50.9\%} and a top-5 accuracy of \textbf{79.7\%} on the zero-shot brain-to-image retrieval task, surpassing previous state-of-the-art methods by margins of \textbf{13.7\%} and \textbf{9.8\%}, respectively. 
Code is available at \href{https://github.com/HaitaoWuTJU/Uncertainty-aware-Blur-Prior}{GitHub}.
\end{abstract}

%% file: sec/1_introduction.tex
\section{Introduction}
\label{sec:introduction}

\begin{figure*}[t]
  \centering
    \includegraphics[width=0.96\linewidth]{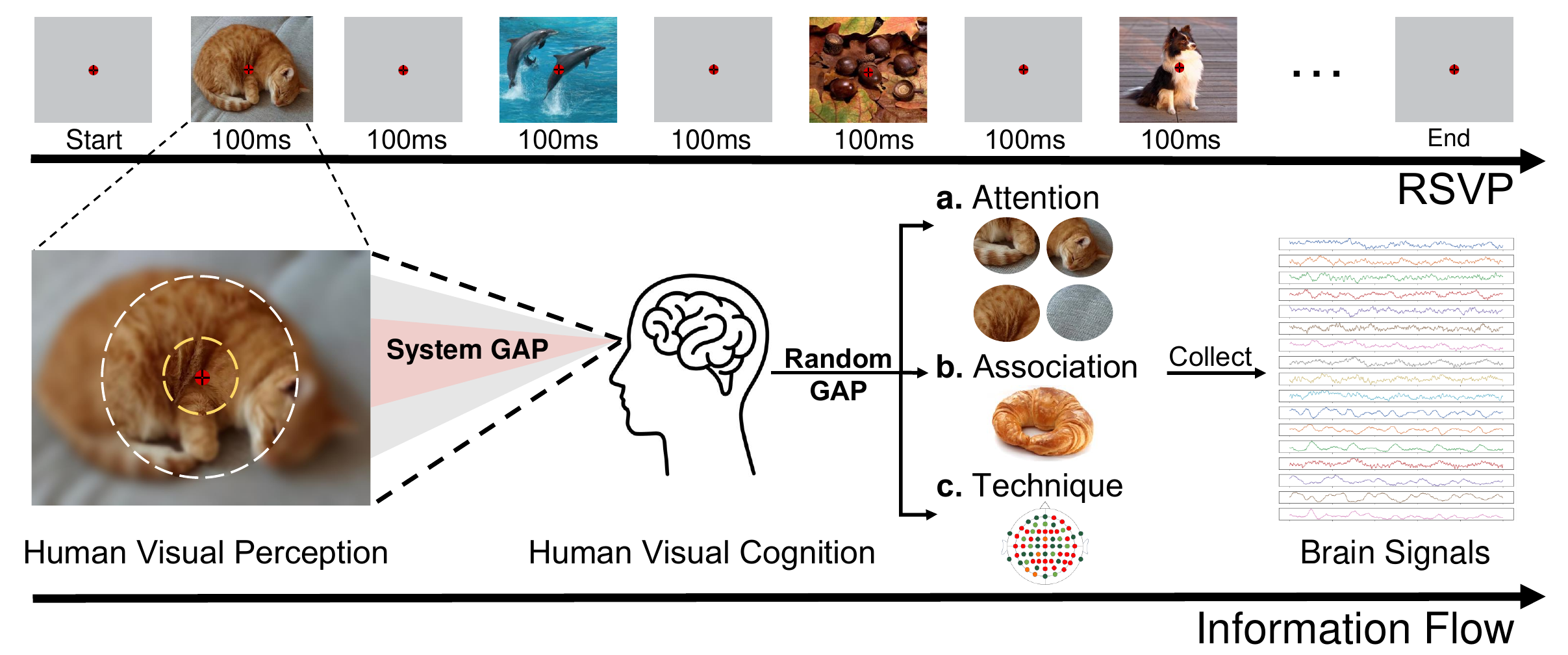}
    \caption{Overview of the information flow during Rapid Serial Visual Presentation (RSVP) and the GAPs in human visual perception and cognition. The top panel illustrates the RSVP paradigm, where a sequence of images is rapidly presented for 100ms each, with a fixation point in the center. The bottom panel highlights the GAPs in the visual processing pipeline: \textbf{System Gap}, which represents the loss of high-frequency details during the transition from raw visual stimuli to visual perception, and \textbf{Random Gap}, which arises due to (a) dynamic perceptual processes (e.g., shifts in visual attention), (b) dynamic cognitive processes (e.g., associating with similar objects or concepts), and (c) low-level technical noise in signal collection.}
    \label{fig:motivation}
\end{figure*}

The human brain is one of the most complex things known in the universe, and extensive studies have been devoted to unraveling its structure and function over the past several decades~\cite{hubel1959receptive,hubel1968receptive,livingstone1984anatomy,nauhaus2012orthogonal,lettvin1959frog, raichle2001default, posner1980attention}. Vision, as the primary sense for humans to perceive the world, involves approximately one-third of the cortical surface. Consequently, the brain plays a crucial role in visual perception and cognition~\cite{van1992information, wang1996optical, tsao2006cortical, liang2018fine}.
To understand the mechanisms between human vision and brain activity, various brain imaging techniques such as Electroencephalogram (EEG), Magnetoencephalography (MEG) and  Functional magnetic resonance imaging (fMRI), are utilized to measure brain responses to visual stimuli. 
EEG is a low-cost, portable method for measuring brain activity by detecting voltage changes caused by neuronal signals, offering high temporal resolution. However, it suffers from a low signal-to-noise ratio due to weak signals being influenced by the skull, external interference, and biological noise.
MEG offers high temporal resolution, but is limited by its high cost.
In contrast, fMRI provides high spatial resolution by detecting changes in blood oxygen levels, but its temporal resolution is limited due to the slower hemodynamic response.

Recently, various methods for decoding brain signals have been proposed~\cite{scotti2024reconstructing,takagi2023high,chen2023seeing,du2023decoding,song2024decoding,li2024visual,chen2024visual,benchetrit2023brain}. These methods aim to retrieve and reconstruct the original visual stimuli by aligning the representations of brain signals with the visual stimuli.
However, they fail to account for the GAPs between brain signals and visual stimuli. Previous studies~\cite{whitney2018ensemble,cohen2016bandwidth,block2011perceptual,buschman2011neural,pylyshyn1999vision} on human perceptual and cognitive capacities have shown that the amount of visual information human can process and remember at any given moment is limited and varies across individuals due to constrained attentional resources~\cite{cavanagh2005tracking,dux2009humans,simons1997change}, limitations in eye movements and scanning~\cite{kowler2011eye,wolfe1994guided}, and the limited capacity of visual working memory~\cite{luck2013visual}. When the digital image modality is transformed into the brain signal modality through human visual perception and cognitive processes, some information is unavoidably lost, which is called the \textbf{System GAP} between human and machine. A key factor contributing to this information loss is the structure of the human eye. As shown in~\cref{fig:motivation}, when an individual observes an object, the resolution of the visual field is not uniform and gradually decreases from the fovea toward the periphery~\cite{curcio1990human}.

\begin{figure*}[t]
  \centering
    \includegraphics[width=1.0\linewidth]{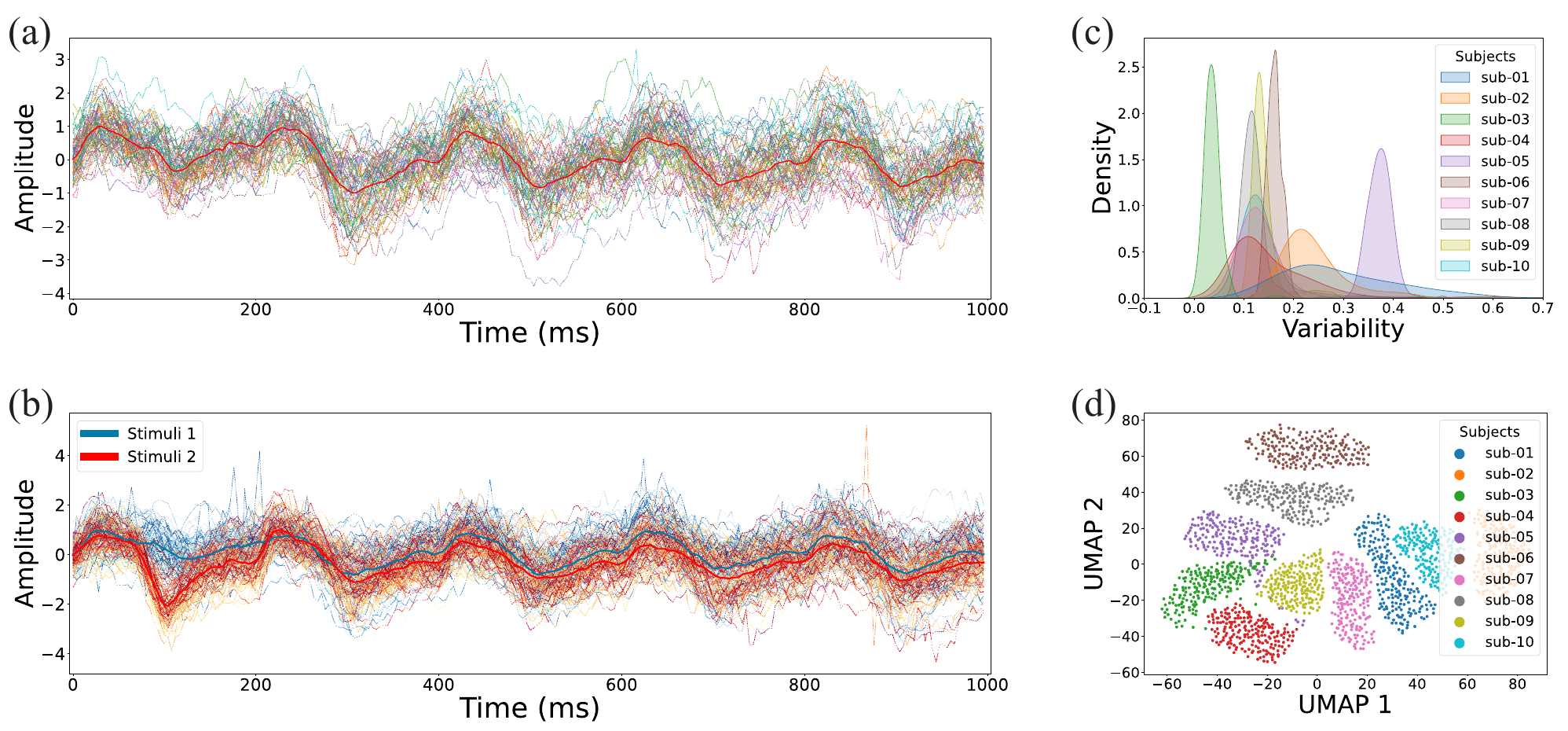}
    \caption{Illustration of brain signals. (a) EEG signals recorded over 80 trials of the same stimulus for Subject 1. The red line indicates the mean across all trials. (b) EEG signals from 80 trials of two stimuli for Subject 1. Cool colors represent Stimulus 1, warm colors represent Stimulus 2. The blue and red lines show the means for Stimulus 1 and Stimulus 2, respectively. (c) Density distribution of EEG signal variability across 10 subjects. Variability is negatively correlated with task performance and see~\cref{tab:correlation} for further details. (d) UMAP projection of EEG signals from 10 subjects, showing distinct clustering patterns.}
    \label{fig:random_gap}
\end{figure*}

Human perception and cognition are inherently dynamic, even when viewing or considering the same image or problem, leading to variability in brain signals responding to same stimuli.
As illustrated in \cref{fig:motivation}, perception can shift as attention is directed toward different parts of an image, while cognition may dynamically associate with related objects or concepts. Additionally, signal acquisition is impacted by technical noise, such as poor electrode-skin contact or instability in signal channels.
These factors contribute to variability in brain signals, as shown in \cref{fig:random_gap}(a), and weaken the information relative to the original visual stimuli, reducing the signal-to-noise ratio.
Consequently, even for two completely different stimuli, the corresponding brain signals are difficult to differentiate due to limited information and excessive noise, as shown in \cref{fig:random_gap}(b). We refer to this variability-induced information mismatch as \textbf{Random GAP}, attributed to its stochastic nature, which makes it exceptionally challenging to quantify. As illustrated in \cref{fig:random_gap}(c)(d), it demonstrates variability both across trials and among different subjects.

The most advanced visual neural decoding methods~\cite{du2023decoding, li2024visual, chen2024visual,song2024decoding} align encoded representations of brain signals with the pretrained embedding of corresponding visual stimuli by contrastive learning~\cite{chen2020simple}. However, when we directly align them, the System GAP and Random GAP may prompt the model to bridge the GAPs. 
Limited by the scarcity of paired data, the gaps become difficult for the model to learn, leading to overfitting on the training set and poor generalization to new data. To address this issue, we aim to mitigate the impact of the GAPs and improve the alignment by introducing priors, thereby preventing the model from overfitting to these gaps.
Our main contributions are summarized as follows:
\begin{enumerate}
    \item We propose the existence of \textbf{System GAP} and \textbf{Random GAP} between visual stimuli and brain signals as shown in \cref{fig:motivation}. The System GAP arises from the inability of brain signals to faithfully reflect visual stimuli, while the Random GAP arises from three factors: the dynamics of perception, the dynamics of cognition, and low-level technical noise in signal collection. These GAPs contribute to the reduction of the fidelity of brain signals in relation to the original visual stimuli.
    \item  We experimentally analyzed the impact of the two types of gaps. Based on observations and experimental analysis, we propose a simple and effective method called \textbf{U}ncertainty-aware \textbf{B}lur \textbf{P}rior (\textbf{UBP}). Our method achieves a top-1 accuracy of \textbf{50.9\%} and a top-5 accuracy of \textbf{79.7\%} on the zero-shot brain-to-image retrieval task, surpassing previous state-of-the-art methods by margins of \textbf{13.7\%} and \textbf{9.8\%}, respectively.
\end{enumerate}

%% file: sec/2-related_work.tex
\section{Related Works}
\label{sec:related_works}
\subsection{Neural Decoding}
Neural decoding refers to the process of interpreting neural signals (e.g., EEG, MEG, fMRI) to infer state of human perception and cognition. In recent years, significant progress has been made in this field, particularly in applications such as motor imagery decoding~\cite{aflalo2015decoding}, visual decoding~\cite{takagi2023high,scotti2024reconstructing,xia2024dream,lu2023minddiffuser,quan2024psychometry,NEURIPS2019_7d2be41b,scotti2024mindeye2,gaziv2022self,fang2024alleviating,xia2024umbrae,bai2023dreamdiffusion}, text decoding~\cite{duan2024dewave}, emotion decoding~\cite{li2022eeg}, inter-subject decoding\cite{wang2024mindbridge,zhou2024clip}, and diagnosis of neurological disorders~\cite{vicchietti2023computational}. 
Visual decoding includes two primary tasks: brain-to-image retrieval and reconstruction.
Several methods~\cite{du2023decoding, li2024visual, chen2024visual,song2024decoding} have been proposed for visual decoding, notably aligning the encoded representations of EEG/MEG signals with the Contrastive Vision-Language Pre-training (CLIP)~\cite{radford2021learning} embedding space.
However, they did not account for the GAPs between brain signals and visual stimuli, leading to overfitting on the training set and poor generalization to new data.

\subsection{Multi-modal Contrastive Learning}
Contrastive representation learning has attained remarkable achievements in multiple domains, including vision~\cite{chen2020simple}, language~\cite{gao2021simcse}, and graph~\cite{you2020graph}. Building on the success of these works, multi-modal contrastive representation learning (MMCL) has emerged, focusing on aligning inputs from multiple modalities within a shared representation space. These models are typically pretrained on large-scale paired datasets using a contrastive loss function.
Recent vision-language contrastive pre-training models, such as CLIP~\cite{radford2021learning} and ALIGN~\cite{jia2021scaling} have demonstrated remarkable zero-shot retrieval and classification performance, along with robust generalization to a wide range of downstream tasks~\cite{stevens2024bioclip,wang2022medclip}. Inspired by the success of these vision-language models, contrastive representation learning across diverse modalities has garnered increasing attention~\cite{lei2024vit,girdhar2023imagebind,nagrani2022learning}.
However, in real-world settings, for certain modality pairs like audio-visual~\cite{guzhov2022audioclip} and 3D-language~\cite{xue2023ulip}, it is challenging to obtain paired data that match precisely. This constraint restricts the generalization capabilities of the pretraining models.
Fortunately, several methods have been proposed to address this issue and provide theoretical analyses~\cite{xue2023ulip,wang2023connecting,liang2022mind}. 
The visual-neural data for neural decoding also suffers from poor matching. Consequently, rough alignment will inevitably lead to a reduction in generalization performance.

\subsection{Uncertainty Quantification}
Uncertainty quantification is crucial for ensuring quality-aware and high-stakes decision-making~\cite{abdar2021review}.
One prominent field is out-of-distribution (OOD) detection~\cite{lu2023uncertainty,charpentier2020posterior,yang2022openood}.
Extreme outliers present during training can negatively impact generalization performance, while outliers encountered during evaluation can undermine the reliability of the assessment.
Uncertainty quantification has also been applied to multimodal fusion in previous works~\cite{han2022trusted,zhang2023provable,xu2024reliable} to enable dynamic fusion.
Several methods~\cite{zhao2020uncertainty,rizvedefense,wang2021combating} leverage uncertainty to identify the incorrect pseudo-labels in unlabeled data, preventing error accumulation during model training.
\cite{ma2025estimating} enhances response reliability by estimating the uncertainty of LLMs. 
In our task, Random Gap lowers the SNR in brain signals, making it essential to quantify uncertainty and dynamically mitigate the impact.

%% file: sec/3_theory.tex
\section{Visual Neural Decoding}
\subsection{Notation}
In this paper, we begin by introducing the basic notation for visual neural decoding. We use paired data \( (x_v, x_b) \), where \( x_v \in \mathbb{R}^{d_V} \) represents an image from the visual domain, and \( x_b \in \mathbb{R}^{d_B} \) represents the corresponding brain signal. $\mathcal{X}_V $ is used to denote the set of all visual data from distribution $\mathcal{P}_V$, and $\mathcal{X}_B $ is employed to denote the set of all brain data from distribution $\mathcal{P}_B$. Their joint multi-modal distribution is $\mathcal{P}_M$.

\subsection{Vision-Brain Contrastive Learning}
The goal of vision-brain contrastive learning is to map brain data $\mathcal{X}_B$ to a $k$-dimensional latent space $\mathcal{H} \in \mathbb{R}^k$ that aligns with the representation of visual data $\mathcal{X}_V$. This is achieved by using a frozen visual encoder $f_V : \mathcal{X}_V \rightarrow \mathcal{H}$ to obtain visual embeddings and training a brain encoder $f_B : \mathcal{X}_B \rightarrow \mathcal{H}$ with parameters $\theta$ to map brain data into the shared latent space $\mathcal{H}$. 
Given the effectiveness of pretrained vision-language models (VLMs) in providing rich visual features, $f_V$ is taken from the vision branch of a pretrained VLM, such as CLIP~\cite{radford2021learning}.

For multi-modal positive and negative pairs, we define an image-brain pair drawn from the paired vision-brain data, i.e., $(x_v, x_b) \sim \mathcal{P}_M$, as positive pairs, and draw independent samples from each domain, $x_v^{-} \sim \mathcal{P}_V$, $x_b^{-} \sim \mathcal{P}_B$, and treat $(x_v, x_b^{-})$, $(x_v^{-}, x_b)$, and $(x_v^{-}, x_b^{-})$ as negative pairs, assuming that the samples in these pairs are independent of each other.
Given positive and negative pairs $(x_v, x_b, x_v^{-}, x_b^{-})$, the corresponding encoders map them to $(h_v, h_b, h_v^{-}, h_b^{-})$. The learning objective is the symmetric cross-entropy (SCE) loss~\cite{wang2019symmetric}, computed as follows:
{ \small
\begin{multline}
\mathcal{L}_\text{SCE}(f_B) = - \mathbb{E}_{x_v, x_b} \log \frac{\exp \left( f_V(x_v)^\top f_B(x_b)/\tau \right)}{\mathbb{E}_{x_b^{-}} \exp \left( f_V(x_v)^\top f_B(x_b^{-}) /\tau \right)}\\
- \mathbb{E}_{x_v, x_b} \log \frac{\exp \left( f_V(x_v)^\top f_B(x_b)/\tau \right)}{\mathbb{E}_{x_v^{-}} \exp \left( f_V(x_v^{-})^\top f_B(x_b)/\tau \right)}.
\label{eq:compute_loss}
\end{multline}
}

%% file: sec/4_method.tex
\section{Method}
Our method consists of Blur Prior and Uncertainty-aware components, addressing the System GAP and Random GAP, respectively. The algorithmic flow of our framework is illustrated in Algorithm~\ref{alg:ubp} and the details are as follows.

\begin{algorithm}[t]
  \caption{Uncertainty-aware Blur Prior Framework}
  \label{alg:ubp}
  \begin{algorithmic}[1]
    \State \textbf{Input:} Multimodal training dataset $\mathcal{P}_M$
    \State \textbf{Model:} Brain encoder $f_B$ with random parameters $\theta$, pretrained vision encoder $f_V$ with parameters $\phi$, temperature parameter $\tau$, learning rate $\eta$
    \State \textbf{Output:} Trained model $f_B$
    \For{each iteration}
        \State Obtain training sample $(x_v,x_b)$ from dataset $\mathcal{P}_M$ 
        \State \textcolor{darkred}{Obtain $\tilde{x}_v$ by \cref{eq:get_blur_x} with blur radius $r$}
        \State $h_b$ = $f_B$($x_b$); $h_v$ = $f_V$($\tilde{x}_v$)
    
        \State Compute loss $\mathcal{L}$ by \cref{eq:compute_loss}
        \State \textcolor{darkred}{Update $r$ for sample $(x_v,x_b)$ by \cref{eq:get_r}}
        \State Update model parameters $\theta \leftarrow \theta - \eta \nabla \mathcal{L}$
    \EndFor
    \State \textbf{return} trained model $f_B$
  \end{algorithmic}
\end{algorithm}

\subsection{Blur Prior}
Due to the existence of the System GAP between the human visual system and the original visual stimuli, a discrepancy in information arises, particularly in the loss of high-frequency details. Aligning brain signals with the images may cause the model to overfit to the high-frequency details in the images. To mitigate the System GAP, we propose a simple prior, which applies Gaussian blur to the original images, making the image modality better aligned with the brain signal modality.

Based on the characteristics of the experimental paradigm, where the focal point is concentrated on the red dot in the center of the image, we synthesized images of the macular and peripheral regions of the human eye to simulate the decrease in resolution and reduce high-frequency details.
Concretely, a uniformly blurred image is generated first:
\begin{equation}
x_{\text{blur}}(i, j) = \sum_{m=-k}^{k} \sum_{n=-k}^{k} x(i - m, j - n) \cdot G(m, n),
\end{equation}
where $r=2k+1$ denotes the radius of the Gaussian kernel, and $x(i - m, j - n)$ represents the pixel value in the original image $x$, while $ G(m, n) $ denotes the corresponding weights provided by the Gaussian kernel. The Gaussian kernel \( G(m, n) \) is defined as:
\begin{equation}
G(m, n) = \frac{1}{2 \pi \sigma^2} \exp\left(-\frac{m^2 + n^2}{2 \sigma^2}\right),
\end{equation}
where $ \sigma $ is the standard deviation, which controls the intensity of the blur.
The fovea blur image is blended with the original image and the uniformly blurred image as:
\begin{equation}
\tilde{x}_v = \alpha \cdot x + (1 - \alpha) \cdot x_{\text{blur}},
\label{eq:get_blur_x}
\end{equation}
where $\alpha$ is the blending factor, represented as a matrix with values between 0 and 1. For a foveated effect, we define a function of distance from the fovea as:
\begin{equation}
\alpha(i, j) = \exp\left(-\frac{\lambda \cdot d(i, j)}{L}\right),
\end{equation}
where $ d(i, j) $ denotes the Euclidean (2-norm) distance between pixel $ (i, j) $ and the fovea, and $ L $ denotes the maximum possible distance within the image. The parameter $ \lambda$ controls the rate of decay, moderating how quickly the weight $ \alpha(i, j) $ decreases as the distance increases. In our setting, the level of blurriness of the image depends on the radius of the Gaussian kernel $r$, with other factors being fixed.

\subsection{Uncertainty Quantification}
\begin{figure}[t]
  \centering
    \includegraphics[width=1.0\linewidth]{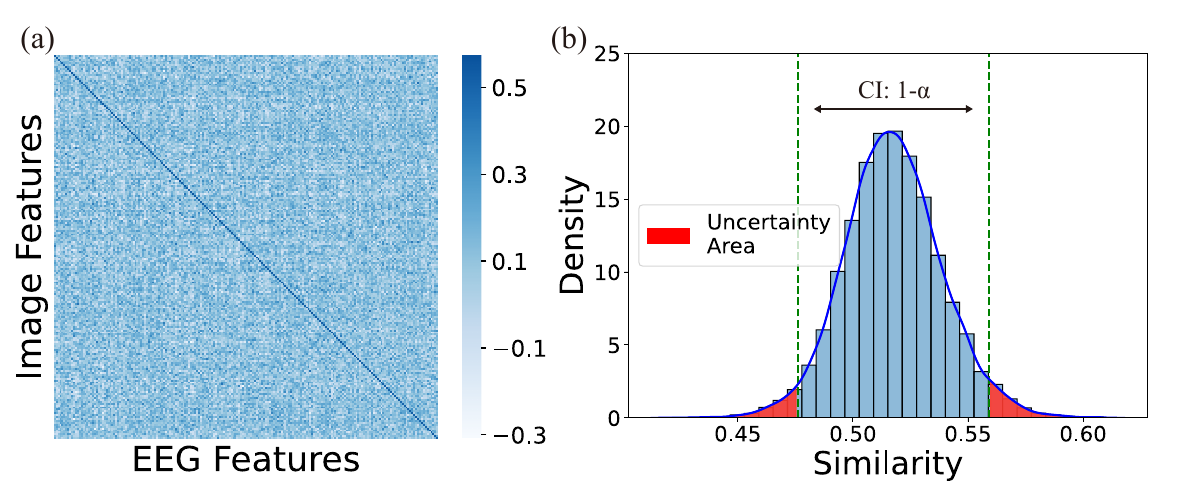}
    \caption{Semantic similarity visualization. (a) Semantic similarity matrix between image features and EEG features. The diagonal represents the similarity between corresponding pairs of features from the two modalities. (b) Density distribution of similarity scores from the diagonal of the matrix. The green dashed lines denote the confidence interval at a significance level of \(1 - \alpha\), indicating the range of similarity scores that are statistically significant. The red areas represent the Uncertainty Area, indicating scores outside the confidence interval.}
    \label{fig:diag_distribution}
\end{figure}

The mismatch between brain signals and the original image stimuli attributes to Random GAP, including dynamics of perception and cognition, along with technical noise, as shown in~\cref{fig:motivation}. Due to the complexity of perception and cognitive processes, which are difficult to disentangle, it is challenging to quantify the contribution of each factor to the Random GAP.
Fortunately, the similarity of paired samples is found to follow a Gaussian distribution, as illustrated in~\cref{fig:diag_distribution}. Based on this observation, outlier pairs falling beyond the confidence interval, indicate a large Random GAP between vision and brain.
For each sample $(x_v, x_b)$, uncertainty is estimated based on its corresponding interval, and 
$r$ is adjusted accordingly to dynamically mitigate the information discrepancy between brain signals and visual stimuli. Overall, a larger discrepancy leads to greater blurring, while a smaller discrepancy results in milder blurring.

Specifically, the \(N\) paired samples are denoted as \(\{(x_v^i, x_b^i)\}_{i=1}^N\). The blurred image is represented as \(x_v \xrightarrow{r} \tilde{x}_v\). The latent features \(h_b \in \mathbb{R}^{N \times d}\) and \(h_v \in \mathbb{R}^{N \times d}\) are obtained through their respective modality encoders. The similarity matrix \(\mathbf{M} \in \mathbb{R}^{N \times N}\) is computed as:

\begin{equation}
\mathbf{M}  = h_b \cdot h_v^\top \cdot \operatorname{softplus}(\tau),
\end{equation}
where $\tau$ is a learned scalar parameter, and $\text{softplus}(\cdot)$ is a smooth, non-linear activation function applied to $\tau$ to ensure positivity. The similarity scores for the \(N\) pairs are represented as \(\mathbf{S} \in \mathbb{R}^{N}\) and computed as:
\begin{equation}
\mathbf{S} = \operatorname{diag}(\mathbf{M}),
\end{equation}
where $\operatorname{diag}(\cdot)$ denotes the diagonal of a matrix. A moving average is applied to \( M \) during iterations to maintain smoothness. The similarity scores approximately follow a normal distribution \(\mathcal{N}(\hat{\mu}, \hat{\sigma}^2)\).  
The mean \(\hat{\mu}\) and variance \(\hat{\sigma}^2\) are computed as follows:
\begin{equation}
\hat{\mu} = \frac{1}{n} \sum_{i=1}^{n} \mathbf{S}_i, \quad
\hat{\sigma}^2 = \frac{1}{n-1} \sum_{i=1}^{n} (\mathbf{S}_i - \hat{\mu})^2.
\end{equation}
The confidence interval for the similarity scores with confidence level $1 - \alpha$ is given by:
\begin{equation}
\left[\hat{\mu}  - z_{\alpha/2} \cdot \hat{\sigma}, \hat{\mu}  + z_{\alpha/2} \cdot \hat{\sigma}\right],
\end{equation}
where $z_{\alpha/2}$ represents the critical value from the standard normal distribution corresponding to a two-sided confidence level of $1 - \alpha$. For the similarity $s$, the corresponding degree of blur is defined as follows:
\begin{equation}
r(s) =
\begin{cases} 
r_0-c, & \text{if}\ s < \hat{\mu} - z_{\alpha/2} \cdot \hat{\sigma}, \\
r_0+c, & \text{if}\ s > \hat{\mu} + z_{\alpha/2} \cdot \hat{\sigma}, \\
r_0, & \text{if}\ \hat{\mu} - z_{\alpha/2} \cdot \hat{\sigma} \leq s \leq \hat{\mu} + z_{\alpha/2} \cdot \hat{\sigma},
\end{cases}
\label{eq:get_r}
\end{equation}
where $r_0$ is the baseline blur radius, and $c$ is a constant that controls the change in blur radius when $s$ is outside this interval.

%% file: sec/5_experiments_results.tex
\input{tabs/eeg}

\section{Experiments and Results}
\label{sec:exp_res}

\subsection{Datasets and Implementation Details}
\textbf{THINGS-EEG}~\citep{gifford2022large} is a large scale EEG dataset including 10 subjects with the Rapid Serial Visual Presentation (RSVP) paradigm~\citep{intraub1981rapid,keysers2001speed,grootswagers2019representational}. The training set includes 1654 concepts with each concept 10 images, and each image repeats 4 times per subject. The test set includes 200 concepts with each concept 1 image, and each image repeats 80 times per subject. 
For data preprocessing, we follow the method detailed in ~\citep{song2024decoding}. Repetitions are averaged for the purpose of high SNR, resulting in a total of 16540 training samples and 200 test samples per subject. The ablation study on channel and time interval selection is provided in Appendix \ref{subs:eeg_feature_selection}. \\
\textbf{THINGS-MEG}~\cite{hebart2023things} involves four participants and consists of 271 channels. It consists of 1854 concepts $\times$ 12 images $\times$ 1 repetition in the training set and 200 concepts $\times$ 1 image $\times$ 12 repetitions in the test set. We follow the same setting described in~\cite{song2024decoding}. Repetitions of the same stimulus are averaged to ensure the SNR.\\
\textbf{Brain Encoders.} We employ a simple yet effective encoder named EEGProject, consisting of two linear layers with residual connection and a normalization layer. The detailed model architecture is provided in the appendix. To further assess the generalizability of our method, we have conducted experiments with additional architectures, including Shallownet~\cite{schirrmeister2017deep}, Deepnet~\cite{schirrmeister2017deep}, EEGnet~\cite{lawhern2018eegnet}, and TSconv~\cite{song2024decoding}.\\
\textbf{Vision Encoders.} Our research employs the visual branches of CLIP models, specifically using pretrained weights from OpenCLIP~\cite{ilharco_gabriel_2021_5143773}. These weights are derived from training multiple models across a diverse range of data sources and computational resources. In the experiments, we utilize several weights, including RN50, RN101, ViT-B/16, ViT-B/32, ViT-L/14, ViT-H/14, ViT-g/14, and ViT-bigG/14. Unless otherwise stated, RN50 is employed as the default model. 

More details on data preprocessing, hyperparameter settings, and hardware configurations are provided in Appendix~\ref{sec:appendix_exp}.

\subsection{Comparison with Baselines}
\input{tabs/meg}

\begin{figure}[t]
  \centering
    \includegraphics[width=0.96\linewidth]{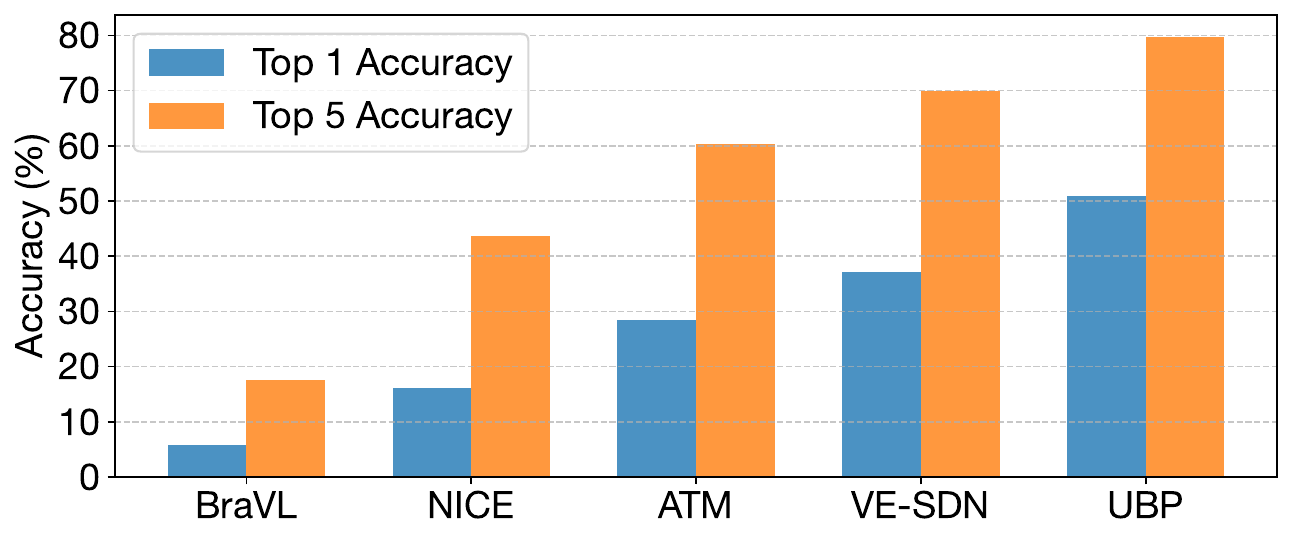}
    \caption{Comparison of Top-1 and Top-5 accuracy (\%) for Intra-subject task on THINGS-EEG.}
    \label{fig:compare_visual}
\end{figure}

\textbf{Baselines.} We compare our approach with recent neural decoding methods. Du et al.~\cite{du2023decoding} propose BraVL, a model based on Mixture of Experts (MoE) that uses multimodal learning of brain-visual-linguistic features. Song et al.~\cite{song2024decoding} present a self-supervised framework for learning image representations from EEG signals, called NICE, incorporating two plug-and-play spatial modules with self-attention and graph attention. Li et al.~\cite{li2024visual} propose a EEG encoder called the Adaptive Thinking Mapper (ATM), which incorporates position encoding and temporospatial encoding. Chen et al.~\cite{chen2024visual} construct a joint semantic space and propose a Visual-EEG Semantic Decouple Framework, called VE-SDN, which explicitly extracts semantic features from both modalities to enable optimal alignment. \\
\textbf{Comparison.} \cref{tab:compare_eeg} and \cref{tab:compare_meg} show quantitative comparisons between our approach and baselines on EEG and MEG test set. Our approach significantly outperforms previous sate-of-the-art in terms of both intra-subject and inter-subject settings. Notably, UBP achieves a top-1 accuracy of 50.9\% and top-5 accuracy of 79.7\% for the zero-shot brain-to-image retrieval task on the THINGS-EEG dataset, and a top-1 accuracy of 26.7\% and top-5 accuracy of 55.2\% on the THINGS-MEG dataset. Notably, we employed two additional evaluation metrics, \textbf{mAP} and \textbf{Similarity Score}, to comprehensively assess performance, as detailed in Appendix~\ref{subs:map}.

\subsection{Effectiveness of Blur Prior}
\input{tabs/data_aug}
\begin{figure}[t]
  \centering
    \includegraphics[width=1.0\linewidth]{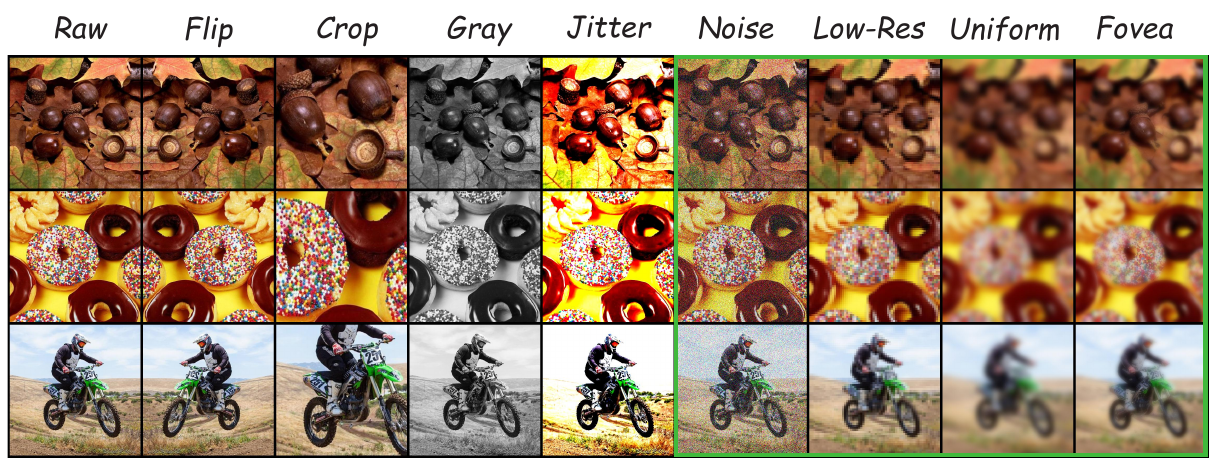}
    \caption{Illustration of various stimuli augmentations and corruptions applied to the visual stimuli. The augmentations (Flip, Crop, Grayscale, Color jitter) modify geometric properties or color distributions, while the corruptions (Gaussian noise, Low resolution, Uniform blur, Fovea blur) degrade image quality by introducing noise, lowering resolution, or simulating optical distortions.}
    \label{fig:data_aug}
\end{figure}
To demonstrate that the Blur Prior is not merely a data augmentation technique but rather a mechanism for bridging the System GAP, visual stimuli processed with different techniques are presented in \cref{fig:data_aug}, with the corresponding performance reported in \cref{tab:data_aug}.
As shown in~\cref{tab:data_aug}, image transformations that degrade high-frequency details significantly enhance retrieval performance, whereas those affecting only geometric properties or color distributions offer limited improvements.
This further supports our motivation that reducing the information mismatch between visual stimuli and brain signals enables the model to mitigate the overfitting issue arising from the System GAP.
Moreover, the proposed Fovea Blur method, drawing inspiration from the human visual system, outperforms other corruptive transformations in terms of performance.
Additionally, when the Random GAP is taken into account, the dynamic blurring method UBP further improves the retrieval performance.



\subsection{Sensitivity Analysis of Various Blur Radius}
\begin{figure}[t]
  \centering
    \includegraphics[width=0.96\linewidth]{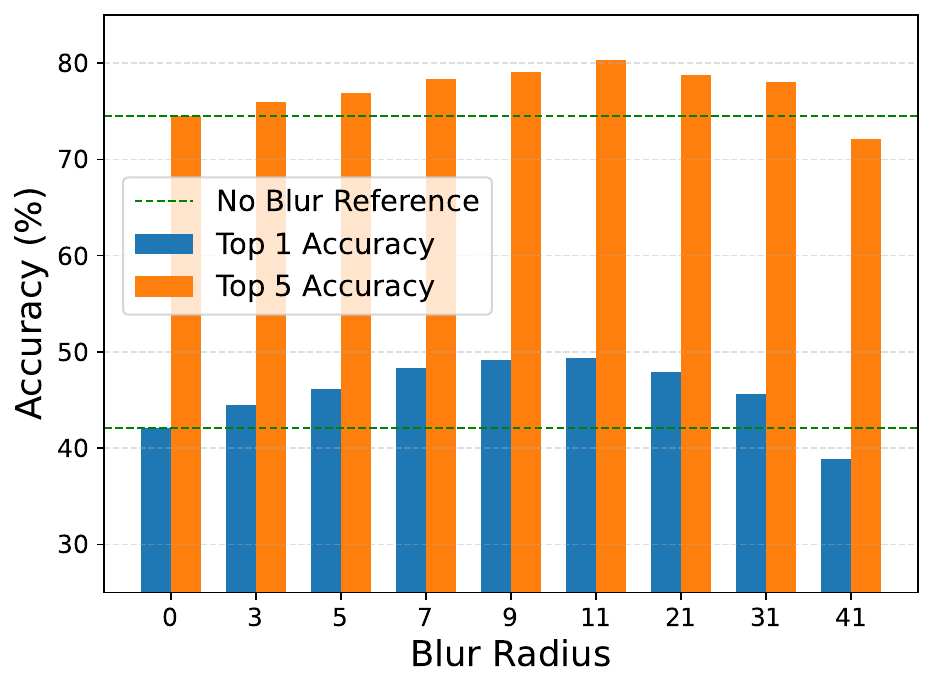}
    \caption{Comparison of Top-1 and Top-5 accuracy (\%) at various blur radius, with reference accuracy for no-blur conditions.}
    \label{fig:different_radius}
\end{figure}

To investigate the effect of different degrees of blur on mitigating System GAP, we applied uniform blur with radius ranging from 0 to 41. The results summarized in~\cref{fig:different_radius} show that as the blur level increases, both top-1 and top-5 accuracy improve, peaking at a blur radius of 11. As the blur level continues to increase, model performance begins to decline, which aligns with our expectation that an appropriate level of blur can reduce the mismatch between visual stimuli and brain signals. Excessive blur, such as a blur radius of 41, leads to a loss of information beyond the optimal level, increasing the information mismatch and resulting in worse performance compared to no blur.

\subsection{Effectiveness of Uncertainty Quantification}
\begin{figure}[t]
  \centering
    \includegraphics[width=1.0\linewidth]{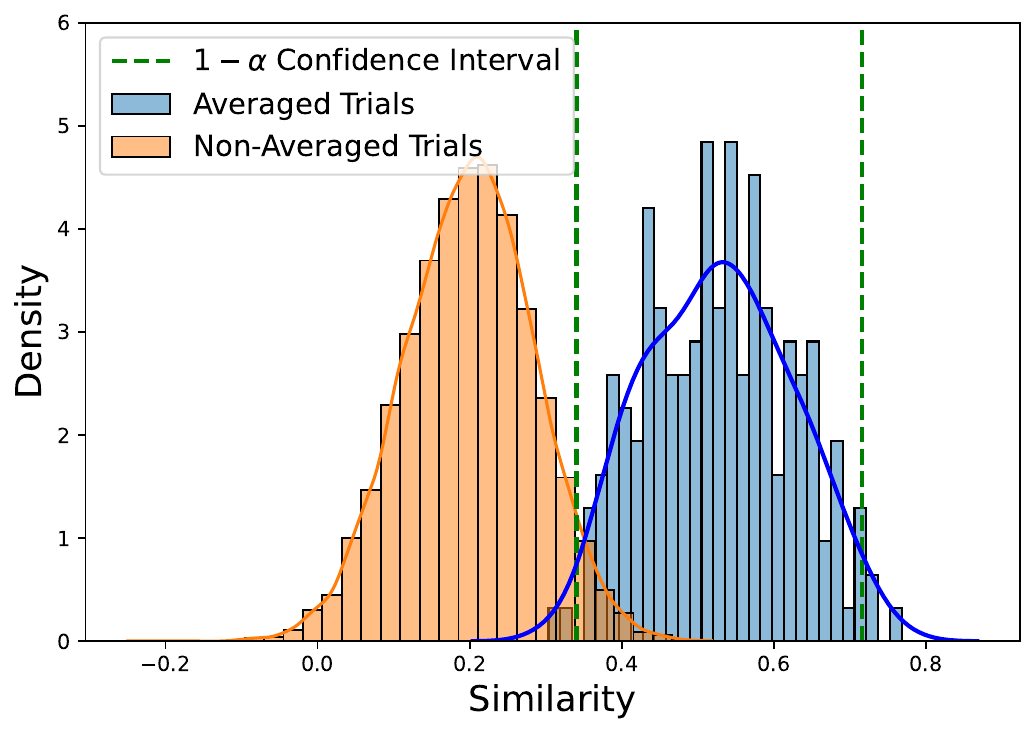}
    \caption{Distribution of similarity scores for averaged and non-averaged EEG trials. The dashed green lines denote the $1 - \alpha$ confidence interval for the averaged trials. Our method effectively distinguishes the two types of trials, with non-averaged samples approximately treated as those with a large Random GAP.}
    \label{fig:sim_no_avg}
\end{figure}
Due to the unavailability of mismatch labels, direct evaluation of uncertainty quantification is challenging. However, non-averaged EEG signals, with their low signal-to-noise ratio, can serve as proxies for outlier samples. As shown in~\cref{fig:sim_no_avg}, our method effectively distinguishes these outlier samples based on the confidence intervals of the similarity distribution, thereby preventing the impact of outlier samples on generalization performance.

\subsection{Ablation Study on Various Encoders}
\begin{figure}[t]
  \centering
    \includegraphics[width=0.95\linewidth]{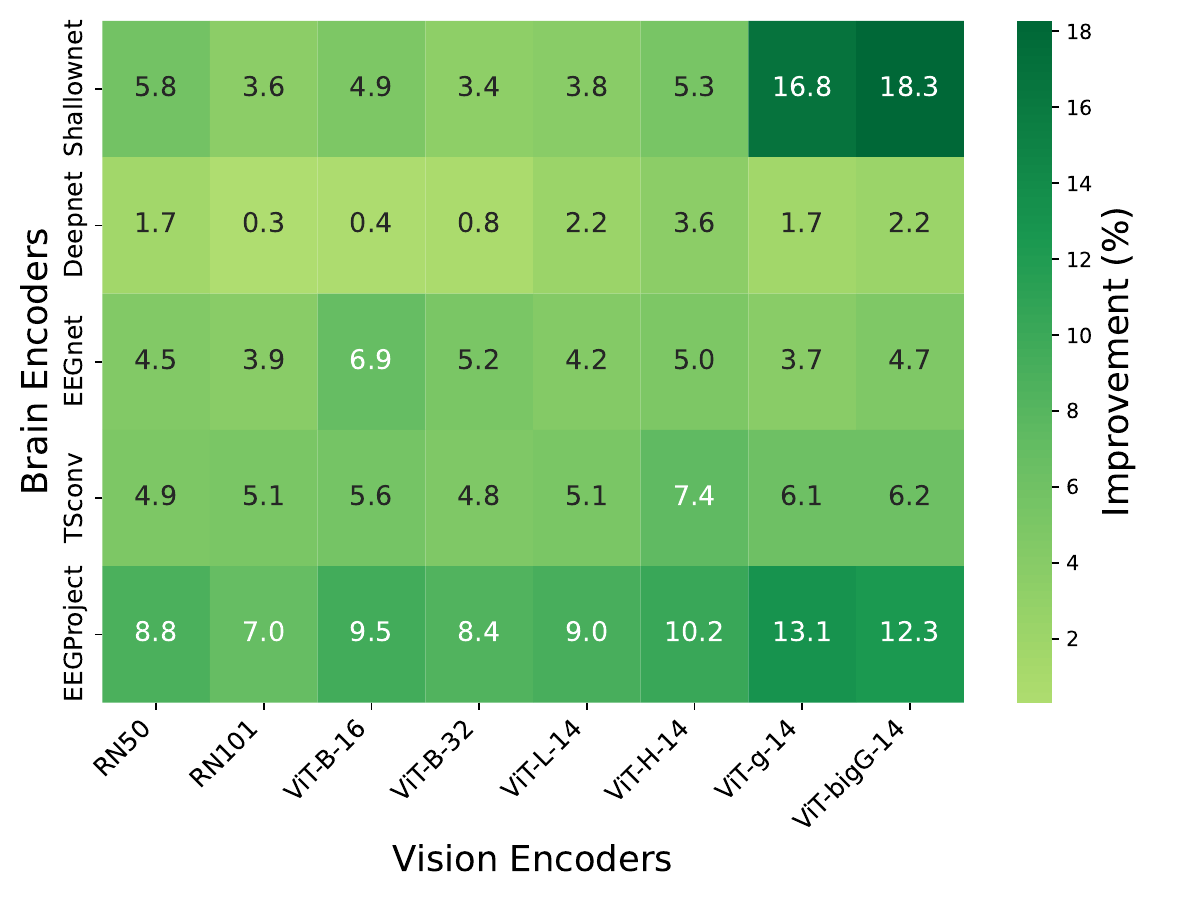}
    \caption{Top-1 accuracy improvement (\%) of UBP across various brain and vision encoder combinations on the THINGS-EEG dataset.}
    \label{fig:backbone_top1_eeg}
\end{figure}


To demonstrate that UBP is not architecture-specific, we conducted comprehensive experiments and trained thousands of models across five brain encoder architectures and eight image encoder architectures, where UBP consistently achieved performance improvements. \cref{fig:backbone_top1_eeg} illustrate the improvements in top-1 accuracy of UBP on the THINGS-EEG. Detailed results, including the top-1 and top-5 accuracy for both baseline and UBP, are provided in the appendix.

\subsection{Robustness to Subject Variability}
As shown in~\cref{tab:correlation}, we report the Pearson and Spearman correlations between subject variability and zero-shot retrieval accuracy. Methods such as NICE-SA, VE-SDN, and Vanilla exhibit strong negative Pearson correlations (e.g., -0.783, -0.687, and -0.761 for top-1, respectively), indicating substantial sensitivity to subject variability.
In contrast, UBP demonstrates improved robustness, with a less negative Pearson correlation (-0.481) compared to Vanilla. UBP demonstrates more stable performance when handling subjects with high variability, with its accuracy not significantly degrading.

\input{tabs/correlation}

%% file: tabs/eeg.tex
\begin{table*}[t]
  \centering
  \caption{Top-1 and Top-5 accuracy (\%) for 200-way zero-shot retrieval on THINGS-EEG}
  \label{tab:compare_eeg}
  \Huge
\resizebox{\linewidth}{!}{
  \begin{tabular}{lcccccccccccccccccccccc}
    \toprule
    & \multicolumn{2}{c}{Subject 1} & \multicolumn{2}{c}{Subject 2} & \multicolumn{2}{c}{Subject 3} & \multicolumn{2}{c}{Subject 4} & \multicolumn{2}{c}{Subject 5} & \multicolumn{2}{c}{Subject 6} & \multicolumn{2}{c}{Subject 7} & \multicolumn{2}{c}{Subject 8} & \multicolumn{2}{c}{Subject 9} & \multicolumn{2}{c}{Subject 10} & \multicolumn{2}{c}{Avg} \\
    \cmidrule(r){2-3} \cmidrule(r){4-5} \cmidrule(r){6-7} \cmidrule(r){8-9} \cmidrule(r){10-11} \cmidrule(r){12-13} \cmidrule(r){14-15} \cmidrule(r){16-17} \cmidrule(r){18-19} \cmidrule(r){20-21} \cmidrule(r){22-23}
    Method & top-1 & top-5 & top-1 & top-5 & top-1 & top-5 & top-1 & top-5 & top-1 & top-5 & top-1 & top-5 & top-1 & top-5 & top-1 & top-5 & top-1 & top-5 & top-1 & top-5 & top-1 & top-5 \\
    \midrule
    \multicolumn{23}{c}{\textbf{Intra-subject}: train and test on one subject} \\
    \midrule
    BraVL~\cite{du2023decoding} & 6.1 & 17.9 & 4.9 & 14.9 & 5.6 & 17.4 & 5.0 & 15.1 & 4.0 & 13.4 & 6.0 & 18.2 & 6.5 & 20.4 & 8.8 & 23.7 & 4.3 & 14.0 & 7.0 & 19.7 & 5.8 & 17.5 \\    
    NICE~\cite{song2024decoding}   &13.2 & 39.5 & 13.5 & 40.3 & 14.5 & 42.7 & 20.6 & 52.7 & 10.1 & 31.5 & 16.5 & 44.0 & 17.0 & 42.1 & 22.9 & 56.1 & 15.4 & 41.6 & 17.4 & 45.8 & 16.1 & 43.6 \\
    NICE-SA~\cite{song2024decoding} & 13.3 & 40.2 & 12.1 & 36.1 & 15.3 & 39.6 & 15.9 & 49.0 & 9.8 & 34.4 & 14.2 & 42.4 & 17.9 & 43.6 & 18.2 & 50.2 & 14.4 & 38.7 & 16.0 & 42.8 & 14.7 & 41.7 \\
    NICE-GA~\cite{song2024decoding} & 15.2 & 40.1 & 13.9 & 40.1 & 14.7 & 42.7 & 17.6 & 48.9 & 9.0 & 29.7 & 16.4 & 44.4 & 14.9 & 43.1 & 20.3 & 52.1 & 14.1 & 39.7 & 19.6 & 46.7 & 15.6 & 42.8 \\
    
    $\text{ATM-S}$~\cite{li2024visual}  &25.6 & 60.4 & 22.0 & 54.5 & 25.0 & 62.4 & 31.4 & 60.9 & 12.9 & 43.0 & 21.3 & 51.1 &30.5 & 61.5 & 38.8 & 72.0 & 34.4 & 51.5 & 29.1 & 63.5 & 28.5 & 60.4 \\
    VE-SDN~\cite{chen2024visual} & 32.6 & 63.7 & 34.4 & 69.9 & 38.7 & 73.5 & 39.8 & 72.0 & 29.4 & 58.6 & 34.5 & 68.8 & 34.5 & 68.3 & 49.3 & 79.8 & 39.0 & 69.6 & 39.8 & 75.3 & 37.2 & 69.9  \\
    \rowcolor{green!20}\textbf{UBP (Ours)}   &41.2 & 70.5 & 51.2 & 80.9 & 51.2 & 82.0 & 51.1 & 76.9 & 42.2 & 72.8 & 57.5 & 83.5 & 49.0 & 79.9 & 58.6 & 85.8 & 45.1 & 76.2 & 61.5 & 88.2 & \textbf{50.9} & \textbf{79.7}\\
    \midrule
    \multicolumn{23}{c}{\textbf{Inter-subject}: leave one subject out for test} \\
    \midrule
    BraVL & 2.3 & 8.0 & 1.5 & 6.3 & 1.4 & 5.9 & 1.7 & 6.7 & 1.5 & 5.6 & 1.8 & 7.2 & 2.1 & 8.1 & 2.2 & 7.6 & 1.6 & 6.4 & 2.3 & 8.5 & 1.8 & 7.0 \\    
    NICE & 7.6 & 22.8 & 5.9 & 20.5 & 6.0 & 22.3 & 6.3 & 20.7 & 4.4 & 18.3 & 5.6 & 22.2 & 5.6 & 19.7 & 6.3 & 22.0 & 5.7 & 17.6 & 8.4 & 28.3 & 6.2 &21.4 \\
    NICE-SA & 7.0 & 22.6 & 6.6 & 23.2 & 7.5 & 23.7 & 5.4 & 21.4 & 6.4 & 22.2 & 7.5 & 22.5 & 3.8 & 19.1 & 8.5 & 24.4 & 7.4 & 22.3 & 9.8 & 29.6 & 7.0 & 23.1 \\
    NICE-GA & 5.9 & 21.4 & 6.4 & 22.7 & 5.5 & 20.1 & 6.1 & 21.0 & 4.7 & 19.5 & 6.2 & 22.5 & 5.9 & 19.1 & 7.3 & 25.3 & 4.8 & 18.3 & 6.2 & 26.3 & 5.9 & 21.6 \\
    $\text{ATM-S}$ &10.5 & 26.8 & 7.1 & 24.8 & \cellcolor{green!10}{11.9} & \cellcolor{green!10}{33.8} & \cellcolor{green!10}{14.7} & \cellcolor{green!10}{39.4} & 7.0 & 23.9 & 11.1 & \cellcolor{green!10}{35.8} & 
\cellcolor{green!10}{16.1} & \cellcolor{green!10}{43.5} & \cellcolor{green!10}{15.0} & \cellcolor{green!10}{40.3} & 4.9 & 22.7 & \cellcolor{green!10}{20.5} & \cellcolor{green!10}{46.5} & 11.8 & \cellcolor{green!10}{33.7} \\
    \rowcolor{green!20}\textbf{UBP (Ours)} &11.5&29.7 & 15.5&40.0 & 9.8&27.0  &13.0&32.3  &8.8&33.8  &11.7&31.0  &10.2&23.8  &12.2&32.2  &15.5&40.5  &16.0&43.5 & \textbf{12.4}&\textbf{33.4} \\

    \bottomrule
  \end{tabular}}
\end{table*}

%% file: tabs/meg.tex
\begin{table}[t]
  \centering
  \caption{Top-1 and Top-5 accuracy (\%) for 200-way zero-shot retrieval on THINGS-MEG}
  \label{tab:compare_meg}
  \huge
  \resizebox{\linewidth}{!}{
  \begin{tabular}{lcccccccccc}
    \toprule
    & \multicolumn{2}{c}{Subject 1} & \multicolumn{2}{c}{Subject 2} & \multicolumn{2}{c}{Subject 3} & \multicolumn{2}{c}{Subject 4}& \multicolumn{2}{c}{Avg} \\
    \cmidrule(r){2-3} \cmidrule(r){4-5} \cmidrule(r){6-7} \cmidrule(r){8-9} \cmidrule(r){10-11}
    Method & top-1 & top-5 & top-1 & top-5 & top-1 & top-5 & top-1 & top-5 & top-1 & top-5 \\
    \midrule
    \multicolumn{11}{c}{\textbf{Intra-subject}: train and test on one subject} \\
    \midrule
    NICE & 9.6 & 27.8 & 18.5 & 47.8 & 14.2 & 41.6 & 9.0 & 26.6 & 12.8 & 36.0\\   
    NICE-SA & 9.8 & 27.8 & 18.6 & 46.4 & 10.5 & 38.4 & 11.7 & 27.2 & 12.7 & 35.0\\ 
    NICE-GA & 8.7 & 30.5 & 21.8 & 56.6 & 16.5 & 49.7 & 10.3 & 32.3 & 14.3 & 42.3\\ 
    \rowcolor{green!20}\textbf{UBP(Ours)} &15.0 & 38.0 & 46.0 & 80.5 & 27.3 & 59.0 & 18.5 & 43.5 & 26.7 & 55.2 \\
    \midrule
    \multicolumn{11}{c}{\textbf{Inter-subject}: leave one subject out for test} \\
    \midrule
     \rowcolor{green!20}\textbf{UBP(Ours)} &2.0 & 5.7 & 1.5 & 17.2 & 2.7 & 10.5 & 2.5 & 8.0 & 2.2 & 10.4\\ 
    \bottomrule
  \end{tabular}}
\end{table}

%% file: tabs/data_aug.tex
\begin{table}[t]
  \centering
  \caption{Top-1 and Top-5 accuracy (\%) for 200-way zero-shot retrieval on THINGS-EEG with different data transformations.}
  \label{tab:data_aug}
  \resizebox{\linewidth}{!}{
  \begin{tabular}{l c c cccc}
    \toprule
     & & &\multicolumn{2}{c}{Intra-subject} &\multicolumn{2}{c}{Inter-subject} \\
    Method &Corrupt &Dynamic & top-1 &top-5  &top-1 &top-5 \\
    \midrule
    
    Vanilla & \xmark &\xmark & 42.1 & 74.5 & 8.5  & 26.6 \\ 
    Flip &\xmark &\xmark & 40.8 & 73.8 & 8.6  & 25.9\\
    Crop &\xmark &\xmark & 41.6 &74.0 & 9.6  & 27.2 \\
    Grayscale &\xmark &\xmark & 38.8 & 72.4 & 9.1  & 27.0 \\
    Color jitter &\xmark &\xmark & 41.3 & 76.2 & 8.5  & 25.7 \\
    \midrule
    \rowcolor{green!20} Noise &\cmark &\xmark & 47.7 & 78.8 & 10.0  & 30.5 \\
    \rowcolor{green!20} Low-Res &\cmark &\xmark &48.1 &  78.4 & 10.8  & 31.9 \\
    \rowcolor{green!20} Uniform blur &\cmark &\xmark & 49.3 & 80.3 & 11.2  & 31.1 \\  
    \rowcolor{green!20} Fovea blur &\cmark &\xmark & 50.2 & 79.1 & 12.3 & 31.7 \\  
    \rowcolor{green!20} UBP &\cmark &\cmark & 50.9 & 79.7 & 12.4 & 33.4 \\  
    \bottomrule
  \end{tabular}}
\end{table}

%% file: tabs/correlation.tex
\begin{table}[t]
\centering
\caption{Pearson and Spearman correlation coefficients between each subject’s mean variability value and the corresponding Top-1 accuracy for different methods.}
\label{tab:correlation}
\small
\begin{tabular}{lcccc}
\toprule
& \multicolumn{2}{c}{\bf Pearson} &\multicolumn{2}{c}{\bf Spearmanr} \\
Method &top-1 &top-5  &top-1 &top-5 \\
\midrule
BraVL &-0.419 &-0.451 & -0.394 &-0.406  \\
NICE    &-0.681 &-0.705& -0.564 &-0.588 \\
NICE-SA    &-0.783 &-0.539& -0.745 &-0.418\\
NICE-GA   &-0.611 &-0.709& -0.382 &-0.450\\
ATM-S &-0.643 &-0.608 & -0.624 &-0.697\\
VE-SDN &-0.687 &-0.810 & -0.787 &-0.758\\
\midrule
Vanilla &-0.761 &-0.721 & -0.636 &-0.690\\
UBP &-0.481 &-0.649 & -0.345 &-0.515\\
\rowcolor{green!20}$\uparrow$ Improvement & 0.280 & 0.072 & 0.291 & 0.175 \\
\bottomrule
\end{tabular}
\end{table}

%% file: sec/6_conclusion.tex
\section{Conclusion}
\label{sec:conclusion}
In this work, we propose the Uncertainty-aware Blur Prior (UBP) to address the System GAP and Random GAP in visual neural decoding. 
UBP leverages uncertainty estimation and biological priors to robustly retrieve natural images from multiple brain modalities.
Extensive experiments demonstrate that UBP outperforms previous state-of-the-art methods, achieving 13.7\% improvement in Top-1 accuracy and 9.8\% improvement in Top-5 accuracy on the THINGS-EEG dataset, along with 12.4\% improvement in Top-1 accuracy and 12.9\% improvement in Top-5 accuracy on the THINGS-MEG dataset. Beyond brain-to-image retrieval, UBP holds potential for applications in stimuli reconstruction and broader multimodal learning contexts. To the best of our knowledge, this is the first effort to incorporate uncertainty awareness and priors into visual neural decoding, offering new perspectives for brain-computer interfaces. Moreover, UBP provides valuable insights for other multimodal tasks, where similar challenges may arise.\\
\textbf{Limitations.} Despite its effectiveness in reducing mismatches between brain signals and visual stimuli, UBP has certain limitations. 
UBP uses a blur prior to approximate high-frequency detail loss, providing a simplified model of the visual system but lacking completeness.
Advanced learnable methods could better bridge this and improve generalization.
Additionally, uncertainty quantification may fail due to the complexity of the Random GAP, which is influenced by perceptual and cognitive dynamics, as well as technical noise. It is promising to investigate advanced uncertainty quantification methods to improve reliability and robustness. \\
\textbf{Acknowledgements.} This work is partially supported by the National Key R\&D Program of China (2022YFF1202400) and the National Natural Science Foundation of China (62376193). The authors appreciate the valuable feedback from anonymous reviewers.

%% file: sec/X_suppl.tex
\clearpage
\appendix

\setcounter{page}{1}
\maketitlesupplementary
\appendix


\section{Experimental details}
\label{sec:appendix_exp}
\subsection{Datasets details}
\textbf{THINGS-EEG}~\citep{gifford2022large} is a large scale EEG dataset included 10 subjects with the Rapid Serial Visual Presentation (RSVP) paradigm~\citep{intraub1981rapid,keysers2001speed,grootswagers2019representational}. The EEG data are collected using 64-channel EASYCAP equipment with the standard 10-10 system~\citep{nuwer1998ifcn}. The training set includes 1654 concepts with each concept 10 images, and each image repeats 4 times (1654 concepts \(\times\) 10 images/concept \(\times\) 4 trials/image) per subject. The test set includes 200 concepts with each concept 1 image, and each image repeats 80 times (200 concepts \(\times\) 1 image/concept \(\times\) 80 trials/image) per subject. 

For data preprocessing, we follow the method detailed in ~\citep{song2024decoding}. Raw EEG data filtered to [0.1, 100] Hz has 63 channels and a sample rate of 1000 Hz. EEG data is epoched into trials ranging from 0 to 1000 ms after stimuli onset with baseline correction using the prior 200 ms average. EEG data is down-sampled to 250 Hz and 17 channels are selected overlying occipital and parietal cortex related to visual\footnote{P7, P5, P3, P1, Pz, P2, P4, P6, P8, PO7, PO3, POz, PO4, PO8, O1, Oz, O2}. For the purpose of high Signal-to-Noise Ratio (SNR), EEG repetitions are
averaged, resulting in total of 16540 training samples and 200 test samples per subject. Additionally, we store EEG data in float16 format to enable faster reading speeds and reduce storage requirements.\\
\\
\textbf{THINGS-MEG}~\cite{hebart2023things} dataset involves four participants and is characterized by 271 channels. The experimental design incorporates a relatively long stimulus duration of 500 ms, followed by a blank screen with a duration of 1000 ± 200 ms. It consists of 1854 concepts$\times$12 images$\times$1 repetitions in the training stage and 200 concepts$\times$1 image$\times$12 repetitions in the test stage. 

We follow the settings described in~\cite{song2024decoding}. During the data processing phase, 200 test concepts are discarded from the training set to construct the zero-shot task, mirroring the procedures in that study. Subsequently, the MEG data are epoched into trials covering the period from 0 to 1000 ms after the stimuli onset. For preprocessing, a band-pass filter within the range of [0.1, 100] Hz is utilized, and baseline correction is carried out after down-sampling the data to 200 Hz. Additionally, we average all MEG repetitions of one image to ensure the signal-to-noise ratio. Additionally, we store EEG data in float16 format to enable faster reading speeds and reduce storage requirements.\\

\subsection{Implementation details}
\textbf{Environment.} Our method is implemented with Python 3.8.19, CUDA 12.0, and PyTorch 2.4.1. The required libraries are specified in the \texttt{requirements.txt} file provided in the repository. The experiments are performed on a machine equipped with an Intel Xeon Platinum 8352V CPU, four V100 GPUs, and 256 GB of RAM.\\
\textbf{Training Configuration.} We use a batch size of 1024 and train the model for 50 epochs. The learning rate is set to 1e-4 for intra-subject setting and 1e-5 for inter-subject setting. Gradient updates are performed using the AdamW\cite{loshchilov2018decoupled} optimizer with weight decay set to 1e-4. Early stopping is employed to monitor training loss and validation performance, concluding the training process to mitigate overfitting when improvements stabilize. Notably, we use the softplus function instead of the exponential function to ensure that temperature parameter $\tau$ remains positive and continuous, as softplus offers a smoother and more stable transition, avoiding the numerical instability of the exponential function. For all above experiments, the hyperparameter $r_0$ is set to 0.25 and c is set to 10. \\
\textbf{Architectures.} We use EEGProject as the brain encoder, detailed as follows:
\begin{mdframed}[
    linewidth=1pt,
    linecolor=black,
    backgroundcolor=white,
    roundcorner=5pt
]
\begin{verbatim}
 Input: Brain Data
        ↓
 Linear Layer (Input_dim → Proj_dim)
        ↓
 ResidualAdd
  |   +---------------------+
  |   | GELU               |
  |   | Linear Layer       |
  |   | Dropout (Rate: 0.3)|
  |   +---------------------+
  +---- ↓
 LayerNorm
        ↓
 Output: Latent
\end{verbatim}
\end{mdframed}

We provided the number of parameters and embedding dimension within different EEG encoders~\cite{schirrmeister2017deep,lawhern2018eegnet,song2024decoding}, in~\cref{tab:brain_encoders}. Compared to other models, EEGProject achieves its performance through a simple yet effective architecture, while remaining lightweight with 5.154M parameters, especially in comparison to the vision branch. 

We also provide the parameter counts for various CLIP vision branch models~\cite{ilharco_gabriel_2021_5143773} to offer a comprehensive comparison across architectures in~\cref{tab:vision_encoders}.

\clearpage
\onecolumn

\section{Results details}
\label{sec:appendix_res}

\subsection{Retrieval Case Analysis}
\begin{figure*}[h!]
  \centering
  \includegraphics[width=0.81\linewidth]{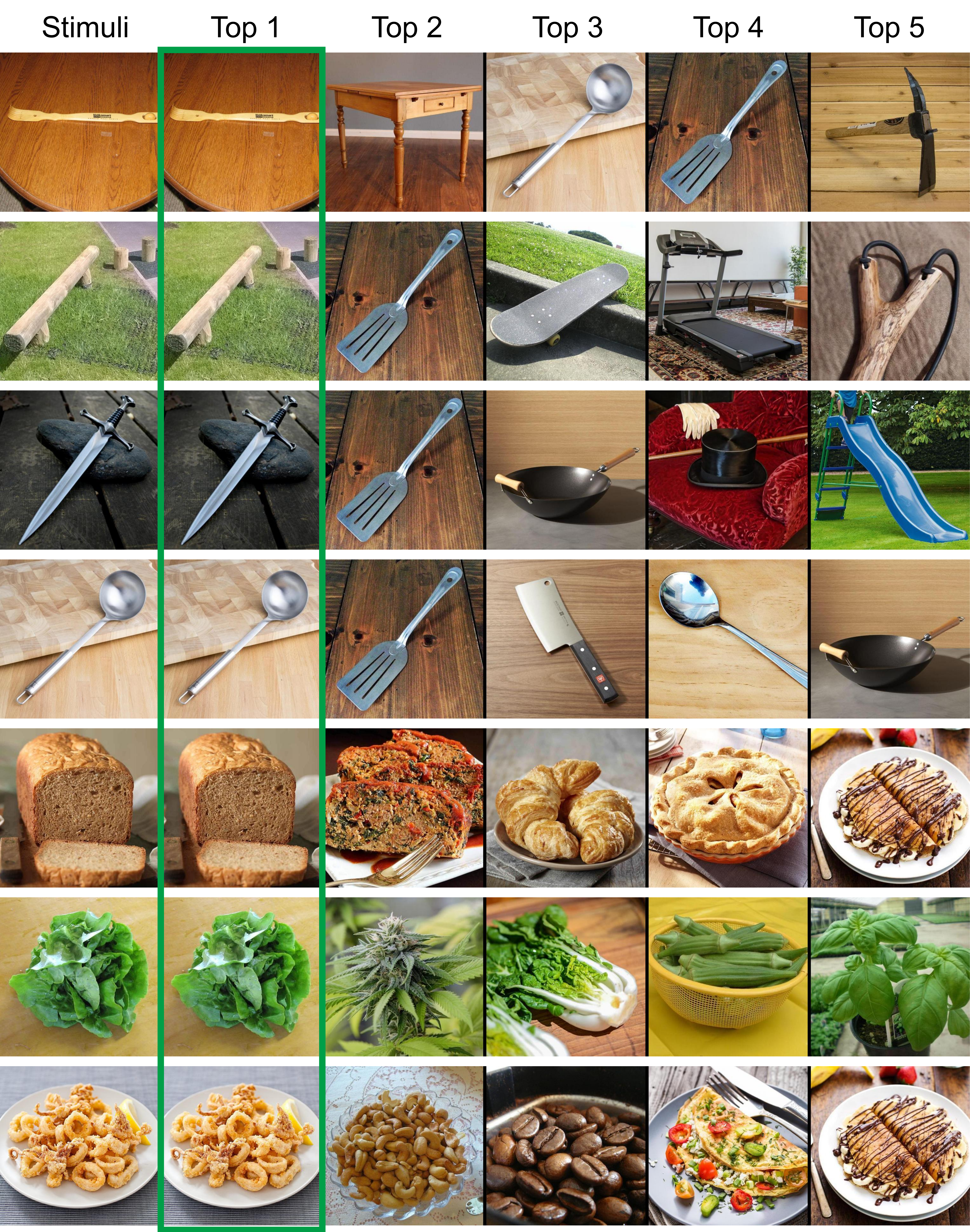}
  \caption{Good Cases: Top-5 Retrieval Results for Various Stimuli.}
  \label{fig:good_cases}
\end{figure*}

\begin{figure*}[h]
  \centering
  \includegraphics[width=0.81\linewidth]{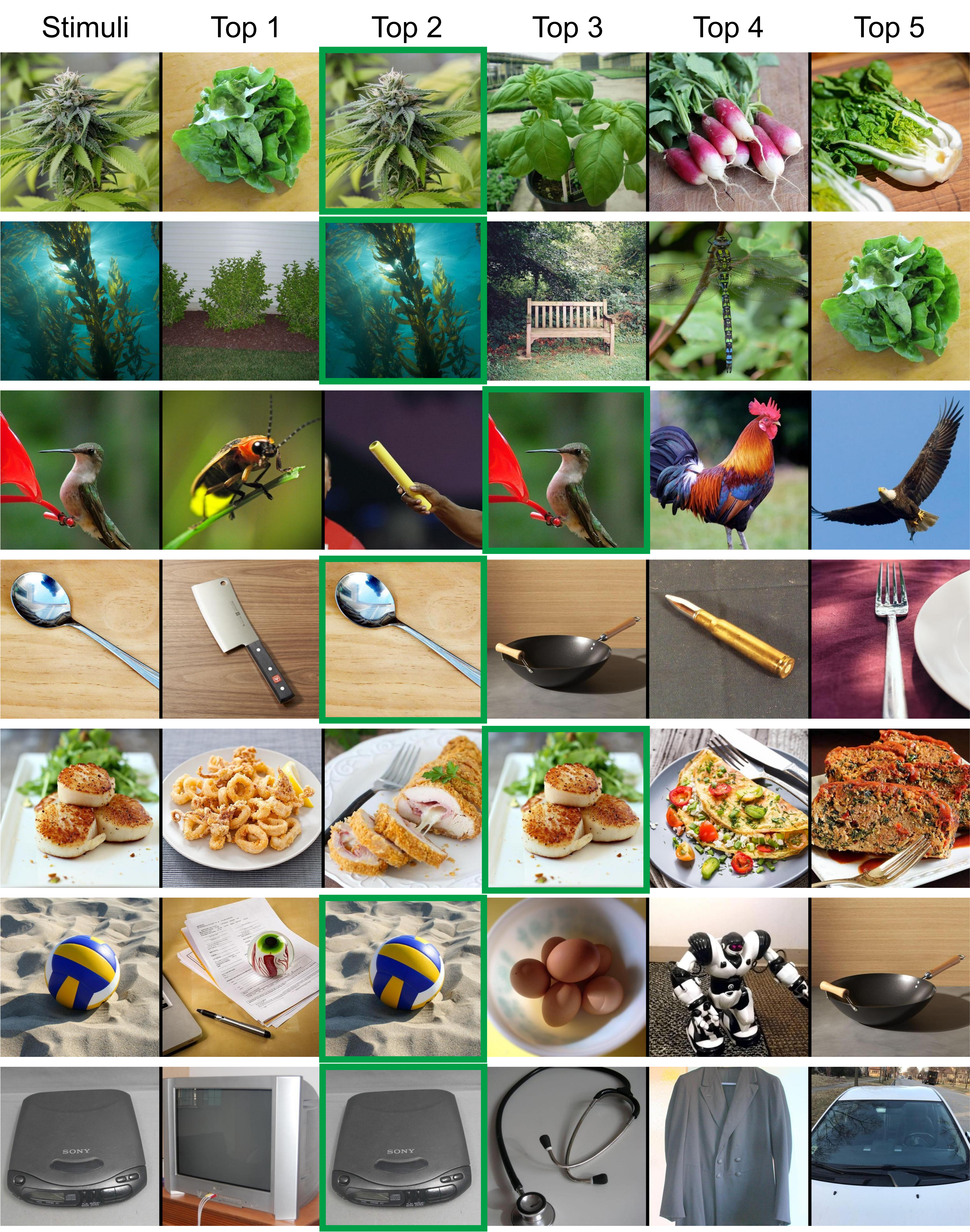}
  \caption{Bad Cases: Top-5 Retrieval Results for Various Stimuli.}
  \label{fig:bad_cases}
\end{figure*}

We present the top-5 retrieval results on THINGS-EEG dataset, including both good cases and bad cases, as shown in~\cref{fig:good_cases} and ~\cref{fig:bad_cases}, respectively. 
Good cases demonstrate the model's capability to effectively align with the target stimuli and retrieve relevant results. An intriguing retrieval result is that the model not only retrieves items with similar materials but also demonstrates \textbf{associations with the orientation and quantity of objects}. These observations warrant further investigation in future studies.

 In contrast, bad cases reveal limitations in distinguishing fine-grained features or addressing semantic inconsistencies. It is challenging to distinguish highly similar stimuli due to the limited information contained in brain signals.

\clearpage
\subsection{THINGS-EEG Results}
\begin{figure*}[!h]
  \centering
  \begin{subfigure}[h]{0.45\linewidth}
    \centering
    \includegraphics[width=\linewidth]{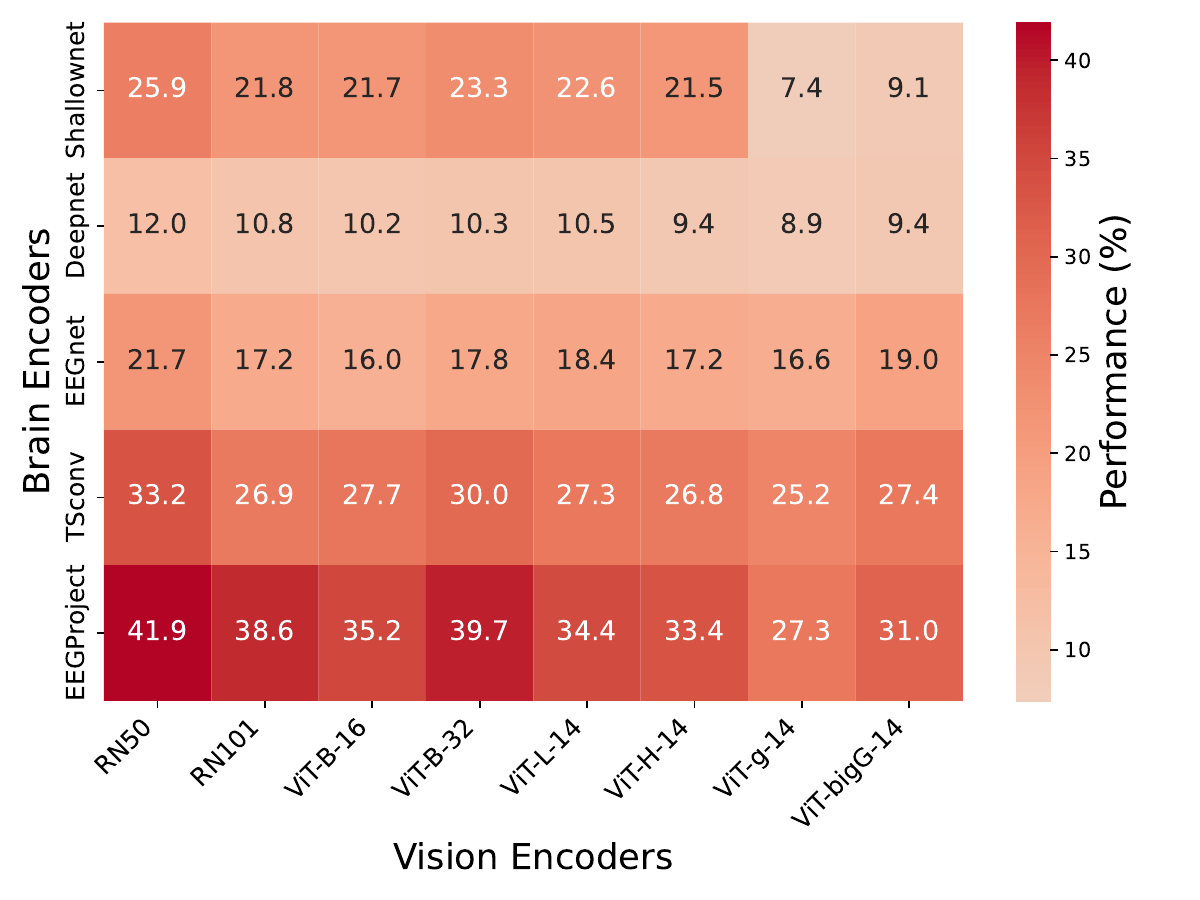}
    \caption{\textbf{Top-1} accuracy (\%) of \textbf{Vanilla} on the THINGS-EEG dataset.}
  \end{subfigure}
  \hfill
  \begin{subfigure}[h]{0.45\linewidth}
    \centering
    \includegraphics[width=\linewidth]{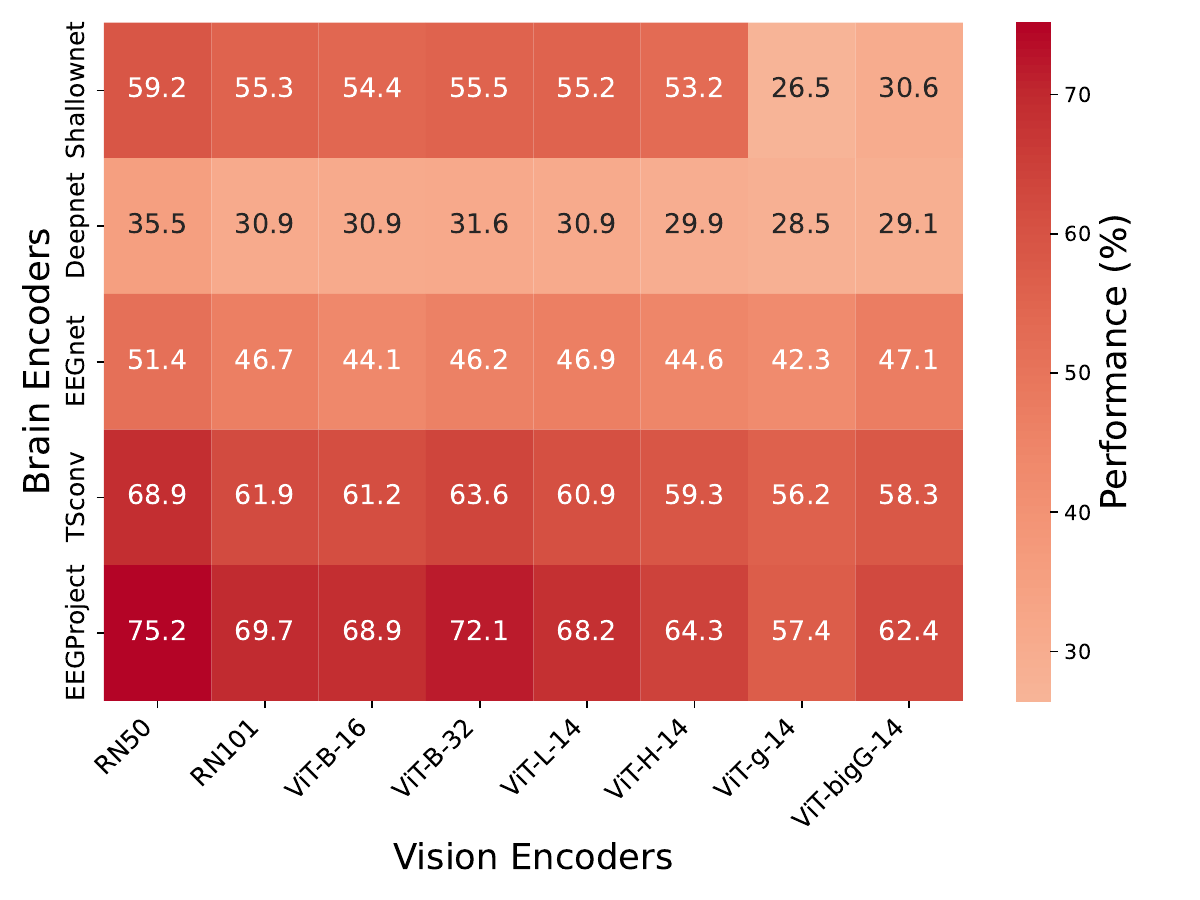}
    \caption{\textbf{Top-5} accuracy (\%) of \textbf{Vanilla} on the THINGS-EEG dataset.}
  \end{subfigure}

  \vspace{0.3cm}
  \begin{subfigure}[h]{0.45\linewidth}
    \centering
    \includegraphics[width=\linewidth]{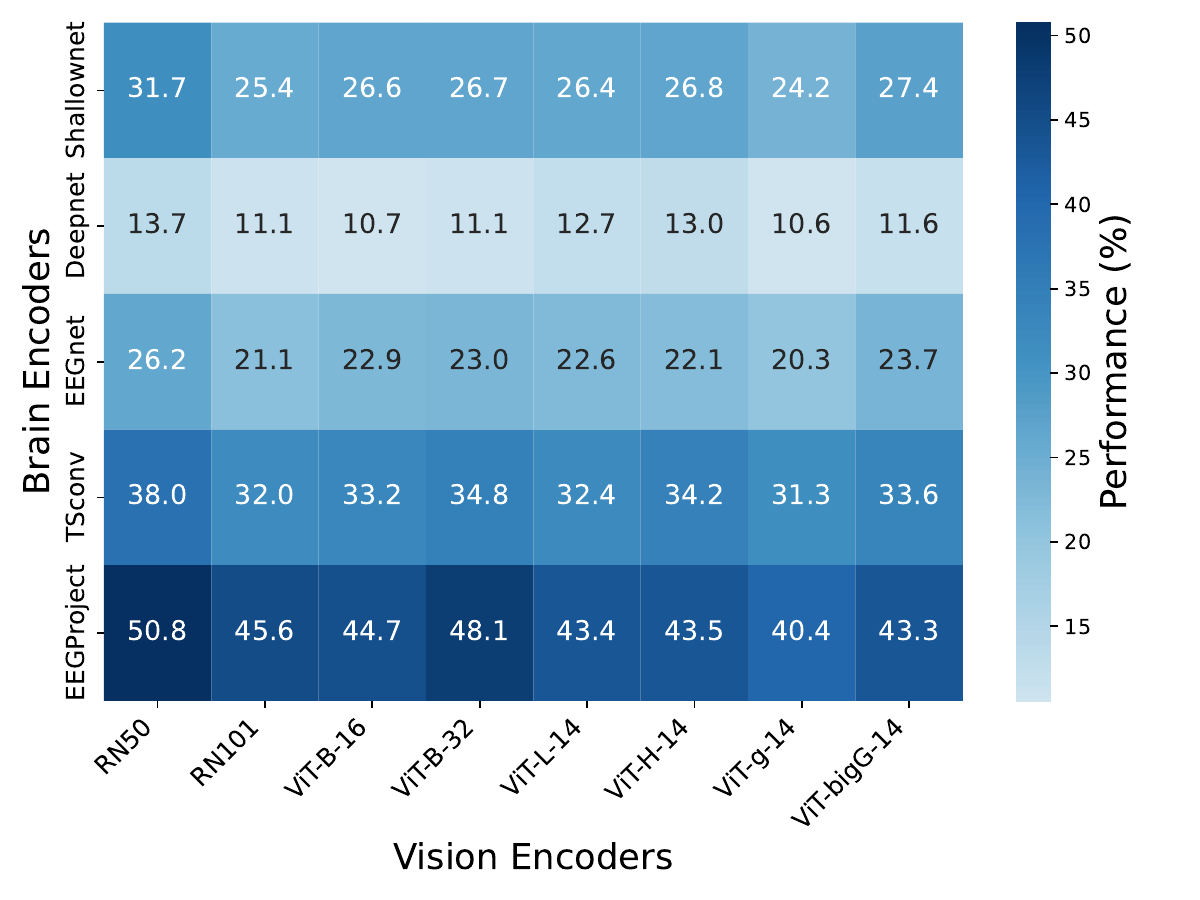}
    \caption{\textbf{Top-1} accuracy (\%) of \textbf{UBP} on the THINGS-EEG dataset.}
  \end{subfigure}
  \hfill
  \begin{subfigure}[h]{0.45\linewidth}
    \centering
    \includegraphics[width=\linewidth]{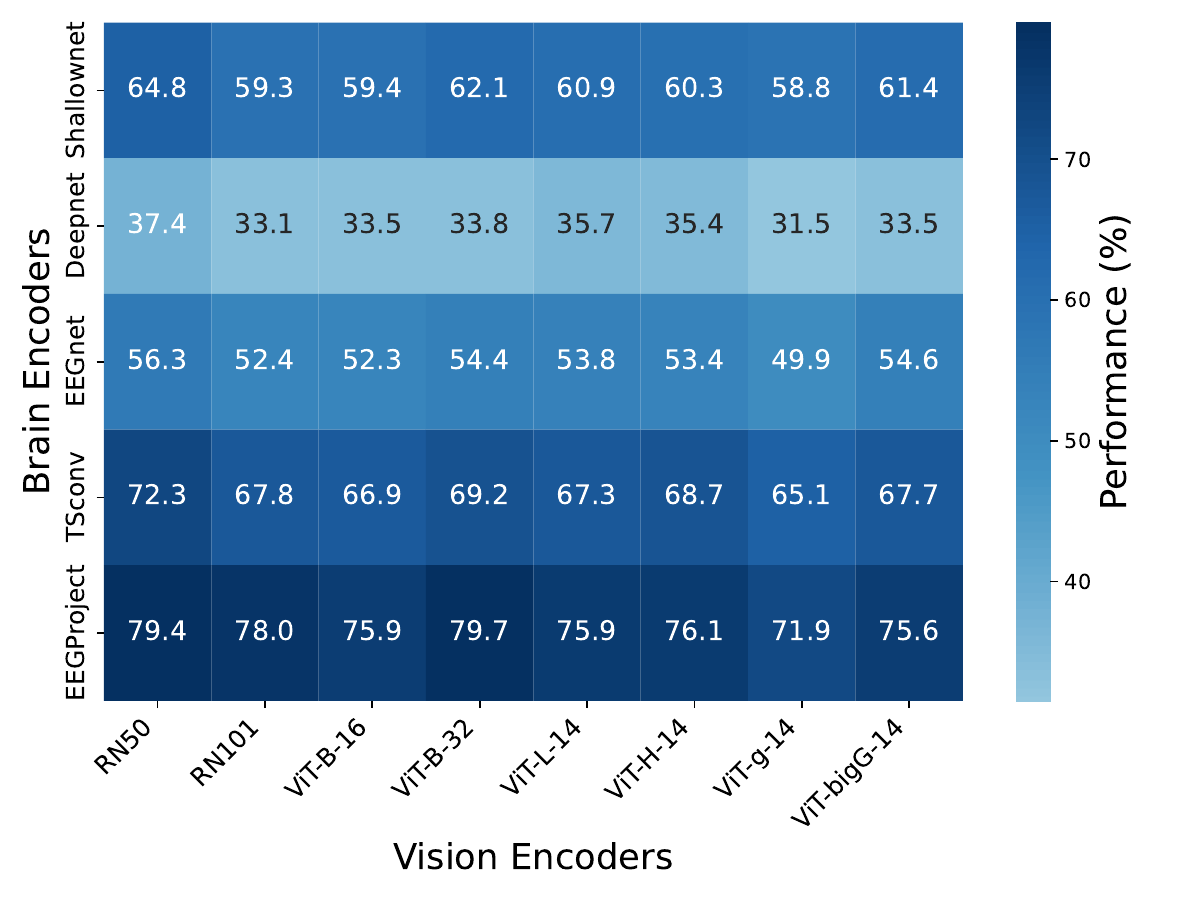}
    \caption{\textbf{Top-5} accuracy (\%) of \textbf{UBP} on the THINGS-EEG dataset.}
  \end{subfigure}

  \vspace{0.3cm}
  \begin{subfigure}[h]{0.45\linewidth}
    \centering
    \includegraphics[width=\linewidth]{figs/backbone_improvement_top1_eeg.pdf}
    \caption{\textbf{Top-1} accuracy \textbf{improvement} (\%) on the THINGS-EEG dataset.}
  \end{subfigure}
  \hfill
  \begin{subfigure}[h]{0.45\linewidth}
    \centering
    \includegraphics[width=\linewidth]{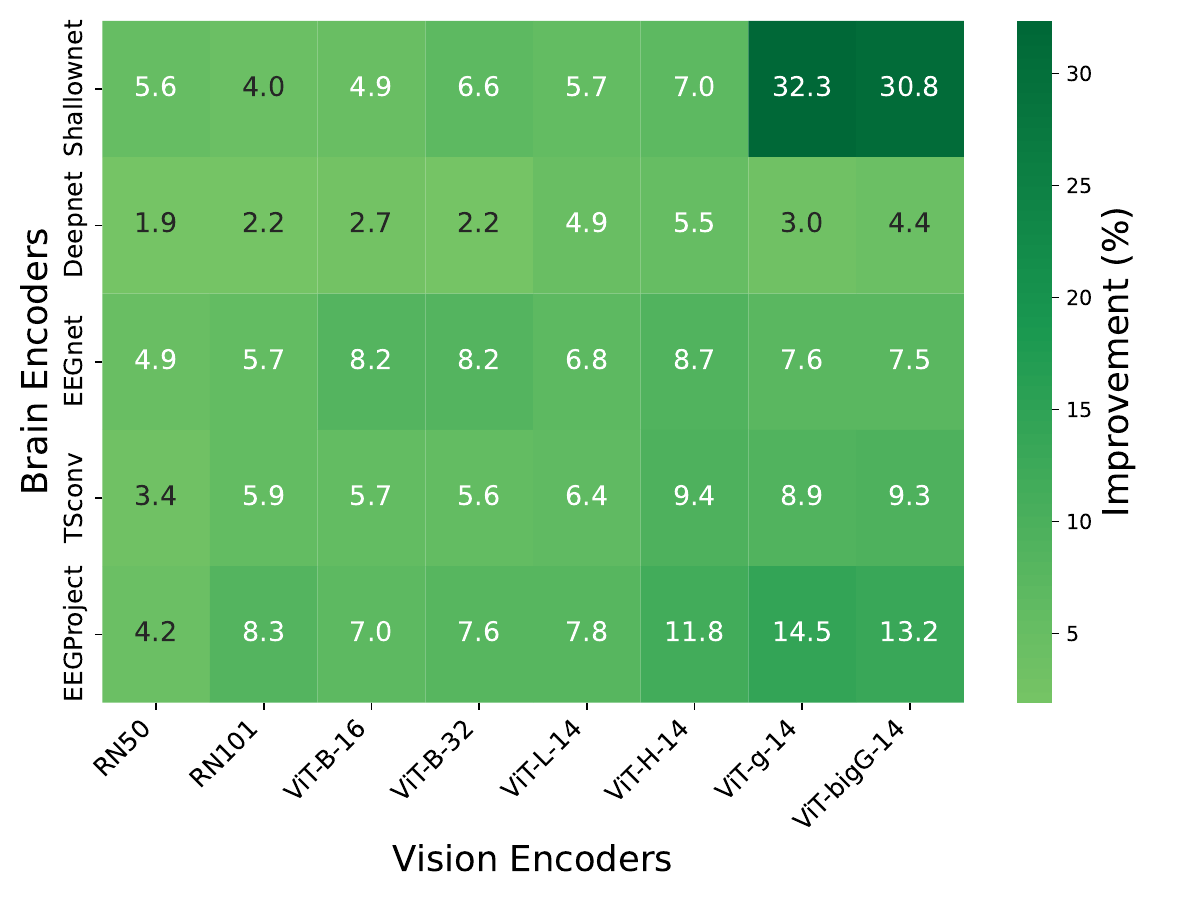}
    \caption{\textbf{Top-5} accuracy \textbf{improvement} (\%) on the THINGS-EEG dataset.}
  \end{subfigure}
  
  \caption{Results on the THINGS-EEG dataset.}
\end{figure*}






\input{tabs/appendix_eeg_backbone}

\clearpage
\subsection{THINGS-MEG Results}

\begin{figure*}[!h]
  \centering
  \begin{subfigure}[h]{0.45\linewidth}
    \centering
    \includegraphics[width=\linewidth]{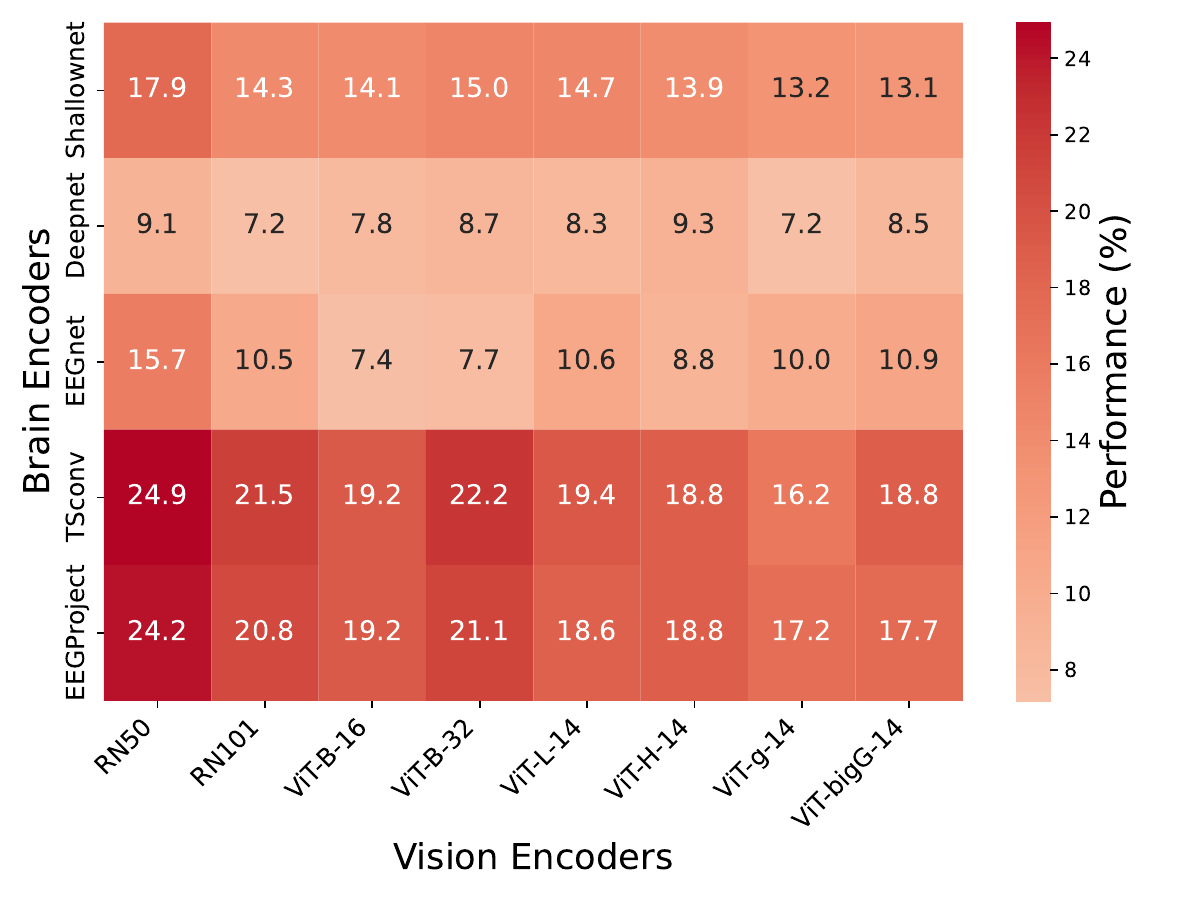}
    \caption{\textbf{Top-1} accuracy (\%) of \textbf{Vanilla} on the THINGS-MEG dataset.}
  \end{subfigure}
  \hfill
  \begin{subfigure}[h]{0.45\linewidth}
    \centering
    \includegraphics[width=\linewidth]{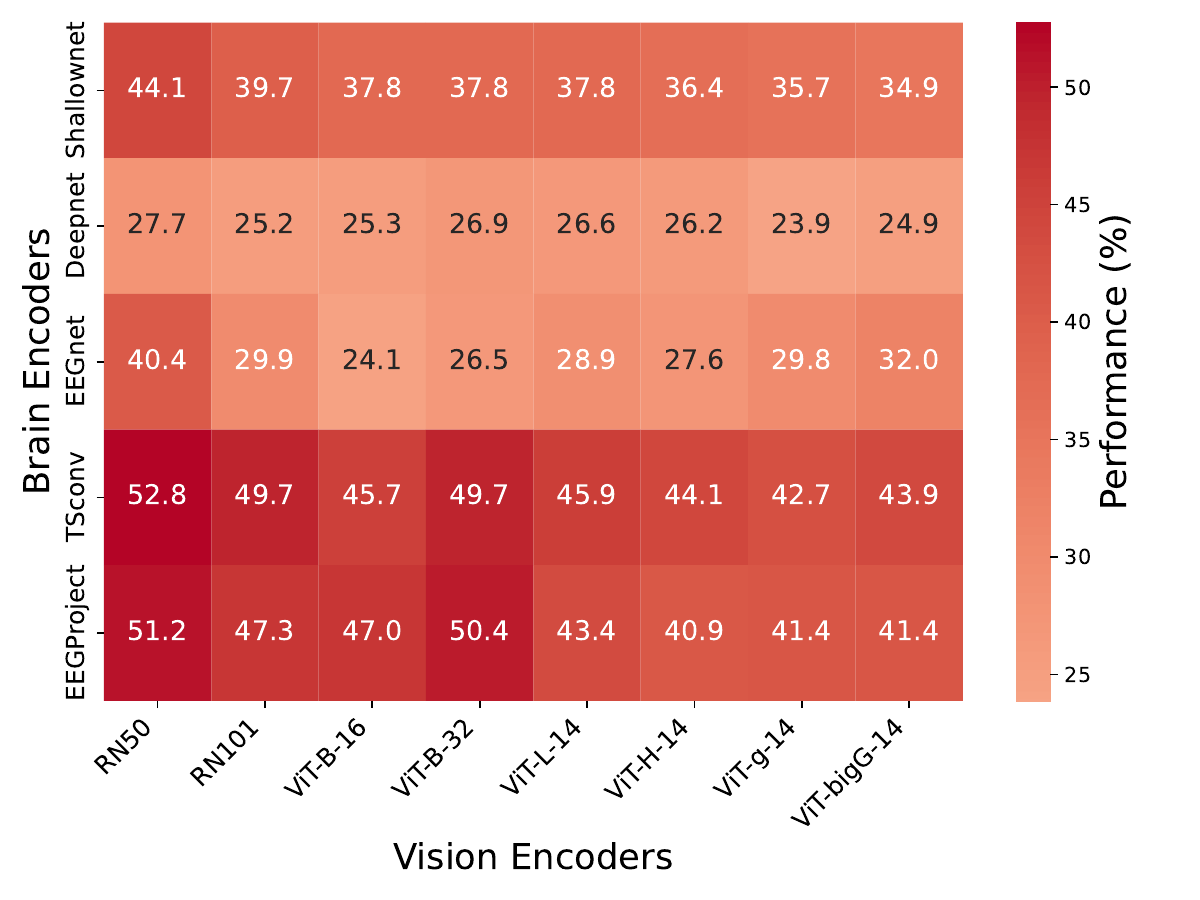}
    \caption{\textbf{Top-5} accuracy (\%) of \textbf{Vanilla} on the THINGS-MEG dataset.}
  \end{subfigure}

  \vspace{0.3cm}
  \begin{subfigure}[h]{0.45\linewidth}
    \centering
    \includegraphics[width=\linewidth]{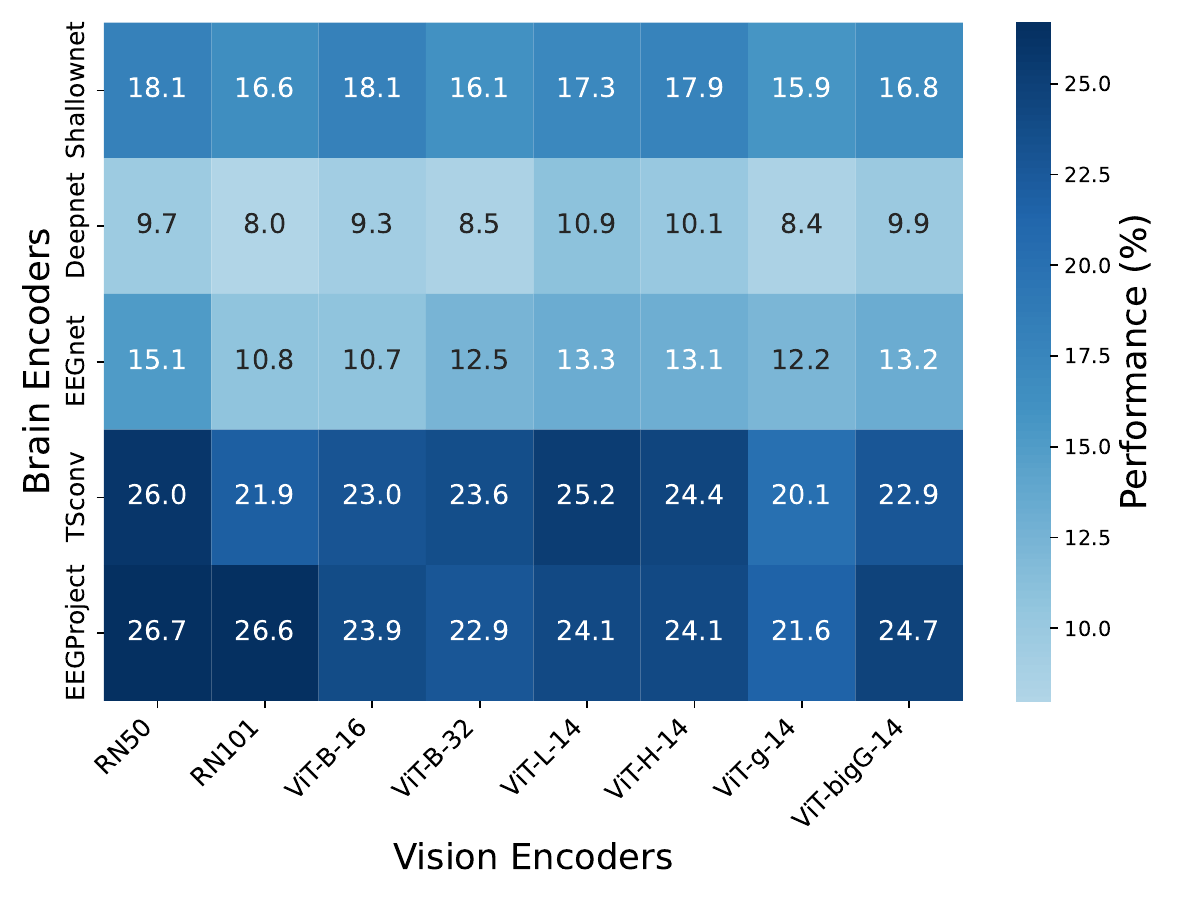}
    \caption{\textbf{Top-1} accuracy (\%) of \textbf{UBP} on the THINGS-MEG dataset.}
  \end{subfigure}
  \hfill
  \begin{subfigure}[h]{0.45\linewidth}
    \centering
    \includegraphics[width=\linewidth]{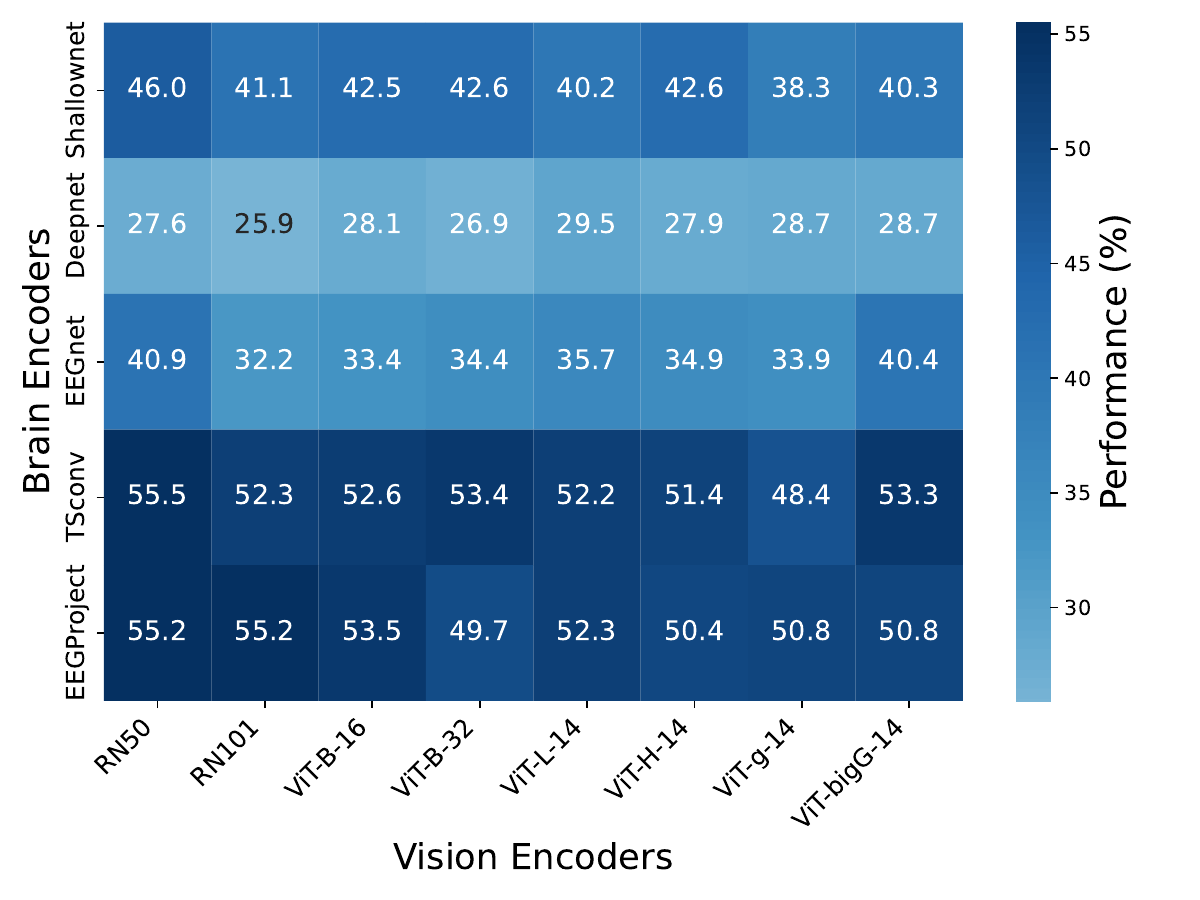}
    \caption{\textbf{Top-5} accuracy (\%) of \textbf{UBP} on the THINGS-MEG dataset.}
  \end{subfigure}

  \vspace{0.3cm}
  \begin{subfigure}[h]{0.45\linewidth}
    \centering
    \includegraphics[width=\linewidth]{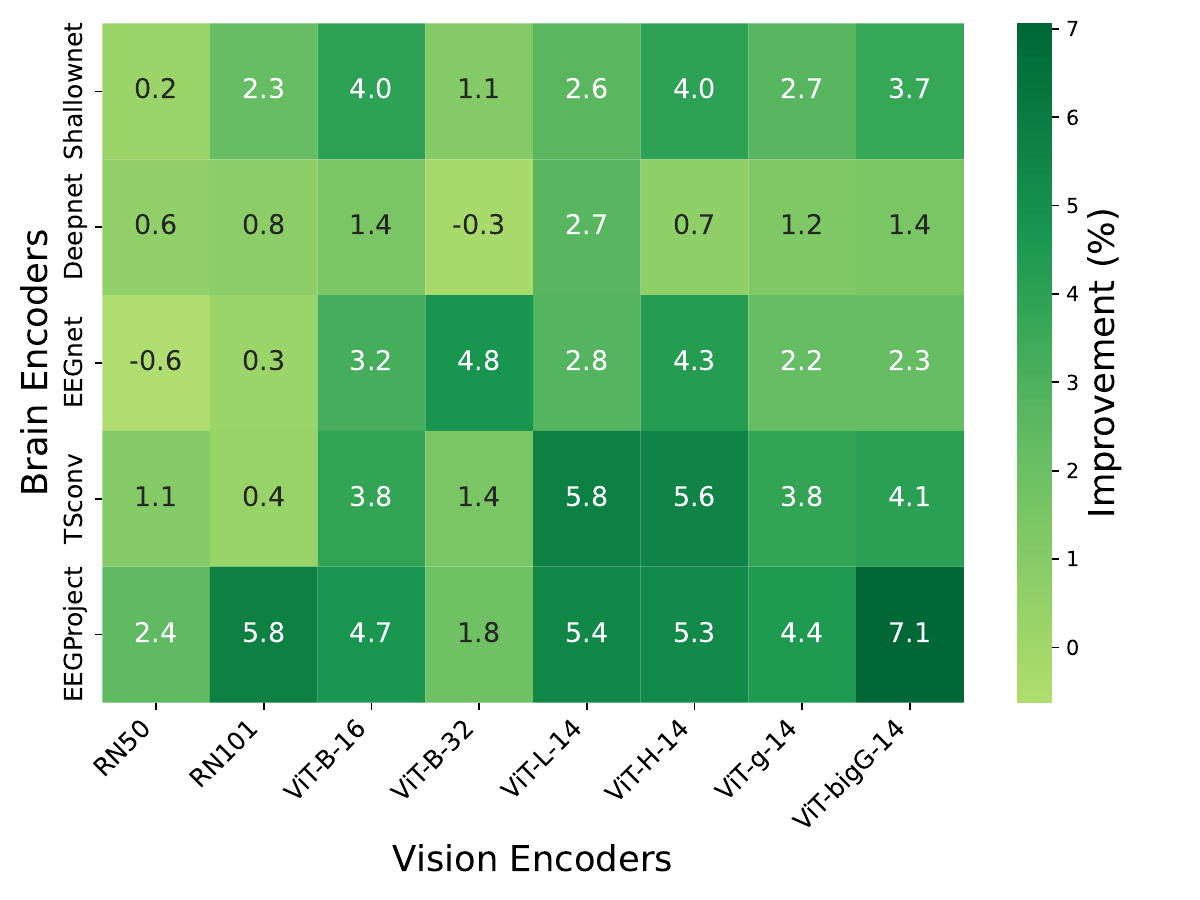}
    \caption{\textbf{Top-1} accuracy \textbf{improvement} (\%) on the THINGS-MEG dataset.}
  \end{subfigure}
  \hfill
  \begin{subfigure}[h]{0.45\linewidth}
    \centering
    \includegraphics[width=\linewidth]{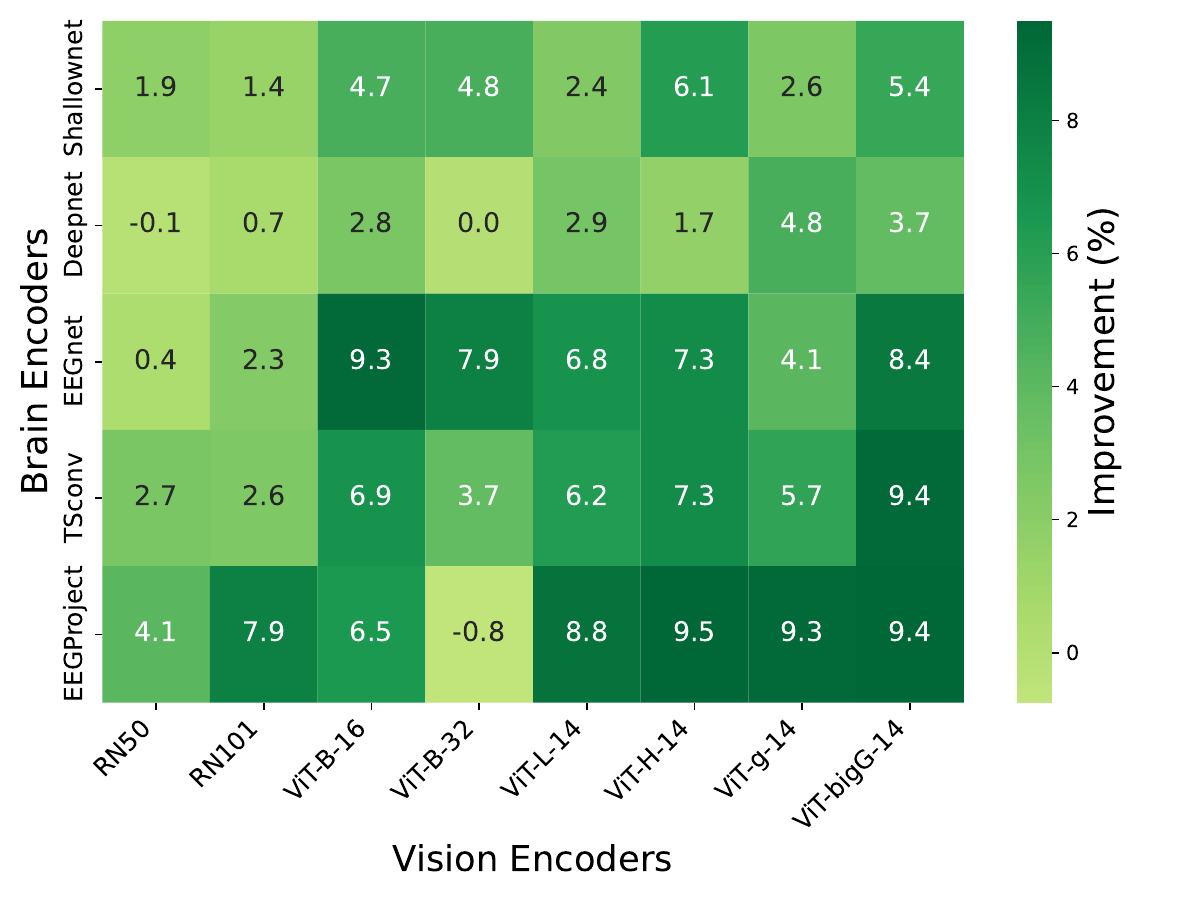}
    \caption{\textbf{Top-5} accuracy \textbf{improvement} (\%) on the THINGS-MEG dataset.}
  \end{subfigure}
  
  \caption{Results on the THINGS-EEG dataset.}
\end{figure*}
%





\twocolumn
\clearpage
\input{tabs/appendix_meg_backbone}

\clearpage

\begin{table}[h]
\centering
\caption{Details of different EEG encoders with Emb dimension of 1024}
\label{tab:brain_encoders}
\begin{tabular}{lcc}
\toprule
\textbf{Brain Encoder} & \textbf{Params}\\  
ShallowNet           & 2.56 M       \\  
DeepNet              & 2.76 M       \\  
EEGNet               & 2.34 M        \\  
TSConv               & 2.56 M        \\  
EEGProject           & 5.40 M        \\  
\bottomrule
\end{tabular}
\end{table}

\begin{table}[h]
\centering
\caption{Details of different Vision encoders}
\label{tab:vision_encoders}
\begin{tabular}{lcc}
\toprule
\textbf{Vision Encoder} & \textbf{Params} & \textbf{Emb dim} \\ 
RN50           & 38.32 M         & 1024              \\ 
RN101              & 56.26 M         & 512            \\ 
ViT-B-16               & 86.19 M         & 512             \\ 
ViT-B-32               & 87.85 M         & 512             \\ 
ViT-L-14           & 303.97 M         & 768             \\ 
ViT-H-14               & 632.08 M         & 1024             \\ 
ViT-g-14           & 1012.65 M         & 1024             \\ 
ViT-bigG-14  & 1844.91 M         & 1280             \\ 
\bottomrule
\end{tabular}
\end{table}

\subsection{EEG Feature Selection Analysis}
\label{subs:eeg_feature_selection}
\begin{figure}[h]
    \centering
    \includegraphics[width=1.0\linewidth]{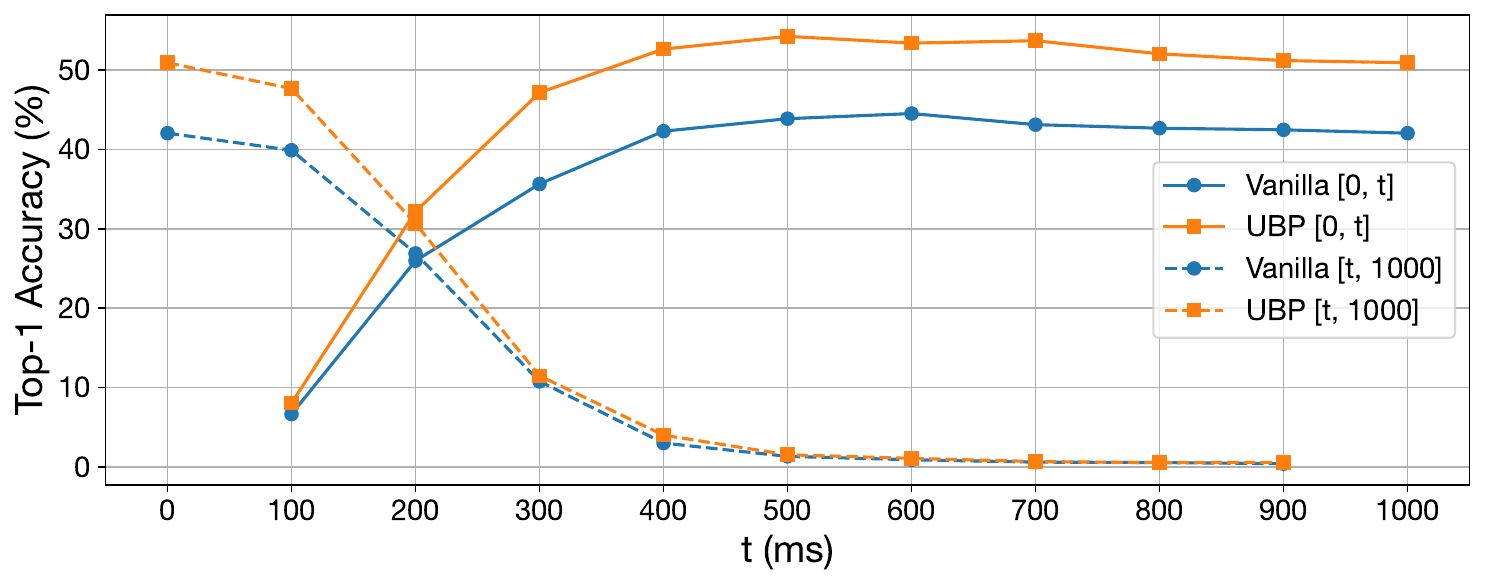}
    \caption{Top-1 ACC (\%) compariation with different epoching.}
    \label{fig:random_gap}
\end{figure}

Unless otherwise specified, EEG data spanning 1000 ms were selected, focusing on 17 visual-related (O+P) channels out of a total of 63, following previous work~\cite{song2024decoding}, where ablation studies are conducted on both channels and epochs.
To verify the influence of the selection of EEG features, we conducted relevant ablation experiments. Tab.~\ref{tab:channels} and Fig.~\ref{fig:random_gap} show consistent improvements across various settings.

\begin{table}[h]
\centering
\caption{Top-1 ACC (\%) compariation with different channels.}
\resizebox{\linewidth}{!}{
\begin{tabular}{llllll}
\toprule
\textbf{Method} & \textbf{Occipital} & \textbf{Parietal} & \textbf{O+P (Our)} & \textbf{Others} & \textbf{All}\\
\midrule
Vanilla & 21.0 & 33.7 & 42.0 & 10.2   & 35.6\\
UBP & 26.9 {\scriptsize (+5.9)} & 40.7 {\scriptsize (+7.0)} & 50.9 {\scriptsize (+8.9)} & 12.1 {\scriptsize (+1.9)} & 42.2 {\scriptsize (+6.6)}\\
\bottomrule
\end{tabular}
}
\label{tab:channels}
\end{table}

\subsection{More Evaluation Metrics}
\label{subs:map}
For a more comprehensive assessment of the method's performance, we provided additional metric values to validate UBP's effectiveness, specifically two new metrics: \textbf{mAP} and \textbf{Similarity Score} of paired samples. We present the performance averaged over 10 subjects. Tab.~\ref{tab:metrics} demonstrates the results of the proposed method evaluated across different metrics. The calculation of metrics can be found in the repo.

\begin{table}[h]
\centering
\caption{Retrieval Performance on THINGS-EEG.}
\resizebox{\linewidth}{!}{
\begin{tabular}{lllll}
\toprule
\textbf{Method} & \textbf{Top-1 Acc} & \textbf{Top-5 Acc} & \textbf{mAP*} & \textbf{Similarity*} \\
\midrule
Vanilla & 42.0 & 75.2 & 56.6 & 0.160 \\
UBP & 50.9 {\scriptsize (+8.9)} & 79.7 {\scriptsize (+4.5)} & 63.8 {\scriptsize (+7.2)} & 0.199 {\scriptsize (+0.039)} \\
\bottomrule
\end{tabular}
}
\label{tab:metrics}
\end{table}

%% file: tabs/appendix_eeg_backbone.tex
\begin{table*}[h!]
  \centering
  \caption{Top-1 and Top-5 Accuracy (\%) on THINGS-EEG with CLIP \textbf{RN50} With/Without UBP.}
 \Huge
  \resizebox{\linewidth}{!}{
  \begin{tabular}{lcccccccccccccccccccccc}
    \toprule
    & \multicolumn{2}{c}{Subject 1} & \multicolumn{2}{c}{Subject 2} & \multicolumn{2}{c}{Subject 3} & \multicolumn{2}{c}{Subject 4}  & \multicolumn{2}{c}{Subject 5}  & \multicolumn{2}{c}{Subject 6}  & \multicolumn{2}{c}{Subject 7}  & \multicolumn{2}{c}{Subject 8} & \multicolumn{2}{c}{Subject 9} & \multicolumn{2}{c}{Subject 10} & \multicolumn{2}{c}{Avg} \\
    \cmidrule(r){2-3} \cmidrule(r){4-5} \cmidrule(r){6-7} \cmidrule(r){8-9} \cmidrule(r){10-11} \cmidrule(r){12-13} \cmidrule(r){14-15} \cmidrule(r){16-17} \cmidrule(r){18-19} \cmidrule(r){20-21} \cmidrule(r){22-23}
    Backbone & top-1 & top-5 & top-1 & top-5 & top-1 & top-5 & top-1 & top-5 & top-1 & top-5 & top-1 & top-5 & top-1 & top-5 & top-1 & top-5 & top-1 & top-5 & top-1 & top-5 & top-1 & top-5\\
    \midrule

    ShallowNet & 18.0 & 48.3 & 24.0 & 54.8 & 23.0 & 58.8 & 30.0 & 63.2 & 20.3 & 45.7 & 27.8 & 62.5 & 26.0 & 60.5 & 32.8 & 69.7 & 24.5 & 59.3 & 33.0 & 69.2 & 25.9 & 59.2\\
    w/ UBP & 24.5 & 57.5 & 25.0 & 60.5 & 31.2 & 64.5 & 35.2 & 66.8 & 20.7 & 49.0 & 37.3 & 70.0 & 33.7 & 65.3 & 42.2 & 78.0 & 27.0 & 63.7 & 40.0 & 72.7 & 31.7 & 64.8\\
    \midrule
    DeepNet & 7.8 & 27.5 & 11.0 & 35.8 & 11.2 & 32.0 & 14.5 & 40.0 & 8.2 & 24.0 & 12.2 & 38.5 & 10.5 & 33.2 & 16.3 & 39.0 & 15.5 & 40.5 & 12.5 & 44.5 & 12.0 & 35.5\\ 
    w/ UBP & 10.5 & 33.2 & 10.8 & 35.3 & 11.7 & 37.0 & 17.8 & 44.2 & 6.5 & 23.3 & 17.0 & 40.5 & 14.2 & 36.2 & 14.5 & 42.0 & 16.0 & 36.7 & 17.5 & 45.7 & 13.7 & 37.4\\
    \midrule
    EEGNet & 11.0 & 38.3 & 18.5 & 51.7 & 19.5 & 51.3 & 24.0 & 56.2 & 16.3 & 41.8 & 24.5 & 57.2 & 19.5 & 43.8 & 30.0 & 62.5 & 24.2 & 50.0 & 29.5 & 61.0 & 21.7 & 51.4\\ 
    w/ UBP & 18.5 & 44.0 & 23.2 & 53.2 & 25.7 & 58.2 & 33.3 & 62.0 & 20.0 & 45.8 & 30.2 & 62.7 & 22.5 & 53.2 & 33.7 & 66.2 & 25.0 & 52.0 & 29.5 & 65.5 & 26.2 & 56.3\\
    \midrule
    TSConv & 29.3 & 61.3 & 27.0 & 63.2 & 32.0 & 73.0 & 35.0 & 72.5 & 25.2 & 59.0 & 38.0 & 74.3 & 31.0 & 66.5 & 43.0 & 76.8 & 31.5 & 66.5 & 40.0 & 75.5 & 33.2 & 68.9\\
    w/ UBP & 38.0 & 69.5 & 28.7 & 68.5 & 38.8 & 75.7 & 41.3 & 74.3 & 29.7 & 61.0 & 42.0 & 76.7 & 36.2 & 70.8 & 48.5 & 79.7 & 33.2 & 67.3 & 44.0 & 79.0 & 38.0 & 72.3\\
    \midrule
    EEGProject & 30.8 & 64.2 & 39.5 & 71.5 & 42.2 & 78.5 & 42.8 & 76.5 & 33.8 & 69.2 & 45.5 & 79.3 & 40.7 & 71.8 & 49.8 & 80.5 & 41.7 & 74.5 & 52.5 & 85.8 & 41.9 & 75.2\\
    w/ UBP & 41.2 & 71.2 & 50.7 & 82.2 & 50.2 & 81.7 & 49.3 & 77.8 & 43.3 & 70.8 & 58.7 & 82.5 & 49.8 & 79.0 & 57.8 & 84.3 & 44.5 & 76.7 & 62.3 & 87.7 & 50.8 & 79.4\\
    \bottomrule
  \end{tabular}}
\end{table*}

\begin{table*}[h!]
  \centering
  \caption{Top-1 and Top-5 Accuracy (\%) on THINGS-EEG with CLIP \textbf{RN101} With/Without UBP.}
 \Huge
  \resizebox{\linewidth}{!}{
  \begin{tabular}{lcccccccccccccccccccccc}
    \toprule
    & \multicolumn{2}{c}{Subject 1} & \multicolumn{2}{c}{Subject 2} & \multicolumn{2}{c}{Subject 3} & \multicolumn{2}{c}{Subject 4}  & \multicolumn{2}{c}{Subject 5}  & \multicolumn{2}{c}{Subject 6}  & \multicolumn{2}{c}{Subject 7}  & \multicolumn{2}{c}{Subject 8} & \multicolumn{2}{c}{Subject 9} & \multicolumn{2}{c}{Subject 10} & \multicolumn{2}{c}{Avg} \\
    \cmidrule(r){2-3} \cmidrule(r){4-5} \cmidrule(r){6-7} \cmidrule(r){8-9} \cmidrule(r){10-11} \cmidrule(r){12-13} \cmidrule(r){14-15} \cmidrule(r){16-17} \cmidrule(r){18-19} \cmidrule(r){20-21} \cmidrule(r){22-23}
    Backbone & top-1 & top-5 & top-1 & top-5 & top-1 & top-5 & top-1 & top-5 & top-1 & top-5 & top-1 & top-5 & top-1 & top-5 & top-1 & top-5 & top-1 & top-5 & top-1 & top-5 & top-1 & top-5\\
    \midrule

    ShallowNet  & 17.8 & 45.7 & 18.2 & 53.0 & 20.7 & 53.5 & 28.0 & 61.0 & 15.2 & 43.0 & 21.8 & 58.0 & 20.3 & 54.8 & 29.7 & 65.3 & 21.0 & 54.2 & 25.2 & 64.2 & 21.8 & 55.3\\
    w/ UBP & 23.5 & 55.3 & 19.5 & 56.0 & 26.0 & 57.7 & 29.7 & 63.7 & 15.5 & 42.5 & 29.7 & 66.0 & 22.5 & 56.5 & 32.8 & 68.3 & 22.0 & 59.5 & 32.5 & 67.5 & 25.4 & 59.3\\
    \midrule
    DeepNet & 8.7 & 23.5 & 11.3 & 29.5 & 5.7 & 23.3 & 15.0 & 38.0 & 7.2 & 21.5 & 10.2 & 31.0 & 10.0 & 28.7 & 9.5 & 33.2 & 12.5 & 35.8 & 17.2 & 44.5 & 10.8 & 30.9\\ 
    w/ UBP & 8.8 & 27.8 & 10.0 & 30.3 & 8.2 & 30.5 & 14.3 & 39.5 & 8.5 & 22.0 & 12.2 & 32.8 & 8.8 & 30.0 & 11.2 & 34.5 & 13.3 & 37.7 & 15.5 & 46.3 & 11.1 & 33.1\\
    \midrule
    EEGNet & 12.5 & 32.8 & 14.5 & 42.7 & 15.3 & 43.8 & 22.5 & 54.7 & 13.2 & 40.3 & 16.5 & 51.5 & 16.0 & 42.7 & 20.0 & 56.7 & 17.7 & 45.7 & 24.2 & 56.2 & 17.2 & 46.7\\ 
    w/ UBP & 18.0 & 43.8 & 18.8 & 52.0 & 19.5 & 51.7 & 22.3 & 55.8 & 18.3 & 46.3 & 26.0 & 60.5 & 18.2 & 43.5 & 21.5 & 58.5 & 21.7 & 50.0 & 26.7 & 62.3 & 21.1 & 52.4\\
    \midrule
    TSConv & 24.8 & 56.5 & 24.8 & 58.5 & 23.3 & 60.5 & 33.5 & 67.8 & 19.0 & 48.5 & 27.5 & 68.2 & 25.0 & 58.0 & 34.3 & 71.5 & 25.5 & 60.5 & 31.7 & 69.0 & 26.9 & 61.9\\
    w/ UBP & 30.8 & 65.0 & 24.2 & 65.2 & 29.5 & 65.2 & 34.0 & 69.5 & 24.2 & 55.0 & 40.0 & 77.0 & 26.5 & 62.3 & 41.2 & 77.8 & 29.7 & 66.2 & 39.5 & 74.8 & 32.0 & 67.8\\
    \midrule
    EEGProject & 29.0 & 59.3 & 36.0 & 66.3 & 40.7 & 72.5 & 42.0 & 73.5 & 30.5 & 58.8 & 39.7 & 72.0 & 39.2 & 70.5 & 41.0 & 77.5 & 37.3 & 67.0 & 50.5 & 79.5 & 38.6 & 69.7\\
    w/ UBP & 35.2 & 69.7 & 45.5 & 80.8 & 46.5 & 78.7 & 47.7 & 77.7 & 40.7 & 71.3 & 49.5 & 83.5 & 42.7 & 75.0 & 55.0 & 81.0 & 39.2 & 75.0 & 53.5 & 87.3 & 45.6 & 78.0\\
    \bottomrule
  \end{tabular}}
\end{table*}

\begin{table*}[h!]
  \centering
  \caption{Top-1 and Top-5 Accuracy (\%) on THINGS-EEG with CLIP \textbf{ViT-B-16} With/Without UBP.}
 \Huge
  \resizebox{\linewidth}{!}{
  \begin{tabular}{lcccccccccccccccccccccc}
    \toprule
    & \multicolumn{2}{c}{Subject 1} & \multicolumn{2}{c}{Subject 2} & \multicolumn{2}{c}{Subject 3} & \multicolumn{2}{c}{Subject 4}  & \multicolumn{2}{c}{Subject 5}  & \multicolumn{2}{c}{Subject 6}  & \multicolumn{2}{c}{Subject 7}  & \multicolumn{2}{c}{Subject 8} & \multicolumn{2}{c}{Subject 9} & \multicolumn{2}{c}{Subject 10} & \multicolumn{2}{c}{Avg} \\
    \cmidrule(r){2-3} \cmidrule(r){4-5} \cmidrule(r){6-7} \cmidrule(r){8-9} \cmidrule(r){10-11} \cmidrule(r){12-13} \cmidrule(r){14-15} \cmidrule(r){16-17} \cmidrule(r){18-19} \cmidrule(r){20-21} \cmidrule(r){22-23}
    Backbone & top-1 & top-5 & top-1 & top-5 & top-1 & top-5 & top-1 & top-5 & top-1 & top-5 & top-1 & top-5 & top-1 & top-5 & top-1 & top-5 & top-1 & top-5 & top-1 & top-5 & top-1 & top-5\\
    \midrule
    ShallowNet & 15.0 & 41.0 & 16.0 & 48.3 & 25.3 & 58.2 & 28.5 & 61.8 & 14.5 & 41.8 & 21.2 & 57.7 & 21.8 & 54.8 & 28.2 & 63.5 & 21.5 & 52.3 & 25.2 & 65.0 & 21.7 & 54.4\\
    w/ UBP & 22.3 & 51.8 & 23.5 & 53.0 & 24.0 & 61.8 & 32.3 & 65.0 & 16.8 & 45.0 & 28.2 & 64.0 & 27.2 & 60.0 & 34.7 & 71.0 & 23.0 & 55.3 & 34.3 & 67.0 & 26.6 & 59.4\\
    \midrule
    DeepNet & 9.0 & 22.5 & 6.7 & 29.3 & 9.0 & 30.5 & 13.5 & 36.8 & 4.7 & 17.8 & 10.0 & 31.0 & 10.0 & 31.2 & 9.8 & 33.2 & 15.3 & 37.5 & 14.2 & 39.0 & 10.2 & 30.9\\ 
    w/ UBP & 10.5 & 26.0 & 8.7 & 32.8 & 9.0 & 32.5 & 15.0 & 42.7 & 5.5 & 23.0 & 12.3 & 34.0 & 9.3 & 31.2 & 12.0 & 35.0 & 13.7 & 39.5 & 10.5 & 38.8 & 10.7 & 33.5\\
    \midrule
    EEGNet & 12.0 & 36.0 & 10.7 & 40.2 & 19.8 & 47.2 & 19.2 & 45.5 & 9.8 & 33.8 & 20.5 & 49.0 & 14.2 & 38.8 & 19.0 & 52.0 & 16.3 & 44.0 & 18.8 & 54.2 & 16.0 & 44.1\\ 
    w/ UBP & 18.8 & 41.5 & 18.5 & 48.5 & 23.5 & 55.0 & 24.7 & 57.0 & 16.5 & 44.0 & 27.0 & 57.2 & 20.7 & 49.8 & 29.3 & 58.5 & 23.0 & 50.2 & 27.5 & 61.2 & 22.9 & 52.3\\
    \midrule
    TSConv & 19.7 & 49.0 & 23.5 & 56.7 & 27.0 & 63.7 & 31.5 & 67.0 & 18.8 & 49.3 & 34.5 & 65.3 & 27.2 & 60.7 & 36.5 & 72.8 & 27.8 & 59.5 & 30.2 & 67.5 & 27.7 & 61.2\\
    w/ UBP & 28.7 & 63.0 & 28.0 & 63.0 & 31.2 & 66.8 & 37.0 & 72.2 & 22.5 & 54.5 & 41.0 & 74.8 & 33.7 & 63.0 & 42.2 & 76.7 & 31.7 & 63.5 & 36.3 & 71.5 & 33.2 & 66.9\\
    \midrule
    EEGProject & 29.5 & 59.5 & 32.8 & 68.0 & 36.5 & 71.8 & 35.3 & 68.0 & 28.2 & 54.0 & 37.2 & 73.3 & 34.3 & 70.0 & 36.8 & 75.0 & 37.3 & 69.2 & 44.0 & 79.7 & 35.2 & 68.9\\
    w/ UBP & 37.3 & 62.7 & 46.5 & 77.5 & 46.5 & 76.2 & 42.8 & 77.2 & 34.7 & 66.8 & 46.8 & 79.2 & 40.3 & 74.5 & 51.7 & 81.2 & 44.8 & 75.7 & 55.5 & 87.5 & 44.7 & 75.9\\
    \bottomrule
  \end{tabular}}
\end{table*}

\begin{table*}[h!]
  \centering
  \caption{Top-1 and Top-5 Accuracy (\%) on THINGS-EEG with CLIP \textbf{ViT-B-32} With/Without UBP.}
 \Huge
  \resizebox{\linewidth}{!}{
  \begin{tabular}{lcccccccccccccccccccccc}
    \toprule
    & \multicolumn{2}{c}{Subject 1} & \multicolumn{2}{c}{Subject 2} & \multicolumn{2}{c}{Subject 3} & \multicolumn{2}{c}{Subject 4}  & \multicolumn{2}{c}{Subject 5}  & \multicolumn{2}{c}{Subject 6}  & \multicolumn{2}{c}{Subject 7}  & \multicolumn{2}{c}{Subject 8} & \multicolumn{2}{c}{Subject 9} & \multicolumn{2}{c}{Subject 10} & \multicolumn{2}{c}{Avg} \\
    \cmidrule(r){2-3} \cmidrule(r){4-5} \cmidrule(r){6-7} \cmidrule(r){8-9} \cmidrule(r){10-11} \cmidrule(r){12-13} \cmidrule(r){14-15} \cmidrule(r){16-17} \cmidrule(r){18-19} \cmidrule(r){20-21} \cmidrule(r){22-23}
    Backbone & top-1 & top-5 & top-1 & top-5 & top-1 & top-5 & top-1 & top-5 & top-1 & top-5 & top-1 & top-5 & top-1 & top-5 & top-1 & top-5 & top-1 & top-5 & top-1 & top-5 & top-1 & top-5\\
    \midrule
    ShallowNet & 15.8 & 45.0 & 17.8 & 49.0 & 27.3 & 57.0 & 28.7 & 65.5 & 17.5 & 41.0 & 25.2 & 60.5 & 20.3 & 54.7 & 30.7 & 63.2 & 22.7 & 56.0 & 27.3 & 63.0 & 23.3 & 55.5\\
    w/ UBP & 20.0 & 53.2 & 22.0 & 55.0 & 26.5 & 63.0 & 35.3 & 69.2 & 19.5 & 48.7 & 27.3 & 65.7 & 27.7 & 62.7 & 37.0 & 73.5 & 22.5 & 60.8 & 29.7 & 69.3 & 26.7 & 62.1\\
    \midrule
    DeepNet & 8.8 & 23.8 & 10.2 & 30.5 & 9.8 & 29.3 & 12.0 & 39.0 & 5.3 & 19.7 & 10.7 & 32.8 & 11.2 & 32.0 & 9.0 & 31.8 & 13.5 & 37.3 & 12.8 & 39.7 & 10.3 & 31.6\\ 
    w/ UBP & 8.7 & 29.0 & 10.2 & 34.3 & 9.3 & 34.3 & 12.5 & 41.2 & 8.5 & 21.0 & 11.2 & 33.5 & 14.0 & 33.0 & 12.0 & 32.3 & 12.0 & 37.0 & 12.8 & 42.0 & 11.1 & 33.8\\
    \midrule
    EEGNet & 14.3 & 36.0 & 14.2 & 43.0 & 18.3 & 46.3 & 24.0 & 53.0 & 12.8 & 36.5 & 20.0 & 50.2 & 15.5 & 43.8 & 20.3 & 53.5 & 18.8 & 43.8 & 20.2 & 56.0 & 17.8 & 46.2\\ 
    w/ UBP & 18.5 & 40.5 & 21.7 & 54.7 & 20.7 & 56.0 & 25.0 & 58.2 & 19.2 & 45.3 & 30.2 & 60.5 & 20.3 & 48.3 & 27.0 & 63.5 & 22.0 & 54.0 & 25.2 & 63.5 & 23.0 & 54.4\\
    \midrule
    TSConv & 19.8 & 55.8 & 22.7 & 55.8 & 33.8 & 65.0 & 34.5 & 72.5 & 20.0 & 52.5 & 33.7 & 68.3 & 25.3 & 60.5 & 40.3 & 74.5 & 33.7 & 62.5 & 36.5 & 69.0 & 30.0 & 63.6\\
    w/ UBP & 30.8 & 63.5 & 27.5 & 64.0 & 37.5 & 70.8 & 36.8 & 72.3 & 25.8 & 57.7 & 43.5 & 76.0 & 34.8 & 68.0 & 41.5 & 78.7 & 34.0 & 69.0 & 36.2 & 72.5 & 34.8 & 69.2\\
    \midrule
    EEGProject & 30.7 & 59.5 & 39.2 & 70.5 & 43.0 & 79.5 & 40.3 & 73.5 & 31.2 & 65.0 & 42.0 & 74.3 & 36.8 & 68.8 & 45.7 & 75.0 & 41.2 & 71.3 & 46.5 & 83.2 & 39.7 & 72.1\\
    w/ UBP & 37.5 & 70.0 & 46.5 & 80.5 & 52.8 & 85.5 & 47.2 & 80.8 & 37.5 & 70.7 & 54.5 & 81.7 & 43.8 & 79.5 & 58.8 & 81.5 & 46.5 & 77.0 & 56.0 & 89.7 & 48.1 & 79.7\\
    \bottomrule
  \end{tabular}}
\end{table*}

\begin{table*}[h!]
  \centering
  \caption{Top-1 and Top-5 Accuracy (\%) on THINGS-EEG with CLIP \textbf{ViT-L-14} With/Without UBP.}
 \Huge
  \resizebox{\linewidth}{!}{
  \begin{tabular}{lcccccccccccccccccccccc}
    \toprule
    & \multicolumn{2}{c}{Subject 1} & \multicolumn{2}{c}{Subject 2} & \multicolumn{2}{c}{Subject 3} & \multicolumn{2}{c}{Subject 4}  & \multicolumn{2}{c}{Subject 5}  & \multicolumn{2}{c}{Subject 6}  & \multicolumn{2}{c}{Subject 7}  & \multicolumn{2}{c}{Subject 8} & \multicolumn{2}{c}{Subject 9} & \multicolumn{2}{c}{Subject 10} & \multicolumn{2}{c}{Avg} \\
    \cmidrule(r){2-3} \cmidrule(r){4-5} \cmidrule(r){6-7} \cmidrule(r){8-9} \cmidrule(r){10-11} \cmidrule(r){12-13} \cmidrule(r){14-15} \cmidrule(r){16-17} \cmidrule(r){18-19} \cmidrule(r){20-21} \cmidrule(r){22-23}
    Backbone & top-1 & top-5 & top-1 & top-5 & top-1 & top-5 & top-1 & top-5 & top-1 & top-5 & top-1 & top-5 & top-1 & top-5 & top-1 & top-5 & top-1 & top-5 & top-1 & top-5 & top-1 & top-5\\
    \midrule
    ShallowNet & 16.3 & 45.3 & 17.8 & 51.2 & 19.7 & 57.5 & 27.0 & 60.2 & 17.5 & 45.5 & 26.3 & 60.8 & 20.0 & 48.3 & 30.5 & 63.2 & 21.8 & 54.5 & 29.5 & 65.7 & 22.6 & 55.2\\
    w/ UBP & 15.7 & 47.8 & 23.3 & 52.0 & 24.5 & 61.2 & 32.2 & 66.7 & 17.2 & 50.5 & 33.2 & 69.2 & 25.7 & 61.5 & 35.5 & 71.0 & 23.5 & 58.8 & 33.2 & 70.2 & 26.4 & 60.9\\
    \midrule
    DeepNet & 8.3 & 23.0 & 9.0 & 29.0 & 9.0 & 29.7 & 12.8 & 34.7 & 7.3 & 20.5 & 11.5 & 33.3 & 8.3 & 31.7 & 13.0 & 36.3 & 13.5 & 36.5 & 12.0 & 33.7 & 10.5 & 30.9\\ 
    w/ UBP & 10.2 & 28.7 & 11.8 & 33.5 & 12.0 & 34.3 & 16.5 & 41.2 & 6.7 & 23.7 & 14.5 & 39.2 & 13.2 & 37.5 & 12.5 & 40.5 & 15.3 & 37.5 & 13.8 & 41.0 & 12.7 & 35.7\\
    \midrule
    EEGNet & 12.8 & 35.8 & 13.3 & 42.0 & 16.3 & 47.8 & 20.8 & 50.0 & 14.2 & 34.5 & 23.8 & 52.7 & 14.3 & 41.8 & 24.2 & 58.8 & 17.5 & 47.7 & 27.0 & 58.5 & 18.4 & 46.9\\ 
    w/ UBP & 14.3 & 40.5 & 20.7 & 45.5 & 21.7 & 54.0 & 29.0 & 59.5 & 18.0 & 45.7 & 27.5 & 61.3 & 19.7 & 53.0 & 26.3 & 61.3 & 21.8 & 54.0 & 27.5 & 63.0 & 22.6 & 53.8\\
    \midrule
    TSConv & 21.7 & 55.7 & 24.0 & 56.5 & 25.2 & 61.0 & 31.2 & 65.5 & 20.0 & 47.2 & 31.2 & 69.0 & 25.7 & 58.2 & 37.7 & 70.0 & 21.8 & 56.0 & 34.0 & 70.0 & 27.3 & 60.9\\
    w/ UBP & 26.5 & 60.0 & 26.0 & 60.5 & 32.3 & 66.5 & 36.3 & 72.0 & 24.5 & 56.0 & 39.0 & 75.2 & 28.2 & 65.3 & 42.8 & 78.5 & 29.0 & 63.5 & 39.0 & 76.0 & 32.4 & 67.3\\
    \midrule
    EEGProject & 26.7 & 55.3 & 33.5 & 66.5 & 35.2 & 73.3 & 32.0 & 70.0 & 27.5 & 51.7 & 36.2 & 72.2 & 33.5 & 72.8 & 44.5 & 75.5 & 29.5 & 67.0 & 45.3 & 77.7 & 34.4 & 68.2\\
    w/ UBP & 31.2 & 66.2 & 39.0 & 73.8 & 43.3 & 77.5 & 43.8 & 76.2 & 31.0 & 69.5 & 51.5 & 83.0 & 41.0 & 74.5 & 56.2 & 80.5 & 42.5 & 74.0 & 54.8 & 84.3 & 43.4 & 75.9\\
    \bottomrule
  \end{tabular}}
\end{table*}

\begin{table*}[h!]
  \centering
  \caption{Top-1 and Top-5 Accuracy (\%) on THINGS-EEG with CLIP \textbf{ViT-H-14} With/Without UBP.}
 \Huge
  \resizebox{\linewidth}{!}{
  \begin{tabular}{lcccccccccccccccccccccc}
    \toprule
    & \multicolumn{2}{c}{Subject 1} & \multicolumn{2}{c}{Subject 2} & \multicolumn{2}{c}{Subject 3} & \multicolumn{2}{c}{Subject 4}  & \multicolumn{2}{c}{Subject 5}  & \multicolumn{2}{c}{Subject 6}  & \multicolumn{2}{c}{Subject 7}  & \multicolumn{2}{c}{Subject 8} & \multicolumn{2}{c}{Subject 9} & \multicolumn{2}{c}{Subject 10} & \multicolumn{2}{c}{Avg} \\
    \cmidrule(r){2-3} \cmidrule(r){4-5} \cmidrule(r){6-7} \cmidrule(r){8-9} \cmidrule(r){10-11} \cmidrule(r){12-13} \cmidrule(r){14-15} \cmidrule(r){16-17} \cmidrule(r){18-19} \cmidrule(r){20-21} \cmidrule(r){22-23}
    Backbone & top-1 & top-5 & top-1 & top-5 & top-1 & top-5 & top-1 & top-5 & top-1 & top-5 & top-1 & top-5 & top-1 & top-5 & top-1 & top-5 & top-1 & top-5 & top-1 & top-5 & top-1 & top-5\\
    \midrule
    ShallowNet & 15.2 & 42.2 & 16.8 & 47.0 & 25.0 & 56.5 & 25.5 & 59.3 & 11.5 & 43.0 & 23.2 & 55.0 & 18.0 & 51.0 & 27.8 & 62.7 & 22.7 & 55.5 & 29.5 & 60.0 & 21.5 & 53.2\\
    w/ UBP & 21.0 & 51.0 & 21.8 & 50.3 & 24.2 & 62.7 & 32.0 & 64.3 & 17.8 & 45.7 & 30.5 & 69.5 & 24.2 & 60.0 & 31.7 & 71.8 & 26.2 & 59.5 & 38.3 & 67.8 & 26.8 & 60.3\\
    \midrule
    DeepNet & 8.0 & 24.8 & 9.0 & 27.2 & 11.5 & 31.0 & 9.0 & 30.3 & 4.5 & 20.7 & 9.5 & 27.8 & 9.5 & 30.0 & 8.8 & 32.0 & 11.2 & 34.0 & 13.3 & 41.0 & 9.4 & 29.9\\ 
    w/ UBP & 9.3 & 28.7 & 14.3 & 36.5 & 14.0 & 37.3 & 14.0 & 36.5 & 8.5 & 22.2 & 13.8 & 33.8 & 11.7 & 36.5 & 14.5 & 38.0 & 14.0 & 38.5 & 15.7 & 45.8 & 13.0 & 35.4\\
    \midrule
    EEGNet & 9.3 & 30.8 & 17.3 & 43.5 & 18.8 & 46.0 & 18.5 & 47.3 & 12.5 & 36.2 & 21.2 & 49.0 & 14.0 & 40.5 & 21.2 & 51.0 & 17.7 & 48.5 & 21.0 & 53.5 & 17.2 & 44.6\\ 
    w/ UBP & 15.0 & 42.5 & 19.7 & 49.0 & 22.0 & 54.0 & 26.0 & 58.5 & 20.5 & 44.0 & 22.5 & 58.0 & 21.2 & 49.2 & 26.0 & 60.0 & 21.5 & 54.5 & 27.0 & 63.8 & 22.1 & 53.4\\
    \midrule
    TSConv & 20.5 & 50.0 & 22.0 & 55.8 & 27.2 & 60.5 & 29.5 & 61.8 & 19.0 & 49.8 & 30.5 & 61.5 & 22.7 & 57.8 & 35.0 & 70.0 & 27.0 & 59.0 & 34.7 & 66.7 & 26.8 & 59.3\\
    w/ UBP & 28.7 & 63.3 & 29.7 & 65.0 & 32.8 & 69.5 & 37.0 & 71.0 & 28.7 & 59.0 & 39.5 & 76.2 & 33.2 & 65.3 & 40.0 & 77.8 & 29.7 & 63.7 & 43.0 & 76.5 & 34.2 & 68.7\\
    \midrule
    EEGProject & 19.7 & 54.5 & 31.0 & 62.0 & 39.0 & 70.0 & 35.0 & 62.0 & 26.3 & 54.0 & 38.2 & 68.5 & 29.3 & 61.8 & 39.5 & 72.2 & 34.8 & 62.2 & 40.7 & 76.0 & 33.4 & 64.3\\
    w/ UBP & 32.0 & 64.0 & 45.0 & 75.7 & 45.8 & 81.5 & 42.5 & 77.5 & 34.7 & 66.5 & 47.0 & 81.2 & 40.0 & 73.5 & 51.2 & 82.5 & 45.7 & 72.3 & 51.2 & 86.5 & 43.5 & 76.1\\
    \bottomrule
  \end{tabular}}
\end{table*}

\begin{table*}[h!]
  \centering
  \caption{Top-1 and Top-5 Accuracy (\%) on THINGS-EEG with CLIP \textbf{ViT-g-14} With/Without UBP.}
 \Huge
  \resizebox{\linewidth}{!}{
  \begin{tabular}{lcccccccccccccccccccccc}
    \toprule
    & \multicolumn{2}{c}{Subject 1} & \multicolumn{2}{c}{Subject 2} & \multicolumn{2}{c}{Subject 3} & \multicolumn{2}{c}{Subject 4}  & \multicolumn{2}{c}{Subject 5}  & \multicolumn{2}{c}{Subject 6}  & \multicolumn{2}{c}{Subject 7}  & \multicolumn{2}{c}{Subject 8} & \multicolumn{2}{c}{Subject 9} & \multicolumn{2}{c}{Subject 10} & \multicolumn{2}{c}{Avg} \\
    \cmidrule(r){2-3} \cmidrule(r){4-5} \cmidrule(r){6-7} \cmidrule(r){8-9} \cmidrule(r){10-11} \cmidrule(r){12-13} \cmidrule(r){14-15} \cmidrule(r){16-17} \cmidrule(r){18-19} \cmidrule(r){20-21} \cmidrule(r){22-23}
    Backbone & top-1 & top-5 & top-1 & top-5 & top-1 & top-5 & top-1 & top-5 & top-1 & top-5 & top-1 & top-5 & top-1 & top-5 & top-1 & top-5 & top-1 & top-5 & top-1 & top-5 & top-1 & top-5\\
    \midrule
    ShallowNet & 6.5 & 23.5 & 4.7 & 21.0 & 6.2 & 28.8 & 9.8 & 31.0 & 5.0 & 17.0 & 7.3 & 25.5 & 6.8 & 24.7 & 11.3 & 33.5 & 5.7 & 25.0 & 10.8 & 34.5 & 7.4 & 26.5\\
    w/ UBP & 19.5 & 51.5 & 16.8 & 50.2 & 26.5 & 61.0 & 29.0 & 66.0 & 17.5 & 45.0 & 29.0 & 62.7 & 22.5 & 58.7 & 27.5 & 67.5 & 23.7 & 58.2 & 30.2 & 67.0 & 24.2 & 58.8\\
    \midrule
    DeepNet & 6.5 & 23.7 & 6.8 & 24.8 & 10.8 & 31.5 & 12.2 & 32.3 & 6.5 & 17.0 & 9.5 & 26.2 & 9.0 & 29.0 & 10.5 & 32.2 & 9.8 & 30.8 & 7.0 & 37.7 & 8.9 & 28.5\\ 
    w/ UBP & 9.5 & 26.0 & 8.8 & 26.5 & 11.0 & 33.2 & 12.8 & 33.2 & 6.0 & 21.5 & 12.0 & 33.7 & 11.0 & 36.0 & 9.5 & 33.7 & 11.0 & 32.8 & 14.2 & 38.5 & 10.6 & 31.5\\
    \midrule
    EEGNet & 8.3 & 29.3 & 13.3 & 39.2 & 19.3 & 47.0 & 20.3 & 48.7 & 11.5 & 35.0 & 19.5 & 44.0 & 15.0 & 36.8 & 20.0 & 52.2 & 16.5 & 40.5 & 22.0 & 50.7 & 16.6 & 42.3\\ 
    w/ UBP & 13.3 & 37.5 & 16.0 & 42.7 & 22.7 & 52.2 & 25.0 & 55.8 & 14.0 & 44.2 & 23.0 & 52.5 & 18.8 & 47.5 & 25.2 & 58.0 & 18.0 & 50.7 & 26.8 & 57.8 & 20.3 & 49.9\\
    \midrule
    TSConv & 24.5 & 52.0 & 20.0 & 50.5 & 26.0 & 57.7 & 27.5 & 61.0 & 20.5 & 48.8 & 25.3 & 58.5 & 24.5 & 51.0 & 31.0 & 63.0 & 22.7 & 55.0 & 29.7 & 64.5 & 25.2 & 56.2\\
    w/ UBP & 31.5 & 63.5 & 21.2 & 54.7 & 32.8 & 64.3 & 32.2 & 67.0 & 23.7 & 55.8 & 33.7 & 69.5 & 32.8 & 64.0 & 37.0 & 77.2 & 29.3 & 62.3 & 39.0 & 73.3 & 31.3 & 65.1\\
    \midrule
    EEGProject & 20.3 & 48.5 & 23.5 & 52.3 & 30.0 & 59.7 & 28.5 & 62.7 & 23.3 & 48.3 & 29.0 & 60.2 & 28.2 & 54.8 & 30.7 & 64.0 & 29.0 & 57.8 & 30.5 & 65.5 & 27.3 & 57.4\\
    w/ UBP & 29.0 & 62.0 & 38.5 & 67.5 & 42.2 & 77.2 & 41.0 & 74.8 & 31.0 & 59.5 & 44.3 & 74.8 & 37.3 & 71.0 & 48.5 & 81.0 & 42.0 & 69.7 & 50.2 & 81.2 & 40.4 & 71.9\\
    \bottomrule
  \end{tabular}}
\end{table*}
\begin{table*}[h!]
  \centering
  \caption{Top-1 and Top-5 Accuracy (\%) on THINGS-EEG with CLIP \textbf{ViT-bigG-14} With/Without UBP.}
 \Huge
  \resizebox{\linewidth}{!}{
  \begin{tabular}{lcccccccccccccccccccccc}
    \toprule
    & \multicolumn{2}{c}{Subject 1} & \multicolumn{2}{c}{Subject 2} & \multicolumn{2}{c}{Subject 3} & \multicolumn{2}{c}{Subject 4}  & \multicolumn{2}{c}{Subject 5}  & \multicolumn{2}{c}{Subject 6}  & \multicolumn{2}{c}{Subject 7}  & \multicolumn{2}{c}{Subject 8} & \multicolumn{2}{c}{Subject 9} & \multicolumn{2}{c}{Subject 10} & \multicolumn{2}{c}{Avg} \\
    \cmidrule(r){2-3} \cmidrule(r){4-5} \cmidrule(r){6-7} \cmidrule(r){8-9} \cmidrule(r){10-11} \cmidrule(r){12-13} \cmidrule(r){14-15} \cmidrule(r){16-17} \cmidrule(r){18-19} \cmidrule(r){20-21} \cmidrule(r){22-23}
    Backbone & top-1 & top-5 & top-1 & top-5 & top-1 & top-5 & top-1 & top-5 & top-1 & top-5 & top-1 & top-5 & top-1 & top-5 & top-1 & top-5 & top-1 & top-5 & top-1 & top-5 & top-1 & top-5\\
    \midrule
    ShallowNet & 7.8 & 26.5 & 6.3 & 19.0 & 9.0 & 33.8 & 9.8 & 33.0 & 6.0 & 20.0 & 10.2 & 29.7 & 7.8 & 32.3 & 12.3 & 39.0 & 9.2 & 32.3 & 13.0 & 40.3 & 9.1 & 30.6\\
    w/ UBP & 19.5 & 52.8 & 19.0 & 57.5 & 29.7 & 60.7 & 34.3 & 68.3 & 17.5 & 46.2 & 34.0 & 67.0 & 28.2 & 59.7 & 33.2 & 71.0 & 27.2 & 59.0 & 31.2 & 72.0 & 27.4 & 61.4\\
    \midrule
    DeepNet & 7.3 & 21.5 & 8.0 & 27.8 & 9.0 & 31.5 & 10.5 & 32.5 & 7.5 & 21.7 & 10.5 & 29.3 & 10.8 & 29.7 & 10.0 & 29.7 & 9.8 & 32.0 & 11.0 & 34.7 & 9.4 & 29.1\\ 
    w/ UBP & 7.3 & 21.5 & 8.0 & 27.8 & 9.0 & 31.5 & 10.5 & 32.5 & 7.5 & 21.7 & 10.5 & 29.3 & 10.8 & 29.7 & 10.0 & 29.7 & 9.8 & 32.0 & 11.0 & 34.7 & 9.4 & 29.1\\
    \midrule
    EEGNet & 13.0 & 36.2 & 13.0 & 43.0 & 20.0 & 47.2 & 23.2 & 51.5 & 17.0 & 36.7 & 18.8 & 50.7 & 16.8 & 42.5 & 25.5 & 58.0 & 19.3 & 50.5 & 23.8 & 54.5 & 19.0 & 47.1\\ 
    w/ UBP & 16.3 & 40.7 & 20.5 & 53.7 & 24.0 & 55.8 & 28.0 & 58.8 & 18.5 & 45.0 & 28.0 & 59.7 & 23.2 & 51.2 & 28.7 & 62.5 & 24.5 & 55.5 & 25.5 & 62.7 & 23.7 & 54.6\\
    \midrule
    TSConv & 22.0 & 48.0 & 18.2 & 57.2 & 26.7 & 58.5 & 31.2 & 61.0 & 19.8 & 46.0 & 32.5 & 63.5 & 24.7 & 53.7 & 38.5 & 68.0 & 24.8 & 58.0 & 35.8 & 69.5 & 27.4 & 58.3\\
    w/ UBP & 30.0 & 62.0 & 28.5 & 64.5 & 32.0 & 66.2 & 36.8 & 70.7 & 23.7 & 54.7 & 43.3 & 74.3 & 32.2 & 67.5 & 42.0 & 75.2 & 30.5 & 65.5 & 37.3 & 76.0 & 33.6 & 67.7\\
    \midrule
    EEGProject & 22.0 & 52.5 & 29.7 & 58.5 & 31.5 & 63.7 & 29.7 & 63.7 & 25.2 & 53.5 & 36.2 & 67.3 & 32.5 & 63.0 & 37.5 & 68.2 & 26.3 & 61.3 & 39.2 & 72.0 & 31.0 & 62.4\\
    w/ UBP & 33.8 & 63.7 & 41.8 & 78.2 & 43.8 & 78.2 & 39.7 & 73.8 & 34.0 & 64.0 & 55.8 & 82.7 & 38.7 & 74.8 & 49.0 & 80.3 & 43.5 & 74.3 & 53.2 & 85.5 & 43.3 & 75.6\\
    \bottomrule
  \end{tabular}}
\end{table*}

%% file: tabs/appendix_meg_backbone.tex
\begin{table}[t]
  \centering
  \caption{Top-1 and Top-5 Accuracy (\%) on THINGS-MEG with CLIP \textbf{RN50} With/Without UBP.}
  \label{tab:backbone_meg}
  \Huge
  \resizebox{1.0\linewidth}{!}{
  \begin{tabular}{lcccccccccc}
    \toprule
    & \multicolumn{2}{c}{Subject 1} & \multicolumn{2}{c}{Subject 2} & \multicolumn{2}{c}{Subject 3} & \multicolumn{2}{c}{Subject 4}& \multicolumn{2}{c}{Avg} \\
    \cmidrule(r){2-3} \cmidrule(r){4-5} \cmidrule(r){6-7} \cmidrule(r){8-9} \cmidrule(r){10-11}
    Backbone & top-1 & top-5 & top-1 & top-5 & top-1 & top-5 & top-1 & top-5 & top-1 & top-5 \\
    \midrule
    ShallowNet  & 9.8 & 32.0 & 27.3 & 58.5 & 25.2 & 56.2 & 9.3 & 29.7 & 17.9 & 44.1 \\
w/ UBP & 10.7 & 36.0 & 27.0 & 65.3 & 22.2 & 51.8 & 12.5 & 31.0 & 18.1 & 46.0 \\
\midrule
DeepNet  & 4.3 & 17.0 & 13.3 & 40.0 & 12.8 & 34.7 & 6.2 & 19.2 & 9.1 & 27.7 \\
w/ UBP & 4.5 & 14.8 & 16.5 & 40.5 & 11.3 & 36.0 & 6.7 & 19.2 & 9.7 & 27.6 \\
\midrule
EEGNet  & 10.5 & 26.8 & 25.2 & 59.3 & 19.3 & 47.5 & 8.0 & 28.2 & 15.7 & 40.4 \\
w/ UBP & 7.5 & 24.5 & 29.7 & 66.0 & 14.5 & 46.8 & 8.7 & 26.3 & 15.1 & 40.9 \\
\midrule
TSconv & 14.8 & 40.5 & 39.2 & 73.5 & 30.2 & 60.8 & 15.5 & 36.2 & 24.9 & 52.8 \\
w/ UBP & 18.3 & 45.0 & 40.7 & 76.5 & 28.0 & 59.0 & 17.0 & 41.5 & 26.0 & 55.5 \\
\midrule
EEGProject & 13.2 & 34.5 & 43.3 & 74.5 & 27.0 & 57.0 & 13.5 & 38.8 & 24.2 & 51.2 \\
w/ UBP & 15.0 & 38.0 & 46.0 & 80.5 & 27.3 & 59.0 & 18.5 & 43.5 & 26.7 & 55.2 \\
    \bottomrule
  \end{tabular}}
\end{table}

\begin{table}[t]
  \centering
  \caption{Top-1 and Top-5 Accuracy (\%) on THINGS-MEG with CLIP \textbf{RN101} With/Without UBP.}
  \label{tab:backbone_meg}
  \Huge
  \resizebox{1.0\linewidth}{!}{
  \begin{tabular}{lcccccccccc}
    \toprule
    & \multicolumn{2}{c}{Subject 1} & \multicolumn{2}{c}{Subject 2} & \multicolumn{2}{c}{Subject 3} & \multicolumn{2}{c}{Subject 4}& \multicolumn{2}{c}{Avg} \\
    \cmidrule(r){2-3} \cmidrule(r){4-5} \cmidrule(r){6-7} \cmidrule(r){8-9} \cmidrule(r){10-11}
    Backbone & top-1 & top-5 & top-1 & top-5 & top-1 & top-5 & top-1 & top-5 & top-1 & top-5 \\
    \midrule
    ShallowNet  & 8.5 & 27.8 & 20.0 & 53.0 & 21.0 & 49.7 & 7.7 & 28.2 & 14.3 & 39.7 \\
w/ UBP & 8.7 & 30.0 & 25.0 & 55.3 & 21.5 & 46.3 & 11.0 & 33.0 & 16.6 & 41.1 \\
\midrule
DeepNet  & 4.8 & 15.0 & 11.5 & 36.2 & 8.3 & 30.5 & 4.3 & 19.3 & 7.2 & 25.2 \\
w/ UBP & 3.5 & 17.0 & 12.3 & 38.5 & 9.5 & 27.5 & 6.8 & 20.7 & 8.0 & 25.9 \\
\midrule
EEGNet  & 6.5 & 18.0 & 17.2 & 47.2 & 14.2 & 36.5 & 4.0 & 17.7 & 10.5 & 29.9 \\
w/ UBP & 6.0 & 20.7 & 19.3 & 49.8 & 11.8 & 36.5 & 6.0 & 21.8 & 10.8 & 32.2 \\
\midrule
TSconv & 15.0 & 37.7 & 32.0 & 67.2 & 24.7 & 55.5 & 14.3 & 38.5 & 21.5 & 49.7 \\
w/ UBP & 13.5 & 40.5 & 31.5 & 68.8 & 25.5 & 58.3 & 17.0 & 41.7 & 21.9 & 52.3 \\
\midrule
EEGProject   & 10.7 & 33.2 & 37.3 & 69.5 & 26.2 & 55.5 & 9.0 & 31.0 & 20.8 & 47.3 \\
w/ UBP & 15.3 & 37.7 & 44.8 & 76.5 & 31.0 & 63.5 & 15.5 & 43.3 & 26.6 & 55.2 \\
    \bottomrule
  \end{tabular}}
\end{table}

\begin{table}[b]
  \centering
  \caption{Top-1 and Top-5 Accuracy (\%) on THINGS-MEG with CLIP \textbf{ViT-B-16} With/Without UBP.}
  \label{tab:backbone_meg}
  \Huge
  \resizebox{1.0\linewidth}{!}{
  \begin{tabular}{lcccccccccc}
    \toprule
    & \multicolumn{2}{c}{Subject 1} & \multicolumn{2}{c}{Subject 2} & \multicolumn{2}{c}{Subject 3} & \multicolumn{2}{c}{Subject 4}& \multicolumn{2}{c}{Avg} \\
    \cmidrule(r){2-3} \cmidrule(r){4-5} \cmidrule(r){6-7} \cmidrule(r){8-9} \cmidrule(r){10-11}
    Backbone & top-1 & top-5 & top-1 & top-5 & top-1 & top-5 & top-1 & top-5 & top-1 & top-5 \\
    \midrule
    ShallowNet  & 8.2 & 28.5 & 21.5 & 49.8 & 20.5 & 45.0 & 6.2 & 27.8 & 14.1 & 37.8 \\
w/ UBP & 9.8 & 28.8 & 25.0 & 59.7 & 25.0 & 52.8 & 12.8 & 28.7 & 18.1 & 42.5 \\
\midrule
DeepNet  & 4.0 & 15.3 & 14.8 & 35.3 & 9.0 & 32.2 & 3.5 & 18.3 & 7.8 & 25.3 \\
w/ UBP & 6.8 & 19.0 & 14.8 & 35.3 & 8.0 & 35.8 & 7.5 & 22.3 & 9.3 & 28.1 \\
\midrule
EEGNet  & 3.0 & 12.8 & 14.5 & 40.3 & 9.5 & 29.5 & 2.7 & 13.8 & 7.4 & 24.1 \\
w/ UBP & 5.3 & 20.8 & 18.5 & 54.5 & 13.7 & 38.5 & 5.3 & 19.7 & 10.7 & 33.4 \\
\midrule
TSconv & 10.8 & 33.3 & 29.3 & 62.3 & 26.0 & 54.8 & 10.7 & 32.5 & 19.2 & 45.7 \\
w/ UBP & 13.5 & 40.3 & 35.3 & 70.3 & 29.0 & 62.3 & 14.3 & 37.5 & 23.0 & 52.6 \\
\midrule
EEGProject   & 14.2 & 35.8 & 27.8 & 65.3 & 23.7 & 52.0 & 11.0 & 35.0 & 19.2 & 47.0 \\
w/ UBP & 16.7 & 39.2 & 38.2 & 72.5 & 23.7 & 62.3 & 16.8 & 40.0 & 23.9 & 53.5 \\
    \bottomrule
  \end{tabular}}
\end{table}

\begin{table}[b]
  \centering
  \caption{Top-1 and Top-5 Accuracy (\%) on THINGS-MEG with CLIP \textbf{ViT-B-32} With/Without UBP.}
  \label{tab:backbone_meg}
  \Huge
  \resizebox{1.0\linewidth}{!}{
  \begin{tabular}{lcccccccccc}
    \toprule
    & \multicolumn{2}{c}{Subject 1} & \multicolumn{2}{c}{Subject 2} & \multicolumn{2}{c}{Subject 3} & \multicolumn{2}{c}{Subject 4}& \multicolumn{2}{c}{Avg} \\
    \cmidrule(r){2-3} \cmidrule(r){4-5} \cmidrule(r){6-7} \cmidrule(r){8-9} \cmidrule(r){10-11}
    Backbone & top-1 & top-5 & top-1 & top-5 & top-1 & top-5 & top-1 & top-5 & top-1 & top-5 \\
    \midrule
    ShallowNet  & 8.2 & 24.2 & 24.0 & 54.8 & 18.3 & 46.0 & 9.5 & 26.2 & 15.0 & 37.8 \\
w/ UBP & 7.5 & 30.0 & 23.3 & 54.8 & 21.8 & 55.8 & 11.7 & 29.7 & 16.1 & 42.6 \\
\midrule
DeepNet  & 5.0 & 18.5 & 12.5 & 38.5 & 12.2 & 32.2 & 5.3 & 18.2 & 8.7 & 26.9 \\
w/ UBP & 4.7 & 18.3 & 13.3 & 37.5 & 10.7 & 32.8 & 5.2 & 19.0 & 8.5 & 26.9 \\
\midrule
EEGNet  & 4.5 & 15.7 & 15.7 & 45.5 & 8.5 & 31.2 & 2.2 & 13.5 & 7.7 & 26.5 \\
w/ UBP & 8.0 & 22.5 & 24.5 & 58.0 & 13.3 & 41.0 & 4.2 & 16.3 & 12.5 & 34.4 \\
\midrule
TSconv & 16.3 & 38.5 & 34.3 & 68.5 & 24.2 & 54.0 & 14.0 & 37.7 & 22.2 & 49.7 \\
w/ UBP & 14.8 & 42.2 & 37.3 & 69.2 & 27.8 & 62.0 & 14.8 & 40.2 & 23.6 & 53.4 \\
\midrule
EEGProject   & 10.0 & 35.8 & 34.7 & 70.3 & 25.5 & 54.5 & 14.0 & 41.3 & 21.1 & 50.4 \\
w/ UBP & 15.0 & 38.0 & 34.0 & 66.5 & 29.7 & 63.3 & 12.7 & 31.0 & 22.9 & 49.7 \\
    \bottomrule
  \end{tabular}}
\end{table}

\begin{table}[t]
  \centering
  \caption{Top-1 and Top-5 Accuracy (\%) on THINGS-MEG with CLIP \textbf{ViT-L-14} With/Without UBP.}
  \label{tab:backbone_meg}
  \Huge
  \resizebox{1.0\linewidth}{!}{
  \begin{tabular}{lcccccccccc}
    \toprule
    & \multicolumn{2}{c}{Subject 1} & \multicolumn{2}{c}{Subject 2} & \multicolumn{2}{c}{Subject 3} & \multicolumn{2}{c}{Subject 4}& \multicolumn{2}{c}{Avg} \\
    \cmidrule(r){2-3} \cmidrule(r){4-5} \cmidrule(r){6-7} \cmidrule(r){8-9} \cmidrule(r){10-11}
    Backbone & top-1 & top-5 & top-1 & top-5 & top-1 & top-5 & top-1 & top-5 & top-1 & top-5 \\
    \midrule
    ShallowNet  & 6.5 & 26.0 & 23.0 & 51.7 & 19.0 & 44.0 & 10.2 & 29.5 & 14.7 & 37.8 \\
w/ UBP & 8.7 & 27.5 & 30.8 & 57.7 & 17.5 & 48.0 & 12.0 & 27.5 & 17.3 & 40.2 \\
\midrule
DeepNet  & 4.5 & 18.0 & 12.3 & 37.3 & 11.3 & 34.3 & 5.0 & 16.8 & 8.3 & 26.6 \\
w/ UBP & 6.2 & 18.8 & 18.5 & 44.0 & 12.8 & 34.5 & 6.2 & 20.7 & 10.9 & 29.5 \\
\midrule
EEGNet  & 6.2 & 18.5 & 18.2 & 45.0 & 11.7 & 33.8 & 6.0 & 18.5 & 10.6 & 28.9 \\
w/ UBP & 5.5 & 24.7 & 27.5 & 55.5 & 14.7 & 42.2 & 5.5 & 20.5 & 13.3 & 35.7 \\
\midrule
TSconv & 15.0 & 34.8 & 28.8 & 64.0 & 21.5 & 50.0 & 12.5 & 35.0 & 19.4 & 45.9 \\
w/ UBP & 15.5 & 41.2 & 39.5 & 69.0 & 28.2 & 57.2 & 17.8 & 41.2 & 25.2 & 52.2 \\
\midrule
EEGProject   & 12.2 & 31.2 & 32.5 & 64.0 & 16.8 & 47.0 & 13.0 & 31.5 & 18.6 & 43.4 \\
w/ UBP & 14.8 & 37.5 & 39.0 & 72.5 & 24.5 & 56.2 & 18.0 & 42.8 & 24.1 & 52.3 \\
    \bottomrule
  \end{tabular}}
\end{table}

\begin{table}[t]
  \centering
  \caption{Top-1 and Top-5 Accuracy (\%) on THINGS-MEG with CLIP \textbf{ViT-H-14} With/Without UBP.}
  \label{tab:backbone_meg}
  \Huge
  \resizebox{1.0\linewidth}{!}{
  \begin{tabular}{lcccccccccc}
    \toprule
    & \multicolumn{2}{c}{Subject 1} & \multicolumn{2}{c}{Subject 2} & \multicolumn{2}{c}{Subject 3} & \multicolumn{2}{c}{Subject 4}& \multicolumn{2}{c}{Avg} \\
    \cmidrule(r){2-3} \cmidrule(r){4-5} \cmidrule(r){6-7} \cmidrule(r){8-9} \cmidrule(r){10-11}
    Backbone & top-1 & top-5 & top-1 & top-5 & top-1 & top-5 & top-1 & top-5 & top-1 & top-5 \\
    \midrule
    ShallowNet  & 7.5 & 26.0 & 24.5 & 50.5 & 14.5 & 43.8 & 9.0 & 25.5 & 13.9 & 36.4 \\
w/ UBP & 11.0 & 34.5 & 28.7 & 55.8 & 20.0 & 53.7 & 11.7 & 26.2 & 17.9 & 42.6 \\
\midrule
DeepNet  & 5.3 & 18.3 & 16.0 & 39.2 & 11.0 & 30.0 & 5.0 & 17.2 & 9.3 & 26.2 \\
w/ UBP & 4.3 & 19.5 & 18.0 & 41.8 & 12.7 & 33.2 & 5.2 & 17.0 & 10.1 & 27.9 \\
\midrule
EEGNet  & 5.7 & 18.3 & 16.7 & 45.0 & 8.2 & 31.2 & 4.5 & 15.7 & 8.8 & 27.6 \\
w/ UBP & 7.5 & 25.0 & 23.2 & 53.7 & 16.0 & 41.5 & 5.5 & 19.2 & 13.1 & 34.9 \\
\midrule
TSconv & 12.0 & 31.5 & 30.0 & 58.7 & 20.3 & 51.5 & 13.0 & 34.8 & 18.8 & 44.1 \\
w/ UBP & 14.5 & 39.5 & 39.0 & 70.0 & 30.5 & 58.0 & 13.7 & 38.0 & 24.4 & 51.4 \\
\midrule
EEGProject   & 11.0 & 28.0 & 30.7 & 62.3 & 22.7 & 44.8 & 10.8 & 28.7 & 18.8 & 40.9 \\
w/ UBP & 15.5 & 37.3 & 41.0 & 74.8 & 24.8 & 55.3 & 15.3 & 34.5 & 24.1 & 50.4 \\
    \bottomrule
  \end{tabular}}
\end{table}

\begin{table}[t]
  \centering
  \caption{Top-1 and Top-5 Accuracy (\%) on THINGS-MEG with CLIP \textbf{ViT-g-14} With/Without UBP.}
  \label{tab:backbone_meg}
  \Huge
  \resizebox{1.0\linewidth}{!}{
  \begin{tabular}{lcccccccccc}
    \toprule
    & \multicolumn{2}{c}{Subject 1} & \multicolumn{2}{c}{Subject 2} & \multicolumn{2}{c}{Subject 3} & \multicolumn{2}{c}{Subject 4}& \multicolumn{2}{c}{Avg} \\
    \cmidrule(r){2-3} \cmidrule(r){4-5} \cmidrule(r){6-7} \cmidrule(r){8-9} \cmidrule(r){10-11}
    Backbone & top-1 & top-5 & top-1 & top-5 & top-1 & top-5 & top-1 & top-5 & top-1 & top-5 \\
    \midrule
    ShallowNet  & 9.5 & 25.7 & 17.5 & 47.0 & 17.3 & 42.7 & 8.5 & 27.3 & 13.2 & 35.7 \\
w/ UBP & 10.5 & 28.2 & 25.5 & 51.0 & 19.2 & 46.5 & 8.3 & 27.5 & 15.9 & 38.3 \\
\midrule
DeepNet  & 5.0 & 16.0 & 11.5 & 33.7 & 9.8 & 29.7 & 2.5 & 16.0 & 7.2 & 23.9 \\
w/ UBP & 4.3 & 17.2 & 14.2 & 42.5 & 9.0 & 33.8 & 6.2 & 21.2 & 8.4 & 28.7 \\
\midrule
EEGNet  & 6.5 & 20.3 & 17.0 & 46.3 & 12.0 & 32.5 & 4.5 & 20.3 & 10.0 & 29.8 \\
w/ UBP & 6.2 & 24.0 & 23.3 & 53.8 & 13.7 & 37.5 & 5.5 & 20.2 & 12.2 & 33.9 \\
\midrule
TSconv & 10.8 & 33.8 & 26.0 & 59.3 & 18.2 & 48.5 & 10.0 & 29.3 & 16.2 & 42.7 \\
w/ UBP & 11.5 & 34.7 & 32.0 & 65.0 & 24.2 & 56.5 & 12.5 & 37.3 & 20.1 & 48.4 \\
\midrule
EEGProject   & 11.0 & 28.7 & 27.5 & 59.5 & 19.5 & 44.3 & 10.7 & 33.2 & 17.2 & 41.4 \\
w/ UBP & 13.5 & 40.0 & 35.8 & 69.8 & 23.8 & 55.0 & 13.5 & 38.3 & 21.6 & 50.8 \\
    \bottomrule
  \end{tabular}}
\end{table}

\begin{table}[b]
  \centering
  \caption{Top-1 and Top-5 Accuracy (\%) on THINGS-MEG with CLIP \textbf{ViT-bigG-14} With/Without UBP.}
  \label{tab:backbone_meg}
  \Huge
  \resizebox{1.0\linewidth}{!}{
  \begin{tabular}{lcccccccccc}
    \toprule
    & \multicolumn{2}{c}{Subject 1} & \multicolumn{2}{c}{Subject 2} & \multicolumn{2}{c}{Subject 3} & \multicolumn{2}{c}{Subject 4}& \multicolumn{2}{c}{Avg} \\
    \cmidrule(r){2-3} \cmidrule(r){4-5} \cmidrule(r){6-7} \cmidrule(r){8-9} \cmidrule(r){10-11}
    Backbone & top-1 & top-5 & top-1 & top-5 & top-1 & top-5 & top-1 & top-5 & top-1 & top-5 \\
    \midrule
    ShallowNet  & 9.0 & 25.8 & 18.5 & 47.5 & 16.3 & 42.2 & 8.5 & 24.0 & 13.1 & 34.9 \\
w/ UBP & 6.5 & 25.5 & 25.5 & 55.8 & 25.5 & 51.2 & 9.8 & 28.7 & 16.8 & 40.3 \\
\midrule
DeepNet  & 4.2 & 18.2 & 13.0 & 36.0 & 11.3 & 29.0 & 5.5 & 16.5 & 8.5 & 24.9 \\
w/ UBP & 4.7 & 16.3 & 16.0 & 43.0 & 12.0 & 35.3 & 7.0 & 20.2 & 9.9 & 28.7 \\
\midrule
EEGNet  & 5.7 & 23.0 & 19.0 & 49.2 & 13.3 & 38.2 & 5.7 & 17.5 & 10.9 & 32.0 \\
w/ UBP & 8.2 & 28.5 & 23.7 & 61.0 & 16.3 & 50.2 & 4.5 & 22.0 & 13.2 & 40.4 \\
\midrule
TSconv & 11.2 & 34.5 & 28.5 & 59.5 & 23.7 & 49.8 & 11.7 & 32.0 & 18.8 & 43.9 \\
w/ UBP & 13.2 & 40.5 & 36.5 & 68.5 & 28.7 & 65.7 & 13.0 & 38.5 & 22.9 & 53.3 \\
\midrule
EEGProject   & 9.0 & 30.0 & 32.8 & 59.5 & 20.3 & 46.5 & 8.7 & 29.5 & 17.7 & 41.4 \\
w/ UBP & 13.5 & 33.3 & 40.3 & 70.0 & 26.7 & 63.2 & 18.5 & 36.8 & 24.7 & 50.8 \\
    \bottomrule
  \end{tabular}}
\end{table}

%% file: main.bbl
\begin{thebibliography}{82}
\providecommand{\natexlab}[1]{#1}
\providecommand{\url}[1]{\texttt{#1}}
\expandafter\ifx\csname urlstyle\endcsname\relax
  \providecommand{\doi}[1]{doi: #1}\else
  \providecommand{\doi}{doi: \begingroup \urlstyle{rm}\Url}\fi

\bibitem[Abdar et~al.(2021)Abdar, Pourpanah, Hussain, Rezazadegan, Liu, Ghavamzadeh, Fieguth, Cao, Khosravi, Acharya, et~al.]{abdar2021review}
Moloud Abdar, Farhad Pourpanah, Sadiq Hussain, Dana Rezazadegan, Li Liu, Mohammad Ghavamzadeh, Paul Fieguth, Xiaochun Cao, Abbas Khosravi, U~Rajendra Acharya, et~al.
\newblock A review of uncertainty quantification in deep learning: Techniques, applications and challenges.
\newblock \emph{Information fusion}, 76:\penalty0 243--297, 2021.

\bibitem[Aflalo et~al.(2015)Aflalo, Kellis, Klaes, Lee, Shi, Pejsa, Shanfield, Hayes-Jackson, Aisen, Heck, et~al.]{aflalo2015decoding}
Tyson Aflalo, Spencer Kellis, Christian Klaes, Brian Lee, Ying Shi, Kelsie Pejsa, Kathleen Shanfield, Stephanie Hayes-Jackson, Mindy Aisen, Christi Heck, et~al.
\newblock Decoding motor imagery from the posterior parietal cortex of a tetraplegic human.
\newblock \emph{Science}, 348\penalty0 (6237):\penalty0 906--910, 2015.

\bibitem[Bai et~al.(2023)Bai, Wang, Cao, Ge, Yuan, and Shan]{bai2023dreamdiffusion}
Yunpeng Bai, Xintao Wang, Yan-pei Cao, Yixiao Ge, Chun Yuan, and Ying Shan.
\newblock Dreamdiffusion: Generating high-quality images from brain eeg signals.
\newblock \emph{arXiv preprint arXiv:2306.16934}, 2023.

\bibitem[Beliy et~al.(2019)Beliy, Gaziv, Hoogi, Strappini, Golan, and Irani]{NEURIPS2019_7d2be41b}
Roman Beliy, Guy Gaziv, Assaf Hoogi, Francesca Strappini, Tal Golan, and Michal Irani.
\newblock From voxels to pixels and back: Self-supervision in natural-image reconstruction from fmri.
\newblock In \emph{Advances in Neural Information Processing Systems}. Curran Associates, Inc., 2019.

\bibitem[Benchetrit et~al.(2023)Benchetrit, Banville, and King]{benchetrit2023brain}
Yohann Benchetrit, Hubert Banville, and Jean-R{\'e}mi King.
\newblock Brain decoding: toward real-time reconstruction of visual perception.
\newblock \emph{arXiv preprint arXiv:2310.19812}, 2023.

\bibitem[Block(2011)]{block2011perceptual}
Ned Block.
\newblock Perceptual consciousness overflows cognitive access.
\newblock \emph{Trends in cognitive sciences}, 15\penalty0 (12):\penalty0 567--575, 2011.

\bibitem[Buschman et~al.(2011)Buschman, Siegel, Roy, and Miller]{buschman2011neural}
Timothy~J Buschman, Markus Siegel, Jefferson~E Roy, and Earl~K Miller.
\newblock Neural substrates of cognitive capacity limitations.
\newblock \emph{Proceedings of the National Academy of Sciences}, 108\penalty0 (27):\penalty0 11252--11255, 2011.

\bibitem[Cavanagh and Alvarez(2005)]{cavanagh2005tracking}
Patrick Cavanagh and George~A Alvarez.
\newblock Tracking multiple targets with multifocal attention.
\newblock \emph{Trends in cognitive sciences}, 9\penalty0 (7):\penalty0 349--354, 2005.

\bibitem[Charpentier et~al.(2020)Charpentier, Z{\"u}gner, and G{\"u}nnemann]{charpentier2020posterior}
Bertrand Charpentier, Daniel Z{\"u}gner, and Stephan G{\"u}nnemann.
\newblock Posterior network: Uncertainty estimation without ood samples via density-based pseudo-counts.
\newblock \emph{Advances in neural information processing systems}, 33:\penalty0 1356--1367, 2020.

\bibitem[Chen et~al.(2024)Chen, He, Liu, and Yang]{chen2024visual}
Hongzhou Chen, Lianghua He, Yihang Liu, and Longzhen Yang.
\newblock Visual neural decoding via improved visual-eeg semantic consistency.
\newblock \emph{arXiv preprint arXiv:2408.06788}, 2024.

\bibitem[Chen et~al.(2020)Chen, Kornblith, Norouzi, and Hinton]{chen2020simple}
Ting Chen, Simon Kornblith, Mohammad Norouzi, and Geoffrey Hinton.
\newblock A simple framework for contrastive learning of visual representations.
\newblock In \emph{International conference on machine learning}, pages 1597--1607. PMLR, 2020.

\bibitem[Chen et~al.(2023)Chen, Qing, Xiang, Yue, and Zhou]{chen2023seeing}
Zijiao Chen, Jiaxin Qing, Tiange Xiang, Wan~Lin Yue, and Juan~Helen Zhou.
\newblock Seeing beyond the brain: Conditional diffusion model with sparse masked modeling for vision decoding.
\newblock In \emph{Proceedings of the IEEE/CVF Conference on Computer Vision and Pattern Recognition}, pages 22710--22720, 2023.

\bibitem[Cohen et~al.(2016)Cohen, Dennett, and Kanwisher]{cohen2016bandwidth}
Michael~A Cohen, Daniel~C Dennett, and Nancy Kanwisher.
\newblock What is the bandwidth of perceptual experience?
\newblock \emph{Trends in cognitive sciences}, 20\penalty0 (5):\penalty0 324--335, 2016.

\bibitem[Curcio et~al.(1990)Curcio, Sloan, Kalina, and Hendrickson]{curcio1990human}
Christine~A Curcio, Kenneth~R Sloan, Robert~E Kalina, and Anita~E Hendrickson.
\newblock Human photoreceptor topography.
\newblock \emph{Journal of comparative neurology}, 292\penalty0 (4):\penalty0 497--523, 1990.

\bibitem[Du et~al.(2023)Du, Fu, Li, and He]{du2023decoding}
Changde Du, Kaicheng Fu, Jinpeng Li, and Huiguang He.
\newblock Decoding visual neural representations by multimodal learning of brain-visual-linguistic features.
\newblock \emph{IEEE Transactions on Pattern Analysis and Machine Intelligence}, 45\penalty0 (9):\penalty0 10760--10777, 2023.

\bibitem[Duan et~al.(2024)Duan, Chau, Wang, Wang, and Lin]{duan2024dewave}
Yiqun Duan, Charles Chau, Zhen Wang, Yu-Kai Wang, and Chin-teng Lin.
\newblock Dewave: Discrete encoding of eeg waves for eeg to text translation.
\newblock \emph{Advances in Neural Information Processing Systems}, 36, 2024.

\bibitem[Dux and Marois(2009)]{dux2009humans}
Paul~E Dux and R{\'e}ne Marois.
\newblock How humans search for targets through time: A review of data and theory from the attentional blink.
\newblock \emph{Attention, perception \& psychophysics}, 71\penalty0 (8):\penalty0 1683, 2009.

\bibitem[Fang et~al.(2024)Fang, Zheng, and Pan]{fang2024alleviating}
Tao Fang, Qian Zheng, and Gang Pan.
\newblock Alleviating the semantic gap for generalized fmri-to-image reconstruction.
\newblock \emph{Advances in Neural Information Processing Systems}, 36, 2024.

\bibitem[Gao et~al.(2021)Gao, Yao, and Chen]{gao2021simcse}
Tianyu Gao, Xingcheng Yao, and Danqi Chen.
\newblock Simcse: Simple contrastive learning of sentence embeddings.
\newblock In \emph{Proceedings of the 2021 Conference on Empirical Methods in Natural Language Processing}, pages 6894--6910, 2021.

\bibitem[Gaziv et~al.(2022)Gaziv, Beliy, Granot, Hoogi, Strappini, Golan, and Irani]{gaziv2022self}
Guy Gaziv, Roman Beliy, Niv Granot, Assaf Hoogi, Francesca Strappini, Tal Golan, and Michal Irani.
\newblock Self-supervised natural image reconstruction and large-scale semantic classification from brain activity.
\newblock \emph{NeuroImage}, 254:\penalty0 119121, 2022.

\bibitem[Gifford et~al.(2022)Gifford, Dwivedi, Roig, and Cichy]{gifford2022large}
Alessandro~T Gifford, Kshitij Dwivedi, Gemma Roig, and Radoslaw~M Cichy.
\newblock A large and rich eeg dataset for modeling human visual object recognition.
\newblock \emph{NeuroImage}, 264:\penalty0 119754, 2022.

\bibitem[Girdhar et~al.(2023)Girdhar, El-Nouby, Liu, Singh, Alwala, Joulin, and Misra]{girdhar2023imagebind}
Rohit Girdhar, Alaaeldin El-Nouby, Zhuang Liu, Mannat Singh, Kalyan~Vasudev Alwala, Armand Joulin, and Ishan Misra.
\newblock Imagebind: One embedding space to bind them all.
\newblock In \emph{Proceedings of the IEEE/CVF Conference on Computer Vision and Pattern Recognition}, pages 15180--15190, 2023.

\bibitem[Grootswagers et~al.(2019)Grootswagers, Robinson, and Carlson]{grootswagers2019representational}
Tijl Grootswagers, Amanda~K Robinson, and Thomas~A Carlson.
\newblock The representational dynamics of visual objects in rapid serial visual processing streams.
\newblock \emph{NeuroImage}, 188:\penalty0 668--679, 2019.

\bibitem[Guzhov et~al.(2022)Guzhov, Raue, Hees, and Dengel]{guzhov2022audioclip}
Andrey Guzhov, Federico Raue, J{\"o}rn Hees, and Andreas Dengel.
\newblock Audioclip: Extending clip to image, text and audio.
\newblock In \emph{ICASSP 2022-2022 IEEE International Conference on Acoustics, Speech and Signal Processing (ICASSP)}, pages 976--980. IEEE, 2022.

\bibitem[Han et~al.(2022)Han, Zhang, Fu, and Zhou]{han2022trusted}
Zongbo Han, Changqing Zhang, Huazhu Fu, and Joey~Tianyi Zhou.
\newblock Trusted multi-view classification with dynamic evidential fusion.
\newblock \emph{IEEE transactions on pattern analysis and machine intelligence}, 45\penalty0 (2):\penalty0 2551--2566, 2022.

\bibitem[Hebart et~al.(2023)Hebart, Contier, Teichmann, Rockter, Zheng, Kidder, Corriveau, Vaziri-Pashkam, and Baker]{hebart2023things}
Martin~N Hebart, Oliver Contier, Lina Teichmann, Adam~H Rockter, Charles~Y Zheng, Alexis Kidder, Anna Corriveau, Maryam Vaziri-Pashkam, and Chris~I Baker.
\newblock Things-data, a multimodal collection of large-scale datasets for investigating object representations in human brain and behavior.
\newblock \emph{Elife}, 12:\penalty0 e82580, 2023.

\bibitem[Hubel and Wiesel(1968)]{hubel1968receptive}
David~H Hubel and Torsten~N Wiesel.
\newblock Receptive fields and functional architecture of monkey striate cortex.
\newblock \emph{The Journal of physiology}, 195\penalty0 (1):\penalty0 215--243, 1968.

\bibitem[Hubel et~al.(1959)Hubel, Wiesel, et~al.]{hubel1959receptive}
David~H Hubel, Torsten~N Wiesel, et~al.
\newblock Receptive fields of single neurones in the cat’s striate cortex.
\newblock \emph{J physiol}, 148\penalty0 (3):\penalty0 574--591, 1959.

\bibitem[Ilharco et~al.(2021)Ilharco, Wortsman, Wightman, Gordon, Carlini, Taori, Dave, Shankar, Namkoong, Miller, Hajishirzi, Farhadi, and Schmidt]{ilharco_gabriel_2021_5143773}
Gabriel Ilharco, Mitchell Wortsman, Ross Wightman, Cade Gordon, Nicholas Carlini, Rohan Taori, Achal Dave, Vaishaal Shankar, Hongseok Namkoong, John Miller, Hannaneh Hajishirzi, Ali Farhadi, and Ludwig Schmidt.
\newblock Openclip, 2021.
\newblock If you use this software, please cite it as below.

\bibitem[Intraub(1981)]{intraub1981rapid}
Helene Intraub.
\newblock Rapid conceptual identification of sequentially presented pictures.
\newblock \emph{Journal of Experimental Psychology: Human Perception and Performance}, 7\penalty0 (3):\penalty0 604, 1981.

\bibitem[Jia et~al.(2021)Jia, Yang, Xia, Chen, Parekh, Pham, Le, Sung, Li, and Duerig]{jia2021scaling}
Chao Jia, Yinfei Yang, Ye Xia, Yi-Ting Chen, Zarana Parekh, Hieu Pham, Quoc Le, Yun-Hsuan Sung, Zhen Li, and Tom Duerig.
\newblock Scaling up visual and vision-language representation learning with noisy text supervision.
\newblock In \emph{International conference on machine learning}, pages 4904--4916. PMLR, 2021.

\bibitem[Keysers et~al.(2001)Keysers, Xiao, F{\"o}ldi{\'a}k, and Perrett]{keysers2001speed}
Christian Keysers, D-K Xiao, Peter F{\"o}ldi{\'a}k, and David~I Perrett.
\newblock The speed of sight.
\newblock \emph{Journal of cognitive neuroscience}, 13\penalty0 (1):\penalty0 90--101, 2001.

\bibitem[Kowler(2011)]{kowler2011eye}
Eileen Kowler.
\newblock Eye movements: The past 25 years.
\newblock \emph{Vision research}, 51\penalty0 (13):\penalty0 1457--1483, 2011.

\bibitem[Lawhern et~al.(2018)Lawhern, Solon, Waytowich, Gordon, Hung, and Lance]{lawhern2018eegnet}
Vernon~J Lawhern, Amelia~J Solon, Nicholas~R Waytowich, Stephen~M Gordon, Chou~P Hung, and Brent~J Lance.
\newblock Eegnet: a compact convolutional neural network for eeg-based brain--computer interfaces.
\newblock \emph{Journal of neural engineering}, 15\penalty0 (5):\penalty0 056013, 2018.

\bibitem[Lei et~al.(2024)Lei, Ge, Yi, Zhang, Gao, Sun, Ge, Shan, and Shou]{lei2024vit}
Weixian Lei, Yixiao Ge, Kun Yi, Jianfeng Zhang, Difei Gao, Dylan Sun, Yuying Ge, Ying Shan, and Mike~Zheng Shou.
\newblock Vit-lens: Towards omni-modal representations.
\newblock In \emph{Proceedings of the IEEE/CVF Conference on Computer Vision and Pattern Recognition}, pages 26647--26657, 2024.

\bibitem[Lettvin et~al.(1959)Lettvin, Maturana, McCulloch, and Pitts]{lettvin1959frog}
Jerome~Y Lettvin, Humberto~R Maturana, Warren~S McCulloch, and Walter~H Pitts.
\newblock What the frog's eye tells the frog's brain.
\newblock \emph{Proceedings of the IRE}, 47\penalty0 (11):\penalty0 1940--1951, 1959.

\bibitem[Li et~al.(2024)Li, Wei, Li, Zou, and Liu]{li2024visual}
Dongyang Li, Chen Wei, Shiying Li, Jiachen Zou, and Quanying Liu.
\newblock Visual decoding and reconstruction via eeg embeddings with guided diffusion.
\newblock \emph{Advances in Neural Information Processing Systems}, 2024.

\bibitem[Li et~al.(2022)Li, Zhang, Tiwari, Song, Hu, Yang, Zhao, Kumar, and Marttinen]{li2022eeg}
Xiang Li, Yazhou Zhang, Prayag Tiwari, Dawei Song, Bin Hu, Meihong Yang, Zhigang Zhao, Neeraj Kumar, and Pekka Marttinen.
\newblock Eeg based emotion recognition: A tutorial and review.
\newblock \emph{ACM Computing Surveys}, 55\penalty0 (4):\penalty0 1--57, 2022.

\bibitem[Liang et~al.(2018)Liang, Fratzl, Goldey, Ramesh, Sugden, Morgan, Chen, and Andermann]{liang2018fine}
Liang Liang, Alex Fratzl, Glenn Goldey, Rohan~N Ramesh, Arthur~U Sugden, Josh~L Morgan, Chinfei Chen, and Mark~L Andermann.
\newblock A fine-scale functional logic to convergence from retina to thalamus.
\newblock \emph{Cell}, 173\penalty0 (6):\penalty0 1343--1355, 2018.

\bibitem[Liang et~al.(2022)Liang, Zhang, Kwon, Yeung, and Zou]{liang2022mind}
Victor~Weixin Liang, Yuhui Zhang, Yongchan Kwon, Serena Yeung, and James~Y Zou.
\newblock Mind the gap: Understanding the modality gap in multi-modal contrastive representation learning.
\newblock \emph{Advances in Neural Information Processing Systems}, 35:\penalty0 17612--17625, 2022.

\bibitem[Livingstone and Hubel(1984)]{livingstone1984anatomy}
Margaret~S Livingstone and David~H Hubel.
\newblock Anatomy and physiology of a color system in the primate visual cortex.
\newblock \emph{Journal of Neuroscience}, 4\penalty0 (1):\penalty0 309--356, 1984.

\bibitem[Loshchilov and Hutter(2019)]{loshchilov2018decoupled}
Ilya Loshchilov and Frank Hutter.
\newblock Decoupled weight decay regularization.
\newblock In \emph{International Conference on Learning Representations}, 2019.

\bibitem[Lu et~al.(2023{\natexlab{a}})Lu, Zhu, Zhai, Zheng, and Cao]{lu2023uncertainty}
Fan Lu, Kai Zhu, Wei Zhai, Kecheng Zheng, and Yang Cao.
\newblock Uncertainty-aware optimal transport for semantically coherent out-of-distribution detection.
\newblock In \emph{Proceedings of the IEEE/CVF Conference on Computer Vision and Pattern Recognition}, pages 3282--3291, 2023{\natexlab{a}}.

\bibitem[Lu et~al.(2023{\natexlab{b}})Lu, Du, Zhou, Wang, and He]{lu2023minddiffuser}
Yizhuo Lu, Changde Du, Qiongyi Zhou, Dianpeng Wang, and Huiguang He.
\newblock Minddiffuser: Controlled image reconstruction from human brain activity with semantic and structural diffusion.
\newblock In \emph{Proceedings of the 31st ACM International Conference on Multimedia}, pages 5899--5908, 2023{\natexlab{b}}.

\bibitem[Luck and Vogel(2013)]{luck2013visual}
Steven~J Luck and Edward~K Vogel.
\newblock Visual working memory capacity: from psychophysics and neurobiology to individual differences.
\newblock \emph{Trends in cognitive sciences}, 17\penalty0 (8):\penalty0 391--400, 2013.

\bibitem[Ma et~al.(2025)Ma, Chen, Wang, and Zhang]{ma2025estimating}
Huan Ma, Jingdong Chen, Guangyu Wang, and Changqing Zhang.
\newblock Estimating llm uncertainty with logits.
\newblock \emph{arXiv preprint arXiv:2502.00290}, 2025.

\bibitem[Nagrani et~al.(2022)Nagrani, Seo, Seybold, Hauth, Manen, Sun, and Schmid]{nagrani2022learning}
Arsha Nagrani, Paul~Hongsuck Seo, Bryan Seybold, Anja Hauth, Santiago Manen, Chen Sun, and Cordelia Schmid.
\newblock Learning audio-video modalities from image captions.
\newblock In \emph{European Conference on Computer Vision}, pages 407--426. Springer, 2022.

\bibitem[Nauhaus et~al.(2012)Nauhaus, Nielsen, Disney, and Callaway]{nauhaus2012orthogonal}
Ian Nauhaus, Kristina~J Nielsen, Anita~A Disney, and Edward~M Callaway.
\newblock Orthogonal micro-organization of orientation and spatial frequency in primate primary visual cortex.
\newblock \emph{Nature neuroscience}, 15\penalty0 (12):\penalty0 1683--1690, 2012.

\bibitem[Nuwer et~al.(1998)Nuwer, Comi, Emerson, Fuglsang-Frederiksen, Gu{\'e}rit, Hinrichs, Ikeda, Luccas, and Rappelsburger]{nuwer1998ifcn}
Marc~R Nuwer, Giancarlo Comi, Ronald Emerson, Anders Fuglsang-Frederiksen, Jean-Michel Gu{\'e}rit, Hermann Hinrichs, Akio Ikeda, Fransisco Jose~C Luccas, and Peter Rappelsburger.
\newblock Ifcn standards for digital recording of clinical eeg.
\newblock \emph{Electroencephalography and clinical Neurophysiology}, 106\penalty0 (3):\penalty0 259--261, 1998.

\bibitem[Posner et~al.(1980)Posner, Snyder, and Davidson]{posner1980attention}
Michael~I Posner, Charles~R Snyder, and Brian~J Davidson.
\newblock Attention and the detection of signals.
\newblock \emph{Journal of experimental psychology: General}, 109\penalty0 (2):\penalty0 160, 1980.

\bibitem[Pylyshyn(1999)]{pylyshyn1999vision}
Zenon Pylyshyn.
\newblock Is vision continuous with cognition?: The case for cognitive impenetrability of visual perception.
\newblock \emph{Behavioral and brain sciences}, 22\penalty0 (3):\penalty0 341--365, 1999.

\bibitem[Quan et~al.(2024)Quan, Wang, Tian, Ma, and Yang]{quan2024psychometry}
Ruijie Quan, Wenguan Wang, Zhibo Tian, Fan Ma, and Yi Yang.
\newblock Psychometry: An omnifit model for image reconstruction from human brain activity.
\newblock In \emph{Proceedings of the IEEE/CVF Conference on Computer Vision and Pattern Recognition}, pages 233--243, 2024.

\bibitem[Radford et~al.(2021)Radford, Kim, Hallacy, Ramesh, Goh, Agarwal, Sastry, Askell, Mishkin, Clark, et~al.]{radford2021learning}
Alec Radford, Jong~Wook Kim, Chris Hallacy, Aditya Ramesh, Gabriel Goh, Sandhini Agarwal, Girish Sastry, Amanda Askell, Pamela Mishkin, Jack Clark, et~al.
\newblock Learning transferable visual models from natural language supervision.
\newblock In \emph{International conference on machine learning}, pages 8748--8763. PMLR, 2021.

\bibitem[Raichle et~al.(2001)Raichle, MacLeod, Snyder, Powers, Gusnard, and Shulman]{raichle2001default}
Marcus~E Raichle, Ann~Mary MacLeod, Abraham~Z Snyder, William~J Powers, Debra~A Gusnard, and Gordon~L Shulman.
\newblock A default mode of brain function.
\newblock \emph{Proceedings of the national academy of sciences}, 98\penalty0 (2):\penalty0 676--682, 2001.

\bibitem[Rizve et~al.()Rizve, Duarte, Rawat, and Shah]{rizvedefense}
Mamshad~Nayeem Rizve, Kevin Duarte, Yogesh~S Rawat, and Mubarak Shah.
\newblock In defense of pseudo-labeling: An uncertainty-aware pseudo-label selection framework for semi-supervised learning.
\newblock In \emph{International Conference on Learning Representations}.

\bibitem[Schirrmeister et~al.(2017)Schirrmeister, Springenberg, Fiederer, Glasstetter, Eggensperger, Tangermann, Hutter, Burgard, and Ball]{schirrmeister2017deep}
Robin~Tibor Schirrmeister, Jost~Tobias Springenberg, Lukas Dominique~Josef Fiederer, Martin Glasstetter, Katharina Eggensperger, Michael Tangermann, Frank Hutter, Wolfram Burgard, and Tonio Ball.
\newblock Deep learning with convolutional neural networks for eeg decoding and visualization.
\newblock \emph{Human brain mapping}, 38\penalty0 (11):\penalty0 5391--5420, 2017.

\bibitem[Scotti et~al.(2024{\natexlab{a}})Scotti, Banerjee, Goode, Shabalin, Nguyen, Dempster, Verlinde, Yundler, Weisberg, Norman, et~al.]{scotti2024reconstructing}
Paul Scotti, Atmadeep Banerjee, Jimmie Goode, Stepan Shabalin, Alex Nguyen, Aidan Dempster, Nathalie Verlinde, Elad Yundler, David Weisberg, Kenneth Norman, et~al.
\newblock Reconstructing the mind's eye: fmri-to-image with contrastive learning and diffusion priors.
\newblock \emph{Advances in Neural Information Processing Systems}, 36, 2024{\natexlab{a}}.

\bibitem[Scotti et~al.(2024{\natexlab{b}})Scotti, Tripathy, Villanueva, Kneeland, Chen, Narang, Santhirasegaran, Xu, Naselaris, Norman, et~al.]{scotti2024mindeye2}
Paul~S Scotti, Mihir Tripathy, Cesar Kadir~Torrico Villanueva, Reese Kneeland, Tong Chen, Ashutosh Narang, Charan Santhirasegaran, Jonathan Xu, Thomas Naselaris, Kenneth~A Norman, et~al.
\newblock Mindeye2: Shared-subject models enable fmri-to-image with 1 hour of data.
\newblock \emph{arXiv preprint arXiv:2403.11207}, 2024{\natexlab{b}}.

\bibitem[Simons and Levin(1997)]{simons1997change}
Daniel~J Simons and Daniel~T Levin.
\newblock Change blindness.
\newblock \emph{Trends in cognitive sciences}, 1\penalty0 (7):\penalty0 261--267, 1997.

\bibitem[Song et~al.(2024)Song, Liu, Li, Shi, Wang, and Gao]{song2024decoding}
Yonghao Song, Bingchuan Liu, Xiang Li, Nanlin Shi, Yijun Wang, and Xiaorong Gao.
\newblock Decoding natural images from {EEG} for object recognition.
\newblock In \emph{The Twelfth International Conference on Learning Representations}, 2024.

\bibitem[Stevens et~al.(2024)Stevens, Wu, Thompson, Campolongo, Song, Carlyn, Dong, Dahdul, Stewart, Berger-Wolf, et~al.]{stevens2024bioclip}
Samuel Stevens, Jiaman Wu, Matthew~J Thompson, Elizabeth~G Campolongo, Chan~Hee Song, David~Edward Carlyn, Li Dong, Wasila~M Dahdul, Charles Stewart, Tanya Berger-Wolf, et~al.
\newblock Bioclip: A vision foundation model for the tree of life.
\newblock In \emph{Proceedings of the IEEE/CVF Conference on Computer Vision and Pattern Recognition}, pages 19412--19424, 2024.

\bibitem[Takagi and Nishimoto(2023)]{takagi2023high}
Yu Takagi and Shinji Nishimoto.
\newblock High-resolution image reconstruction with latent diffusion models from human brain activity.
\newblock In \emph{Proceedings of the IEEE/CVF Conference on Computer Vision and Pattern Recognition}, pages 14453--14463, 2023.

\bibitem[Tsao et~al.(2006)Tsao, Freiwald, Tootell, and Livingstone]{tsao2006cortical}
Doris~Y Tsao, Winrich~A Freiwald, Roger~BH Tootell, and Margaret~S Livingstone.
\newblock A cortical region consisting entirely of face-selective cells.
\newblock \emph{Science}, 311\penalty0 (5761):\penalty0 670--674, 2006.

\bibitem[Van~Essen et~al.(1992)Van~Essen, Anderson, and Felleman]{van1992information}
David~C Van~Essen, Charles~H Anderson, and Daniel~J Felleman.
\newblock Information processing in the primate visual system: an integrated systems perspective.
\newblock \emph{Science}, 255\penalty0 (5043):\penalty0 419--423, 1992.

\bibitem[Vicchietti et~al.(2023)Vicchietti, Ramos, Betting, and Campanharo]{vicchietti2023computational}
M{\'a}rio~L Vicchietti, Fernando~M Ramos, Luiz~E Betting, and Andriana~SLO Campanharo.
\newblock Computational methods of eeg signals analysis for alzheimer’s disease classification.
\newblock \emph{Scientific Reports}, 13\penalty0 (1):\penalty0 8184, 2023.

\bibitem[Wang et~al.(1996)Wang, Tanaka, and Tanifuji]{wang1996optical}
Gang Wang, Keiji Tanaka, and Manabu Tanifuji.
\newblock Optical imaging of functional organization in the monkey inferotemporal cortex.
\newblock \emph{Science}, 272\penalty0 (5268):\penalty0 1665--1668, 1996.

\bibitem[Wang et~al.(2024)Wang, Liu, Tan, and Wang]{wang2024mindbridge}
Shizun Wang, Songhua Liu, Zhenxiong Tan, and Xinchao Wang.
\newblock Mindbridge: A cross-subject brain decoding framework.
\newblock In \emph{Proceedings of the IEEE/CVF Conference on Computer Vision and Pattern Recognition}, pages 11333--11342, 2024.

\bibitem[Wang et~al.(2019)Wang, Ma, Chen, Luo, Yi, and Bailey]{wang2019symmetric}
Yisen Wang, Xingjun Ma, Zaiyi Chen, Yuan Luo, Jinfeng Yi, and James Bailey.
\newblock Symmetric cross entropy for robust learning with noisy labels.
\newblock In \emph{Proceedings of the IEEE/CVF international conference on computer vision}, pages 322--330, 2019.

\bibitem[Wang et~al.(2021)Wang, Li, Guo, and Wang]{wang2021combating}
Zhenyu Wang, Ya-Li Li, Ye Guo, and Shengjin Wang.
\newblock Combating noise: semi-supervised learning by region uncertainty quantification.
\newblock \emph{Advances in Neural Information Processing Systems}, 34:\penalty0 9534--9545, 2021.

\bibitem[Wang et~al.(2022)Wang, Wu, Agarwal, and Sun]{wang2022medclip}
Zifeng Wang, Zhenbang Wu, Dinesh Agarwal, and Jimeng Sun.
\newblock Medclip: Contrastive learning from unpaired medical images and text.
\newblock \emph{arXiv preprint arXiv:2210.10163}, 2022.

\bibitem[Wang et~al.(2023)Wang, Zhao, Huang, Liu, Yin, Tang, Li, Wang, Zhang, and Zhao]{wang2023connecting}
Zehan Wang, Yang Zhao, Haifeng Huang, Jiageng Liu, Aoxiong Yin, Li Tang, Linjun Li, Yongqi Wang, Ziang Zhang, and Zhou Zhao.
\newblock Connecting multi-modal contrastive representations.
\newblock \emph{Advances in Neural Information Processing Systems}, 36:\penalty0 22099--22114, 2023.

\bibitem[Whitney and Yamanashi~Leib(2018)]{whitney2018ensemble}
David Whitney and Allison Yamanashi~Leib.
\newblock Ensemble perception.
\newblock \emph{Annual review of psychology}, 69\penalty0 (1):\penalty0 105--129, 2018.

\bibitem[Wolfe(1994)]{wolfe1994guided}
Jeremy~M Wolfe.
\newblock Guided search 2.0 a revised model of visual search.
\newblock \emph{Psychonomic bulletin \& review}, 1:\penalty0 202--238, 1994.

\bibitem[Xia et~al.(2024{\natexlab{a}})Xia, de~Charette, Oztireli, and Xue]{xia2024dream}
Weihao Xia, Raoul de Charette, Cengiz Oztireli, and Jing-Hao Xue.
\newblock Dream: Visual decoding from reversing human visual system.
\newblock In \emph{Proceedings of the IEEE/CVF Winter Conference on Applications of Computer Vision}, pages 8226--8235, 2024{\natexlab{a}}.

\bibitem[Xia et~al.(2024{\natexlab{b}})Xia, de~Charette, Öztireli, and Xue]{xia2024umbrae}
Weihao Xia, Raoul de Charette, Cengiz Öztireli, and Jing-Hao Xue.
\newblock Umbrae: Unified multimodal brain decoding.
\newblock In \emph{European Conference on Computer Vision (ECCV)}, 2024{\natexlab{b}}.

\bibitem[Xu et~al.(2024)Xu, Si, Guan, Zhao, Wu, and Gao]{xu2024reliable}
Cai Xu, Jiajun Si, Ziyu Guan, Wei Zhao, Yue Wu, and Xiyue Gao.
\newblock Reliable conflictive multi-view learning.
\newblock In \emph{Proceedings of the AAAI conference on artificial intelligence}, pages 16129--16137, 2024.

\bibitem[Xue et~al.(2023)Xue, Gao, Xing, Mart{\'\i}n-Mart{\'\i}n, Wu, Xiong, Xu, Niebles, and Savarese]{xue2023ulip}
Le Xue, Mingfei Gao, Chen Xing, Roberto Mart{\'\i}n-Mart{\'\i}n, Jiajun Wu, Caiming Xiong, Ran Xu, Juan~Carlos Niebles, and Silvio Savarese.
\newblock Ulip: Learning a unified representation of language, images, and point clouds for 3d understanding.
\newblock In \emph{Proceedings of the IEEE/CVF conference on computer vision and pattern recognition}, pages 1179--1189, 2023.

\bibitem[Yang et~al.(2022)Yang, Wang, Zou, Zhou, Ding, Peng, Wang, Chen, Li, Sun, et~al.]{yang2022openood}
Jingkang Yang, Pengyun Wang, Dejian Zou, Zitang Zhou, Kunyuan Ding, Wenxuan Peng, Haoqi Wang, Guangyao Chen, Bo Li, Yiyou Sun, et~al.
\newblock Openood: Benchmarking generalized out-of-distribution detection.
\newblock \emph{Advances in Neural Information Processing Systems}, 35:\penalty0 32598--32611, 2022.

\bibitem[You et~al.(2020)You, Chen, Sui, Chen, Wang, and Shen]{you2020graph}
Yuning You, Tianlong Chen, Yongduo Sui, Ting Chen, Zhangyang Wang, and Yang Shen.
\newblock Graph contrastive learning with augmentations.
\newblock \emph{Advances in neural information processing systems}, 33:\penalty0 5812--5823, 2020.

\bibitem[Zhang et~al.(2023)Zhang, Wu, Zhang, Hu, Fu, Zhou, and Peng]{zhang2023provable}
Qingyang Zhang, Haitao Wu, Changqing Zhang, Qinghua Hu, Huazhu Fu, Joey~Tianyi Zhou, and Xi Peng.
\newblock Provable dynamic fusion for low-quality multimodal data.
\newblock In \emph{International conference on machine learning}, pages 41753--41769. PMLR, 2023.

\bibitem[Zhao et~al.(2020)Zhao, Chen, Hu, and Cho]{zhao2020uncertainty}
Xujiang Zhao, Feng Chen, Shu Hu, and Jin-Hee Cho.
\newblock Uncertainty aware semi-supervised learning on graph data.
\newblock \emph{Advances in Neural Information Processing Systems}, 33:\penalty0 12827--12836, 2020.

\bibitem[Zhou et~al.(2024)Zhou, Du, Wang, and He]{zhou2024clip}
Qiongyi Zhou, Changde Du, Shengpei Wang, and Huiguang He.
\newblock Clip-mused: Clip-guided multi-subject visual neural information semantic decoding.
\newblock \emph{arXiv preprint arXiv:2402.08994}, 2024.

\end{thebibliography}
